\documentclass[11pt,draftcls,onecolumn]{IEEEtran}
\usepackage{qtree}
\usepackage{tree-dvips}

\usepackage{graphicx}
\usepackage{amsmath,amssymb} 
\usepackage{lineno}
\usepackage{color}
\usepackage{algorithmic}
\usepackage{overpic}
\usepackage[ruled,vlined]{algorithm2e}

\usepackage{amsthm}

\usepackage{multirow}

\usepackage{url}

\usepackage{subfigure}
\usepackage{overpic}
\usepackage{graphicx}

\usepackage{color}

\usepackage{multirow}

\def\rv{\color{blue}}

\def\x{{\boldsymbol x}}

\def\k{{\boldsymbol k}}

\def\I{{\boldsymbol I}}

\def\z{{\boldsymbol z}}

\def\u{{\boldsymbol u}}
\def\v{{\boldsymbol v}}

\def\W{{\boldsymbol W}}

\def\K{{\boldsymbol K}}

\def\II{{\boldsymbol I}}

\def\Z{{\boldsymbol Z}}

\def\y{{\boldsymbol y}}
\def\u{{\boldsymbol u}}

\def\V{{\boldsymbol V}}

\def\D{{\boldsymbol D}}
\def\HH{{\boldsymbol H}}

\def\gam{{\boldsymbol \gamma}}
\def\sig{{\boldsymbol{\Sigma}}}
\def\bmu{{\boldsymbol \mu}}

\def\x{{\boldsymbol x}}

\def\k{{\boldsymbol k}}

\def\GM{{\boldsymbol \Gamma}}

\def\z{{\boldsymbol z}}

\def\u{{\boldsymbol u}}
\def\v{{\boldsymbol v}}

\def\A{{\boldsymbol A}}
\def\a{{\boldsymbol a}}

\def\W{{\boldsymbol W}}

\def\K{{\boldsymbol K}}

\def\II{{\boldsymbol I}}

\def\Z{{\boldsymbol Z}}

\def\y{{\boldsymbol y}}

\def\V{{\boldsymbol V}}

\def\D{{\boldsymbol D}}
\def\HH{{\boldsymbol H}}
\def\h{{\boldsymbol h}}

\def\rv{}

\usepackage{subfigure}


\graphicspath{{Figures/}}

\ifCLASSINFOpdf
\else
\fi

\begin{document}
%
\title{Image Super-Resolution via Sparse Bayesian Modeling of Natural Images}
%
%

\author{Haichao Zhang,
        David Wipf and
        Yanning Zhang
}

\maketitle

\begin{abstract}
Image super-resolution (SR) is one of the long-standing and active topics in image processing community.
A large body of works for image super resolution formulate the problem with Bayesian modeling techniques and then obtain its Maximum-A-Posteriori (MAP) solution, which actually boils down to a regularized regression task over separable regularization term.
Although straightforward, this approach  cannot exploit the full potential offered by the probabilistic modeling, as only the posterior mode is sought.
Also, the separable property of the regularization term can not capture any correlations between the sparse coefficients, which sacrifices much on its modeling accuracy.
We propose a Bayesian image SR algorithm via sparse modeling of natural images.
The sparsity property of the latent high resolution image is exploited by introducing latent variables into the high-order  Markov Random Field (MRF) which capture the content adaptive variance by pixel-wise adaptation.
The high-resolution image is estimated via Empirical Bayesian estimation scheme, which is substantially faster  than our previous approach based on Markov Chain Monte Carlo sampling~\cite{NBSR_Zhang}.
It is shown that the actual cost function for the proposed approach actually incorporates a non-factorial regularization term over the sparse coefficients.
 Experimental results indicate that the proposed method can generate competitive or better results than \emph{state-of-the-art} SR algorithms.
\end{abstract}


%
\IEEEpeerreviewmaketitle

\section{Introduction}
\label{sec:intro}
Super Resolution (SR) is one of the very active research topics in image processing community.
In practice, many tasks such as astronomical observation and medical diagnostic   rely on high quality images for reliable  and accurate analysis as well as prediction.
However, in many practical situations, due to the inherent limitation of the optical system or other factors, the observed images are often of low resolution, thus limiting the subsequent tasks based on them.
Image SR aims to estimate a high-resolution image (HR) from a single or a set of low-resolution (LR)
observations~\cite{Huang_SR},  which is termed as single image SR and multi-frame SR respectively.
SR has a long history and is a still very active field~\cite{Huang_SR}, with many new techniques emerging~\cite{Freeman_SR,Glasner_super-resolutionfrom,  NLKR_zhang, Yang_SR_TIP}.

Conventional multi-frame SR methods are reconstruction based  methods.
They typically perform motion estimation followed by frame fusion~\cite{Huang_SR}, which rely on the complementary information contained in different frames for resolution enhancement. The success of this kind of approaches relies heavily on the accuracy of the motion estimation procedure as well as the later frame fusion scheme. Furthermore, such reconstruction based methods are inherently limited to the case of small zooming factors~\cite{Limits_SR}.
For single image SR, the simplest methods are the interpolation-based approaches, \emph{e.g.}, bicubic interpolation. However, such analytical interpolation based methods do not exploit the  underlying structures in natural images, such as edges, thus usually blurring the fine details and  introducing artifacts in the interpolated results. More advanced interpolation-based methods take the underlying structures of the image into consideration during interpolation, \emph{e.g.},~\cite{Li01newedge-directed,Takeda07kernelregression}, thus improving the interpolation quality by adjusting the interpolation scheme according to the latent structures.
Recently, methods exploiting the natural image priors for SR have been proposed in the literature~\cite{Freeman_SR, Yang_SR_TIP,  Tappen03exploitingthe, Kim_SR_PAMI, NBSR_Zhang}.
Freeman \emph{et al.} exploited the patch similarity prior with the help of a large training image set for SR~\cite{Freeman_SR}.
Tappen \emph{et al.} developed a Markov Random Field based  SR algorithm utilizing the sparse derivative prior of natural images~\cite{Tappen03exploitingthe}.
Yang \emph{et al.} proposed a SR approach based on the sparse representation prior of image patches with respect to a properly chosen dictionary~\cite{Yang_SR_TIP}.
Kim \emph{et al.} developed a regression-based SR algorithm based on kernel regression and with a sparsity prior of natural images~\cite{Kim_SR_PAMI}.
Apart from the sparsity prior, SR methods exploiting other properties of natural images such as self-similarities and  local/non-local regularities have also been proposed in the literature~\cite{Glasner_super-resolutionfrom, NLKR_zhang}.
{\rv Another property of natural images is the  edge statistics,} which  have been exploited in \cite{Fattal_Edge}, \cite{Softcuts} and \cite{Gradient_SR} as a prior for SR.
While exploiting the natural image statistical properties as a prior, these methods typically formulate the problem within the Bayesian framework and then  estimate the final HR image with a Maximum A Posteriori (MAP)  estimation approach.\footnote{Some of the approaches may be formulated as a regularized regression problem, the solution to which is related to the MAP solution to the corresponding Bayesian formulation. See Section~\ref{sec:SR_SR} for more details.} Although it is demonstrated effective in some applications, this kind of approach discards the further potential enabled by the Bayesian probabilistic modeling.
Another limitation is that the regularization term is typically   designed empirically \emph{a priori}, thus is not guaranteed to be compatible with the image statistics.
There are indeed some Bayesian SR approaches in the literature that do not use MAP solution~\cite{Post_BSR, VBSR, BSR_GMRF, Bayes_SR}; however, they  typically use very simple image priors to ensure the tractability of computation, thus limiting their performance.
A SR approach using  posterior based sampling is proposed in \cite{Post_BSR}. This method is  limited by the fact that it uses a very simple Ising model as its prior~\cite{Post_BSR}.
Multi-frame SR methods based on Variational Bayes (VB) technique have been developed in the past~\cite{VBSR, BSR_GMRF,Bayes_SR}.
In~\cite{VBSR}, the authors used the Total Variation (TV) model as the prior, which prefers piecewise linear images thus is not quite adequate for general natural images.
In~\cite{BSR_GMRF}, the authors used a casual Gaussian MRF model as the prior, which includes  latent variables indicating the  dependency of two adjacent pixels.
Although the model has taken the edge structure of natural images into consideration, it is not flexible enough to capture the statistics of natural images well.
In~\cite{NBSR_Zhang}, a Bayesian image SR method with high-order MRF modelling of natural images is proposed and the HR image is estimated via MCMC sampling.
To sum up, the conventional SR approaches using regularized regression are limited by the fact that they do not exploit the full potential offered by the probabilistic modeling, although various kinds of regularization terms have been designed; on the other hand,  for the previously developed Bayesian SR approach, the prior  is typically limited to very simple models to ensure the tractability in computation, thus limiting their estimation quality. Furthermore, VB technique is typically used for the posterior mean estimation in those approaches, which actually compromises the modeling accuracy, while the sampling based approach is time consuming with high computational cost.

We propose in this paper a single image super-resolution algorithm using sparse bayesian modeling of natural images, which enjoys several advantages:
\begin{enumerate}
\item the proposed method is derived in a Bayesian framework, which can incorporate the \emph{a priori} knowledge on the latent HR image as well as other uncertainties such as the noise level into the framework  in a natural way;
\item  a  high-order MRF model is used for   for capturing the HR image statistics. Different from~\cite{NBSR_Zhang} which uses a Gaussian Scale Mixtures (GSM) with finite discrete scales, our model uses an infinite number of components in the GSM model, with the scales  adaptively learned along with the process of HR image estimation, rather than chosen from a pre-specified finite set of discrete scales as in~\cite{NBSR_Zhang}.
Also, as a global statistical prior is used for the HR image, our algorithm does not require to re-train the model when the zooming factor is changed as required by the sparse representation based SR method~\cite{Yang_SR_TIP}.
Therefore, the proposed method is more flexible.
\item  an empirical Bayesian method is applied for the task of HR image estimation as well as the estimation of the hyper-parameters,
which does not require ad-hoc modifications (tuning the trade-off parameters) as  the MAP approach  to achieve desirable restoration performance. Also, it is less sensitive to the local minima in the solution space than MAP, especially when heavy-tailed priors are applied. The proposed method is much more efficient than the method based on Markov Chain Monte Carlo (MCMC)  sampling~\cite{NBSR_Zhang} and can generate results comparable to or better than that method.
\end{enumerate}

The rest of this paper is organized as follows. We first  give a  review  of some related previous works briefly in Section~\ref{sec:SR_SR}, including the general restoration framework as well as the previous approaches for modeling natural images.
 We introduce our SR framework with Sparse Bayesian modeling of natural images in Section~\ref{sec:EB_SR}.
Then we present an efficient and effective super-resolution algorithm in Section~\ref{sec:EB_SR_Algo}.
Experiments are conducted in Section~\ref{sec:Exp} and the results are compared with several \emph{state-of-the-art} algorithms.
We conclude this paper in Section~\ref{sec:con}.


\section{Previous Works}\label{sec:SR_SR}
We review some of the related previous works in this section, which will lay the foundation for the derivation of our approach later.

\subsection{Image Restoration Model}
Mathematically, the LR image formation process is usually modeled as the convolution of the latent HR image with a low-pass filter which models the low passing property of the imaging system followed with downsampling,
\begin{equation}
  \y = \D\HH\x+ \boldsymbol{\epsilon},
\label{eq:observation_model}
\end{equation}
where $\x\in \mathbb{R}^{m}$ and $\y\in \mathbb{R}^n$ is the vector representation of the HR and LR image, respectively, obtained by lexicographical ordering the images.   $\HH \in \mathbb{R}^{m\times m}$ is the matrix corresponding to the blurring process with blur kernel $\boldsymbol{h}$ and $\D \in \mathbb{R}^{n\times m}$ $(n<m)$ is the matrix representing the downsampling operator. $\boldsymbol{\epsilon} \in \mathbb{R}^n$ is a noise term.
The task of  SR is to estimate the underlying HR image $\x$ given only the LR observation $\y$.
In the sequel, we  review briefly some related works in the literature.

As the task of SR is to estimate a high resolution image from the low resolution observation, the number of unknowns $m$ is larger than that of the knowns $n$, thus the problem is ill-posed. Therefore, regularization techniques based on  the \emph{a priori} knowledge are required to alleviate the ill-posedness of this problem.
Typically, the task of SR is formulated as a regularized least square regression problem. Mathematically,
\begin{eqnarray}\label{eq:reg}
\hat{\x} = \arg \min_{\x} \Vert \y - \D\HH\x\Vert_2^2 + \lambda R(\x),
\end{eqnarray}
where the first term is the fidelity term reflecting the fact that the unknown HR image $\x$ should generate observation that is similar to $\y$ after passing it through the observation system.
The second term, $R(\cdot)$,  is a regularization term on the desired HR image to alleviate the ill-posedness of the SR problem. $\lambda$ is a parameter balancing the contributions of these two terms.
This is a very popular model and has been adopted in many recent SR and other related works~\cite{ Softcuts, Gradient_SR,  Shan_SR, Detial_Syn}.
Following this direction, most of the current works focus on designing different formulations for the regularization term $R(\cdot)$.
The most classical choice for $R(\cdot)$ would be the squared $\mathcal{\ell}_2$-norm, which is based on the assumption that the energy of the HR image should not be too large.  Because of its well known blurring effect for the  $\mathcal{\ell}_2$-norm based regularization, some algorithms apply the $\mathcal{\ell}_2$-norm in the gradient domain.
Recently, many approaches exploiting  the sparsity property of natural images  are developed~\cite{Yang_SR_TIP, Tappen03exploitingthe, Kim_SR_PAMI, Softcuts, Gradient_SR,  Detial_Syn}.
A general approach for designing $R(\cdot)$ utilizing the sparsity property is to use the sparsity property in the transform domain (\emph{e.g.}, wavelet domain, edge domain) as regularization~\cite{ Softcuts, Gradient_SR}.
A novel regularization term called `Softcuts' has been developed in \cite{Softcuts}, by  exploiting the sparse property of the edge profiles.
In \cite{Gradient_SR}, the authors use the sparsity prior of the edges in images by first estimating the gradient of the HR image via transforming that of the LR image with the learned relationship between the LR and HR gradients. Then the regularization term $R(\cdot)$ is designed as the distance between the gradients of the unknown latent HR image and the estimated HR gradients.
A sparse  representation (coding) based SR (ScSR) method has been proposed in \cite{Yang_SR_TIP} based on the property that natural image patches have a sparse representation with respect to a properly chosen dictionary. Specifically, the ScSR method first recovers the sparse representation coefficients from the low resolution patches with respect to a low resolution dictionary, and then reconstructs a high resolution patch with the recovered representation coefficients and a high resolution dictionary, which is trained jointly with the low resolution dictionary.
To further enhance the texture details in the estimated HR images,  several approaches have been proposed by using texture prior offered by domain specific examples~\cite{ Detial_Syn, Sampled_Texture_Prior, Fattal_Texture}.

The regularized SR model (\ref{eq:reg}) is related to a probabilistic model as follows:
\begin{eqnarray}\label{eq:reg_Bayesian}
\begin{split}
p(\x|\y) &= \frac{p(\y|\x)p(\x)}{p(\y)} \propto p(\y|\x)p(\x)\\
 &\propto \exp\left(-\frac{\Vert \y - \D\HH\x\Vert_2^2}{\sigma^2}\right) \exp\left(-\frac{R(\x)}{\eta^2}\right),
 \end{split}
\end{eqnarray}
where $\sigma$ denotes the standard deviation of the noise and $\eta$ is a scale parameter for the prior distribution of $\x$.
At this point, it is easy to see that the solution to (\ref{eq:reg})  corresponds to the Maximum-A-Posteriori (MAP) solution to (\ref{eq:reg_Bayesian}) due to the following relation
\begin{eqnarray}
\begin{split}
\hat{\x} &=    \arg \max_{\x} p(\x|\y) \Leftrightarrow  \arg \min_{\x} -\log p(\x|\y),
\end{split}
\end{eqnarray}
which  corresponds to the mode of the posterior distribution for the HR image $\x$. The regularization parameter $\lambda$ in (\ref{eq:reg}) is related to $\sigma$ and $\eta$ in (\ref{eq:reg_Bayesian})  by $\lambda = \frac{\sigma^2}{\eta^2}$.

 Because of this relation,  many previous works first model the SR problem in a probabilistic framework as (\ref{eq:reg_Bayesian}) and then
  estimate the HR image  by minimizing the negative logarithm of it, which actually boils down to the regularized regression model as shown in (\ref{eq:reg})~\cite{Gradient_SR, Detial_Syn}.
To solve (\ref{eq:reg}), gradient based optimization schemes are typically used~\cite{Gradient_SR, Detial_Syn}.
While some encouraging results can be obtained, the MAP approach could not exploit the full potential offered by  the probabilistic modeling, as only the posterior mode is sought for solution, discarding much information offered by the probabilistic model. Even worse, it is prone to getting stuck in a local minima  when the prior model is a heavy-tailed distribution, which actually leads to a non-convex optimization problem, thus there is no guarantee of obtaining even the posterior mode.

\subsection{Prior Models based on Natural Image Statistics}\label{sec:prior}
The statistics of natural images have received great attention of researchers from different communities
for both understanding the human visual system and designing more effective processing algorithms~\cite{Field}.
Natural image statistics modeling has a fundamental relation with the field of image processing.
Natural images are different from random noise image in that they exhibit structures. Examples are local regularities such as edges~\cite{Takeda07kernelregression} and  self-similarities~\cite{Glasner_super-resolutionfrom, NLKR_zhang}. As a result, the natural images only occupy a tiny fraction of the whole space of all the images generated by all the possibilities enables by the pixel value combinations. Natural images have strong statistical properties known as natural image statistics~\cite{Internal_Statistics, good_model, FOE}.
One of the most well-known properties is that the natural images exhibit heavy-tailed distribution when applying derivative filters  onto them. This can be explained intuitively as follows: on one hand, natural images are locally smooth, therefore, the local difference will be small, thus the marginal distribution will decrease faster than the Gaussian distribution; on the other hand, natural images have many structures such as edges, where the derivative response can be large, which contributes to the heavier tails than the Gaussian distribution. This prior enjoys wide range of applications, including image denoising~\cite{good_model,FOE}, deblurring~\cite{Fergus06removingcamera} and super-resolution~\cite{NBSR_Zhang, Kim_SR_PAMI}.
Apart from this heavy-tailed marginal statistics, natural images also exhibit several other properties, such as scale invariance, which states that the natural images exhibit similar heavy-tailed distribution at different scales~\cite{Internal_Statistics}. Another property is the joint statistics, meaning that the neighboring pixels in the natural images exhibits high dependency~\cite{FOE}.

There are many works on image priors and it is still a very active research topic.  Gaussian model applied onto the derivatives of images is the most classical and widely used one due to its simplicity:
\begin{equation}\label{eq:_Prior_Tik}
p(\x)\propto \exp\left(-\frac{ \Vert \nabla \x\Vert_2^2}{\eta^2}\right),
\end{equation}
where $\nabla \x$ denotes the gradients of image $\x$.
Gaussian prior is related to the Tikhonov regularization in the inverse problem theory.
Although it has the advantage of closed-form solution, it can not generate satisfying solutions in most of the image restoration cases as it typically smooths the image too much.
To preserve the edge structure, Laplacian prior is later used as image prior and is defined as:
\begin{equation}\label{eq:Prior_TV}
p(\x)\propto \exp\left(-\frac{ \Vert \nabla \x \Vert_1}{\eta^2}\right).
 \end{equation}
Laplacian prior is related to the Total Variation (TV) model derived in the Partial Differential Equation field~\cite{Osher_TV}, which has been proved to keep the
image discontinuities better. It is also related to the $\mathcal{\ell}_1$-norm based regularization, which is well known for its ability in promoting sparsity for the solution.
Although the TV prior can preserve the edges, it can not capture natural image statistics well enough, as the resulting images are typically piecewise linear, which is desirable for cartoon-like images, but not for natural images. Natural images follow a distribution with heavier tails than Gaussian and Laplacian distribution,  which has been realized and used by many researchers recently~\cite{Fergus06removingcamera, nips09_fast_deblur_hyperLap}.
To over come this limitation, some researchers proposed to use the hyper-Laplacian  sparse prior for natural images which has been  successfully applied in many fields~\cite{Fergus06removingcamera, nips09_fast_deblur_hyperLap}:
\begin{equation}\label{eq:Prior_Hyper_Lap}
p(\x)\propto \exp\left(-\frac{ \Vert \nabla \x \Vert_{\alpha}}{\eta^2}\right),
 \end{equation}
where $\Vert \x \Vert_{\alpha}$ is defined as $\Vert \x \Vert_{\alpha} = \sum_i \vert\x(i) \vert^{\alpha}$.
  $\alpha$ here is the parameter controlling the sparseness of the desired natural images. For natural images, the responses  with respect to the learned sparse filters are sparse, \emph{i.e.}, with a large number of zero elements and a few responses with large absolute values; therefore, $\alpha$ is typically chosen as $0.5\le\alpha\le 0.8$ in the literature~\cite{nips09_fast_deblur_hyperLap}.


%

An elegant formulation for unifying the above mentioned   probabilistic models of natural images  is  Markov Random Field (MRF)~\cite{FOE}.
MRF is a type of undirected graph, where the node and edge of this graph is defined differently according to different MRF configurations.
In MRF, the probability of a whole image $\x$ is defined as follows based on the computation of local cliques:
\begin{eqnarray}\label{eq:prior}
\begin{split}
p(\x) = \frac{1}{Z} \prod_{c\in \zeta} f(\x_{[c]}),
\end{split}
\end{eqnarray}
where $\zeta$ denotes the set of all pixel locations of the image and $c$ refers to a particular location.
{\rv  $\x_{[c]}$ is the vector made up by the local neighborhood  at $c$ for the image $\x$, which is also referred to as a clique.}
$f(\cdot)$ is a potential function, which takes a clique as input and outputs the potential or energy for that clique.
$Z$ is the partition function to ensure  $p(\x)$ to be integrated to one.
One typical  MRF model is the so called pairwise MRF. In this model, each pixel in the image is defined as a node.
Each pixel is connected with its directed neighbors.

The choice of the potential function $f(\cdot)$ is  crucial for the modeling quality.
The conventional choice is the Gaussian function, as in (\ref{eq:_Prior_Tik}).
With the recent development in natural image statistics, it is demonstrated that the potential function defined as a heavy-tailed function over the response of derivative filters on the clique $\x_c$ can better capture the  image statistics~\cite{FOE}.
Motivated by the property of heavy-tailed marginal distribution of natural images, to improve the modeling quality, researchers suggest to use heavy-tailed potential functions such as Laplacian and hyper-Laplacian as in (\ref{eq:Prior_TV}) and (\ref{eq:Prior_Hyper_Lap}) and Student-t distribution  as used in \cite{FOE}.

In the terminology of MRF, the models (\ref{eq:_Prior_Tik})$\sim$(\ref{eq:Prior_Hyper_Lap}) are pairwise MRF, as only two neighboring pixels are involved in each clique, due to the usage of the first order derivative filters.
However,  the modeling power of pairwise MRF is rather limited, as the pixels in natural images also exhibit long range relations; therefore, high-order MRF model is developed in some recent work~\cite{FOE}.
One challenge in the case of high-order MRF  is that the proper potential function is hard to define in the high dimensional space, as the size of a clique  has been increased.
The recently introduced Field-of-Experts (FoE) model formulates the potential on clique $c$ as a product of several functions defined on different low dimensional spaces respectively~\cite{FOE}:
\begin{eqnarray}\label{eq:expert}
\begin{split}
f(\x_{[c]}) & = \prod_{l=1}^L \phi\Big((\k_l\ast \x )_c; {\Theta}_l\Big)\\
&= \prod_{l=1}^L \phi\left((\K_l \x)_{c}; {\Theta}_l\right),
\end{split}
\end{eqnarray}
which is known as the Product-of-Expert (PoE) model in the literature~\cite{PoE}.
{\rv $(\k_l\ast \x )_c$ refers to the $c$-th pixel in the filtered image by $\k_l$.}
$\mathcal{K}=\{\k_l\}_{l=1}^L$  is a set of derivative filters, either designed under some criteria or learned from the data~\cite{good_model, FOE}.
$\K_l \in \mathbb{R}^{m\times m}$ is the matrix representation corresponding to the derivative filter $\k_l$, with the following relation: $\K_l\x = \k_l\ast\x$.
The clique $\x_{[c]}$  in this model is actually the local patch covered by the  filter $\k_l$ ($l\in\{1, \cdots, L\}$) at $c$.
The model in (\ref{eq:expert}) consists of  $L$ expert functions to model the high-dimensional distribution by using the products of them.
$\{\Theta_l\}_{l=1}^L$ is the parameter set associated with the PoE.

In FoE, the filters are learned from the training data~~\cite{FOE}.
The advantage of learned filters over pre-specified ones (such as $\{\k_1=[-1, 1]$, $\k_2 = \k_1^T$\}) is that the learned filters have the potential to  capture the statistics of natural images better than the simple horizontal and vertical derivative filters.
 Note that the prior can also be made adaptive to the content of the specific image by allowing its parameters to change according to the content~\cite{Content_Aware}.
One challenge in learning the MRF model is that the normalization term $Z$ is  typically intractable to evaluate, which poses a great difficulty for the learning process. To handle this challenge, several schemes such as {\rv Contrastive Divergence}~\cite{PoE} and Score Matching~\cite{Score_Matching} are proposed in the literature.

\section{Image Super Resolution via Sparse Bayesian Modeling}
\label{sec:EB_SR}
\begin{figure}[ht]
\centering
\begin{overpic}[viewport = 50 50 550 300, clip, width=7cm]{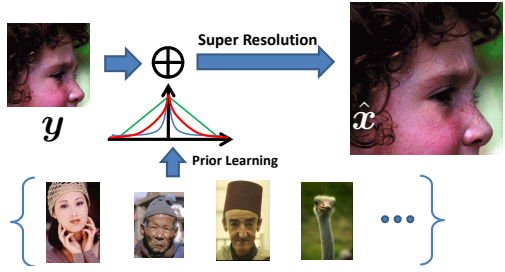}
\put(9,25){\Large\textcolor{black}{$\y$}}
\put(68,26){\Large\textcolor{white}{$\hat{\x}$}}
\end{overpic}
\caption{{\bf Bayesian SR Framework}. In the proposed SR framework, we first learn a set of  filters from the training images with a flexible high-order MRF model, with Gaussian Scale Mixture model with continuous scale as its potentials. The learned filter is then used together with the potential model as the prior for the underlying HR image to be estimated.
The proper scale is adapted to the image and is estimated together with the HR image estimation.
The HR image is estimated via an empirical Bayesian approach.
In this framework, we are actually `transfer' the prior learned from training images to the unknown HR image.}
\label{fig:framework}
\end{figure}

As mentioned above, the conventional SR approaches using regularized regression can not exploit the full potential of the probabilistic model and are also  often faced with non-convex optimization problem when  heavy-tailed sparse priors are used, thus is prone to be trapped in a local minima; on the other hand, previously proposed Bayesian SR approach\footnote{By Bayesian SR, we refer to those approaches that use the MMSE estimation criteria, rather than those using the MAP criteria, even though they are formulated in a Bayesian framework.} only used very simple prior model for natural images due to the limitation in computation~\cite{VBSR, BSR_GMRF, Bayes_SR}, or resorting to MCMC sampling when a more advanced prior is used in the Bayesian model~\cite{NBSR_Zhang}.
In this section, we present an effective and efficient Bayesian SR approach which can use advanced prior model learned from natural images as~\cite{NBSR_Zhang}, while reducing the computational overhead significantly.
The framework of the proposed approach is illustrated in Figure~\ref{fig:framework}.
In the sequel, we first formulate  the SR problem in the Bayesian framework and then present an effective algorithm for generating the SR image.
In terms of Bayesian inference, the latent SR image is estimated via the posterior distribution  $p(\x|\y; \Theta)$. Using the Bayes rule, we can get:
\begin{eqnarray}\label{eq:Bayes}
p(\x|\y; \Theta) =\frac{ p(\y|\x; \Theta) p(\x)}{p(\y)},
\end{eqnarray}
therefore, the task for the Bayes modeling includes the configuration of the  likelihood $p(\y|\x; \Theta)$  and the prior $ p(\x)$, which is presented in detail in the sequel.

\subsection{Bayesian Image Super Resolution Formulation}
In this subsection, we first present the likelihood and prior model, and then derive the posterior for the latent HR image, from which the posterior mean is computed as the  estimation of the HR image.
\subsubsection{Likelihood}
The noise term $\boldsymbol{\epsilon}$ is assumed to be known only up to the statistical property, {\emph{i.e.}}. its statistical distribution.
The statistical distribution of the noise is application dependent.
In many situations, the noise term can be modeled with Gaussian distribution.
In this work, we also use a Gaussian  distribution  for modeling the statistics of noise.
Assuming the elements in the noise term $\boldsymbol{\epsilon}$  follow \emph{i.i.d.} Gaussian distribution and the standard deviation  is $\sigma$, then we have:
\begin{eqnarray}
\label{eq:noise}
\boldsymbol{\epsilon}  \sim \mathcal{N}(\mathbf{0}, \sigma^2 \II).
\end{eqnarray}
With  (\ref{eq:noise}) and the observation  model  defined in (\ref{eq:observation_model}),
  we can derive the following conditional distribution for the observed image:
\begin{eqnarray}\label{eq:likelihood}
\begin{split}
p(\y|\x, \sigma^2) = \mathcal{N}(\y | \D\HH\x, \sigma^2 \II),
\end{split}
\end{eqnarray}
where $\II\in \mathbb{R}^{n\times n}$ is the identity matrix.
The noise level $\sigma^2$ is not assumed to be known. Rather, we treat it as a random variable and place a hyperprior on it as shown later.

\subsubsection{Prior Model for Image}
The definition of the prior model for the HR image is crucial for the final HR estimation quality.
The design of such a prior should reflect the desired property of natural HR images.
The Gaussian prior reflects the smoothness property of natural images.
However, this prior typically causes blurring effects for the structures in natural images, as it fails in capturing the heavy-tailed distribution of the natural image statistics.
Also, the conventional first-order MRF model for natural images are limited by the  pairwise interaction, as it is not flexible enough for capturing the image statistics. The recently developed FoE model is a high-order MRF model that can capture the statistics of natural images better by learning the high-order filters from the data~\cite{FOE}.
Therefore, in this work, we use a high-order FoE model as the prior for the underlying HR image $\x$:
\begin{eqnarray}\label{eq:x_prior}
\begin{split}
p(\x) = \frac{1}{Z} \prod_{c\in \zeta} f(\x_{[c]})= \frac{1}{Z} \prod_{c\in \zeta} \left( \prod_{l=1}^L \phi\big( (\k_l \ast \x)_{c}; {\Theta}_l\big)\right),
\end{split}
\end{eqnarray}
where Gaussian  scale mixtures (GSM) model (\ref{eq:GSM_expert}) is used as the expert function $\phi(\cdot)$ rather than Student-t distribution as used in~\cite{FOE}, and the Gaussian, Laplacian and hyper-Laplacian functions used in (\ref{eq:_Prior_Tik}), (\ref{eq:Prior_TV}) and  (\ref{eq:Prior_Hyper_Lap}) respectively.
This model is very flexible and can be adapted easily for capturing the natural image statistics.
In our previous work~\cite{NBSR_Zhang}, we have used \emph{discrete} and \emph{finite} Gaussian Scale Mixture (GSM) as the expert function:
\begin{eqnarray}\label{eq:GSM_expert}
\phi\big((\k_l\ast\x)_c; \Theta_l\big) = \sum_{j=1}^J \beta_{lj} \mathcal{N} \big((\k_l\ast\x)_c; 0, \frac{\eta^2_l}{s_j}\big),
\end{eqnarray}
where $\beta_{lj}$
is the normalized weight for the $j$-th component in the Gaussian mixture for modeling the  responds of the \mbox{$l$-th} filter. $\eta^2_l$ is the base variance for the $J$ Gaussian components in the $l$-th expert function.
 $s_j$ is the scale parameter for the $j$-th component.
The usage of this expert function allows efficient sampling as used in~\cite{NBSR_Zhang}. However, in this model, the scales are pre-defined and restricted to be one of the values in the discrete set.

Now we resort to variational approach for modeling the expert function $\phi(\cdot)$ as:
\begin{eqnarray}\label{eq:GSM_expert}
\phi\big((\k_l\ast\x)_c | \gamma_{cl} \big) = \max_{\gamma_{cl} \ge 0}  \mathcal{N} \big((\k_l\ast\x)_c| 0, \gamma_{cl} \big) \psi(\gamma_{cl}),
\end{eqnarray}
which is actually the superposition of zero-mean-Gaussian with different variances.

Using (\ref{eq:x_prior}) and (\ref{eq:GSM_expert}), we can have the following prior model for the latent HR image $\x$ as:
\begin{eqnarray}\label{eq_prior_x}
\begin{split}
p(\x|\{\gam_l\}_{l=1}^L, \tau) = \mathcal{N} (\x| \mathbf{0}, \V^{-1}_{\gamma}),
\end{split}
\end{eqnarray}
where $\V^{-1}_{\gam} = \sum_{l=1}^L \K^{\top} \GM^{-1}_l \K_l$ and $\GM_l = {\rm diag} (\gam_l)$.

\subsubsection{Hyperpriors on Hyperparameters}
We do not assume the noise level to be known in our model. Rather, we treat the noise level as a random variable and perform estimation for it jointly with other random variables.
Specifically, to model this random variable, we use a Gamma distribution as its hyperprior:
\begin{eqnarray}\label{eq:noise_prior}
p(\tau) = \mathcal{G}(\tau;a^o,b^o) =  \frac{1}{\Gamma(a^o)} (b^o)^{a^o} \tau^{a^o-1} e^{-b^o\tau} ,
\end{eqnarray}
where $\tau = \sigma^{-2}$ which is also referred to as precision. $a^o$ and $b^o$ are non-negative parameters controlling  the shape  and scale of the Gamma distribution, respectively.
For the latent variables $\{\gam_l\}_{l=1}^L$, we also place gamma priors over them, which will promote sparsity combined with the prior for $\x$x:
\begin{eqnarray}\label{eq:gam_l}
\begin{split}
p(\{\gam_l\}_{l=1}^L) = \prod_{l=1}^L \prod_{c=1}^N \mathcal{G} (\gam_{cl} ; a, b),
\end{split}
\end{eqnarray}
where $a$ and $b$ are the hyper-parameters for the hyper-priors.

\subsubsection{Posterior}
By substituting  (\ref{eq:likelihood}), (\ref{eq_prior_x}),  (\ref{eq:noise_prior}) and (\ref{eq:gam_l})  into (\ref{eq:Bayes}), we can derive the following posterior for the latent HR image $\x$ as well as the precision $\tau$:
\begin{eqnarray}\label{eq:posterior}
\begin{split}
&p(\x, \{\gam_l\}_{l=1}^L, \tau|\y)  \\
&\propto p(\y|\x, \{\gam_l\}_{l=1}^L, \tau) p(\x|\{\gam_l\}_{l=1}^L, \tau)  p(\{\gam_l\}_{l=1}^L) p(\tau)\\
&\propto \mathcal{N}(\y| \D\HH\x, \tau^{-1} \II) \cdot \mathcal{N} (\x| \mathbf{0}, \V^{-1}_{\gamma}) \cdot \prod_{l=1}^L \prod_{c=1}^N \mathcal{G} (\gam_{cl}; a, b)  \cdot \mathcal{G}(\tau;a^o,b^o).
\end{split}
\end{eqnarray}

\section{Empirical Bayesian Super-Resolution Algorithm}
\label{sec:EB_SR_Algo}

\subsection{Empirical Bayesian SR Algorithm}
After obtaining the posterior (\ref{eq:posterior}), we should perform inference based on it.
However, there are several challenges for this.
One of the difficulties is the fact that (\ref{eq:posterior}) is typically intractable as we can not perform the normalization of it.
Therefore, approximate schemes have to be adopted.
One type of approach is to approximate the posterior distribution with separable distribution under criteria such as KL-divergence, thus transforming the
problem to another easy-to-solve problem. A representative example is the Variational Bayes (VB) method~\cite{EL}, which has  been applied to image deblurring and compressive sensing recently~\cite{Fergus06removingcamera, VB_CS}.
While approaches based on the VB technique are computationally efficient as it typically offers analytic formulas for updating the variables, it compromises the modeling accuracy for computational efficiency.
Apart from approximating the original model with tractable ones as in VB, another line of solution is based on generative sampling from the posterior distribution. After  obtaining  enough samples from the posterior distribution,  the  empirical mean calculated from the generated samples is used  as the estimation for the posterior mean. This approach has been used in our previous work on image SR~\cite{NBSR_Zhang}, which is, however,
time consuming and not practical for real applications.

Rather than using VB approximation or sampling,  Tipping suggested instead to decompose the posterior into two parts as follows~\cite{Tipping_SBL_RVM_JMLR01}:
\begin{eqnarray}\label{eq:posterior_decomp}
\begin{split}
p(\x, \{\gam_l\}_{l=1}^L, \tau|\y) =  p(\x|\y, \{\gam_l\}_{l=1}^L, \tau) p(\{\gam_l\}_{l=1}^L, \tau|\y).
\end{split}
\end{eqnarray}
The reason for this decomposition is that while the full posterior is not analytically computable due to the normalization factor, the
first part of (\ref{eq:posterior_decomp})  enables this analytical calculation.
The empirical Bayesian approach first estimates the parameters and latent variables via maximizing the marginal likelihood by integrating out the unknown variable $\x$, and then uses the estimated optimal parameters for evaluating the conditional posterior for  $\x$ using the first part of (\ref{eq:posterior_decomp})~\cite{Tipping_SBL_RVM_JMLR01, Wipf_Latent_Variable_TIT11}.
This approach is first applied in Relevance Vector Machine (RVM) as an alternative for Support Vector Machine (SVM) and also used in  Automatic Relevance Determination (ARD)~\cite{Tipping_SBL_RVM_JMLR01}.
 automatic relevance determination
This approach has been introduced into single processing community for basis selection and sparse representation by Wipf \emph{et al.} in \cite{Wipf04sparsebayesian},
which has later been extended further in~\cite{WipfR07_SSA, Wipf_Structured_Sparse_NIPS11, Wipf_Latent_Variable_TIT11, Wipf_L1L2}, all showing superior performance in various signal processing applications.
Some theoretical analysis on the optimality conditions has been carried out in \cite{Wipf_Latent_Variable_TIT11,Wipf_L1L2}.
In this work, we further exploit the possibilities of this promising paradigm for image processing tasks.
The extension, however, is non-trivial, due to the high-dimensional property inherent to image processing tasks.

Specifically, decomposing (\ref{eq:posterior}) into two parts according to (\ref{eq:posterior_decomp}), we have the first part as follows:
\begin{eqnarray}\label{eq:marg_post}
\begin{split}
p(\x|\y, \{\gam_l\}_{l=1}^L, \tau) & = \frac{p(\y| \{\gam_l\}_{l=1}^L, \tau) p(\x| \{\gam_l\}_{l=1}^L, \tau)}{ p(\y| \{\gam_l\}_{l=1}^L, \tau)}\\
& = \mathcal{N}(\x| \boldsymbol{\mu}_{\x}, \boldsymbol{\Sigma}_{\x}).
\end{split}
\end{eqnarray}
where the posterior mean and convariance are:
\begin{eqnarray}\label{eq:x1}
\boldsymbol{\mu}_{\x} = \tau \boldsymbol{\Sigma}_{\x} \HH^{\top} \D^{\top} \y.
\end{eqnarray}
\begin{eqnarray}\label{eq:Sigma}
\begin{split}
\boldsymbol{\Sigma}_{\x} & = \W_{\gamma}^{-1} = \left(\tau  \HH^{\top}\D^{\top} \D\HH + \sum_{l=1}^L \K_l^{\top} \GM^{-1}_l \K_l\right)^{-1}\\
&  = \left( \left[\HH^{\top}\D^{\top}\; \K_1^{\top} \cdots \K_L^{\top} \right]     \left[
                                          \begin{array}{cccc}
                                            \tau\II &  & \cdots & \bf{0} \\
                                             &\GM^{-1}_1 &  &\vdots \\
                                            \vdots &  & \ddots &  \\
                                            \boldsymbol{0} & \cdots &  & \GM^{-1}_L \\
                                          \end{array}
                                        \right]
                                        \left[
                                          \begin{array}{c}
                                            \D\HH \\
                                            \K_1 \\
                                            \vdots \\
                                             \\
                                            \K_L \\
                                          \end{array}
                                        \right]
\right)^{-1} = \left(\K^{\top} \Z \K\right)^{-1},
\end{split}
\end{eqnarray}
where $\GM_l = {\rm diag} (\boldsymbol{\gamma}_l)$.
Basically, the first part guarantees that the posterior mean estimation of the HR image can be obtained via (\ref{eq:x1}) once the hyper-parameters ($\{\gam_l\}_{l=1}^L, \tau$) are known.

The estimation of these hyper-parameters are obtained by minimizing the negative log of the marginal likelihood with the unknown latent HR image $\x$  integrated out is as follows~\cite{Tipping_SBL_RVM_JMLR01, Wipf_Latent_Variable_TIT11, Wipf_L1L2, Wipf_Dualspace_Sparse_NIPS12}:
\begin{eqnarray}
\begin{split}
\{\hat{\gam_l}\}_{l=1}^L, \hat{\tau}x & = \arg \min_{\{\gam_l\}_{l=1}^L, \tau} \mathcal{L}( \{\gam_l\}_{l=1}^L, \tau),
\end{split}
\end{eqnarray}
where $\mathcal{L}( \{\gam_l\}_{l=1}^L, \tau)$  is the  negative log of the marginal likelihood defined as
\begin{eqnarray}
\begin{split}
\mathcal{L}( \{\gam_l\}_{l=1}^L, \tau) & = -\log \int p(\y|\x, \{\gam_l\}_{l=1}^L, \tau) p(\x| \{\gam_l\}_{l=1}^L, \tau) d\x\\
& = -\log p(\y; \{\gam_l\}_{l=1}^L, \tau)\\
& = \y^{\top} \Sigma_{\y}^{-1} \y + \log \vert \Sigma_{\y} \vert +\sum_{l=1}^L f(\gam_l) + \sum_{i=1}^n [\log \tau + f(\tau)],
\end{split}
\end{eqnarray}
where $ \Sigma_{\y} = \tau^{-1} \I + \D\HH \V^{-1}_{\gam} \HH^{\top}\D^{\top}$.
This leads to the following updating rules for $\{\gam_l\}_{l=1}^L$ and $\tau$.
The optimal $\gam_{l}$ can be computed as~\cite{Tipping_SBL_RVM_JMLR01, Wipf_Latent_Variable_TIT11, Wipf_L1L2}:
\begin{eqnarray}\label{eq:gamma}
\begin{split}
\gam_{li} = \frac{ u_{li}^2 + z_{li} + 2b}{1 + 2a},
\end{split}
\end{eqnarray}
where $\u_l \triangleq \K_l\ast\boldsymbol{\mu}_{\x}$ and the optimal $z_{li}$ is given by:
\begin{eqnarray}\label{eq:update_z}
\begin{split}
z_{li} &= \frac{\partial \log |\tau  \K \HH^{\top} \D^{\top}  \D\HH \K^{\top} +  \GM^{-1}|}{ \partial \gamma_{li}}\\
& = [(\tau  \K_l \HH^{\top} \D^{\top}  \D\HH \K_l^{\top} +  \GM^{-1})^{-1}]_{ii}.
\end{split}
\end{eqnarray}
The optimal $\tau$ can be computed as~\cite{Wipf_Dualspace_Sparse_NIPS12}:
\begin{eqnarray}\label{eq:tau2}
\begin{split}
\tau = \frac{\frac{n}{2}+a^o}{\frac{1}{2}\Vert \y - \D\HH\x \Vert_2^2 + 2w + b^o}.
\end{split}
\end{eqnarray}
where
\begin{eqnarray}\label{eq:update_w}
\begin{split}
w &= \frac{\partial \log |\tau   \HH^{\top} \D^{\top}\D\HH +  \V^{-1}_{\gam}|}{ \partial \tau}\\
& =   \sum_{i=1}^n [\D\HH(\tau  \HH^{\top} \D^{\top}  \D\HH  +   \V^{-1}_{\gam})\HH^{\top} \D^{\top}]^{-1}_{ii}.
\end{split}
\end{eqnarray}

\subsection{Covariance Estimation}
The estimation of the covariance $\sig_{\x}$ is required for updating both $\gam_l$ and $\tau$, as shown in (\ref{eq:update_z}) and (\ref{eq:update_w}).
However, as shown in (\ref{eq:Sigma}), explicit calculation of $\sig_{\x}$ requires the inversion of a very large matrix, which is intractable typically and  is also the  case in our task.
In the past, different approaches have been proposed for estimating the covariance matrix approximately, including the  low-rank approximation method using Lanczos algorithm~\cite{Seeger08} and sampling based method~\cite{PaYu11b, Seeger12}, which are time-consuming.
In the sequel, we will present an  efficient approximate covariance estimation approach.
Specifically, we use a diagonal approximation of ${\sig_{\x}}$ as $\tilde{\sig_{\x}}$:
\begin{eqnarray}
\begin{split}
\tilde{\sig_{\x}} &= {\rm diag} \left \{  {\rm diag ({\sig_{\x}}) } \right\}.
\end{split}
\end{eqnarray}
Therefore, the key is to compute $ {\rm diag} ({\sig_{\x}})$.
To achieve this goal, we take advantage of the following two relations:
\begin{eqnarray}\label{eq:diag_approx_I}
\begin{split}
{\rm diag} (\A \A^{\top}) = [\Vert \a_{1.}\Vert_2^2, \Vert \a_{2.}\Vert^2_2, \cdots]^{\top},
\end{split}
\end{eqnarray}
where $\A^{\top} = [\a_{1.},   \a_{2.}, \cdots]$.

The second relation is  for $\A \x = \k\ast \x$ (neglecting the boundary condition), we have
\begin{eqnarray}\label{eq:diag_approx_II}
[\Vert \a_{1.}\Vert_2^2, \Vert \a_{2.}\Vert^2_2, \cdots]^{\top} = {\rm vec} (\Vert \k\Vert_2^2, \Vert \k \Vert_2^2 \cdots) = {\rm vec} ( \k^{.^2} \ast \boldsymbol{1}).
\end{eqnarray}

From (\ref{eq:diag_approx_I}) and (\ref{eq:diag_approx_II}), we have
\begin{eqnarray}
{\rm diag} (\A \A^{\top})  = {\rm vec} ( \k^{.2} \ast  \boldsymbol{1}),
\end{eqnarray}
where $\A$ is the convolution matrix corresponding to the PSF kernel $\k$ (\emph{i.e.}, $\A \x = \k\ast \x$).
A straight forward generalization leads to the following relation:
\begin{eqnarray}\label{eq:diag_apprxo_III}
{\rm diag} (\A {\rm diag}(\boldsymbol{w}) \A^{\top})  = {\rm vec} ({\k}^{.2} \ast \boldsymbol{w}),
\end{eqnarray}
and
\begin{eqnarray}\label{eq:diag_apprxo_III}
{\rm diag} (\A^{\top} {\rm diag}(\boldsymbol{w}) \A)  = {\rm vec} (\bar{\k}^{.2} \ast \boldsymbol{w}),
\end{eqnarray}
where $\bar{\k} =$ {\ttfamily fliplr}({\ttfamily flipud} ($\k$)).

From (\ref{eq:Sigma}) and(\ref{eq:diag_apprxo_III}), we have
\begin{eqnarray}\label{eq:diag_apprx_W}
{\rm diag} (\W_{\gamma}) =  \tau {\rm vec} (\bar{\h}\ast \D^{\top}\D {\h}\ast \boldsymbol{1})) + \sum_{l=1}^L {\rm vec} (\bar{\k}^{.2} \ast (\boldsymbol{\gamma}_l^{-1})).
\end{eqnarray}
Recall the relationship between $\sig$ and $\W_{\gamma}$  as shown in (\ref{eq:Sigma}), we have
\begin{eqnarray}\label{eq:diag_apprxo_IV}
  {\rm diag} ({\sig_{\x}}) =  {\rm diag} ({\tilde{\sig}_{{\x}}}) \approx 1./{\rm diag} (\W_{\gamma}).
\end{eqnarray}
With this covariance approximation,  we can now detail the computation of $\{\gam_l\}_{l=1}^L$ as follows:
\begin{eqnarray}\label{eq:update_gamma_I}
\begin{split}
\gam_l &=  {\rm diag} \langle \u_l \u_l^{\top} \rangle\\
       & = (\bmu_{\u_{l}})^{.2} + {\rm diag} (\sig_{\u_{l}})\\
         &=  (\K_{l} \bmu_{\x})^{.2} +   {\rm diag} \left\{ \K_l  ( \sig_{\x}) \K_l^T \right\}\\
         & \approx (\K_{l} \bmu_{\x})^{.2} + {\rm diag} \left\{ \K_l (\tilde{\sig_{\x}}) \K_l^T \right\}\\
\end{split}
\end{eqnarray}
where $\x^{.2}$ denotes the element-wise square of the vector $\x$.
Plugging (\ref{eq:diag_apprxo_IV}) into (\ref{eq:update_gamma_I}), we have the  update equation for the latent variables $\{\gam_l\}_{l=1}^L$:
\begin{eqnarray}\label{eq:update_gamma_II}
\begin{split}
\gam_l &=  {\rm diag} \langle \z_l \z^T \rangle\\
       & \approx (\K_{l} \bmu_{\x})^{.2} + {\rm diag} \left\{ \K_l (\tilde{\sig_{\x}}) \K_l^T \right\}\\
       & = (\K_{l} \bmu_{\x})^{.2} +  {\rm vec} ({\k}_l^{.2} \ast \v),
\end{split}
\end{eqnarray}
where $\v \triangleq {\rm diag} ({\sig_{\x}})$.
Similarly, from (\ref{eq:update_w}) and (\ref{eq:diag_apprxo_IV}), the updating equation for $w$ reduces to:
\begin{eqnarray}
\begin{split}
w &= \sum_{i=1}^n [{\rm vec}(\bar{\h}\ast\D^{\top}\D{\h} \ast \v)]_i.
\end{split}
\end{eqnarray}

It is worthwhile to point out that the approximate computation of the covariance is only used for updating the parameters.
The computation of the covariance is also involved in the estimation of the HR image $\x$ in (\ref{eq:x1}) but the covariance matrix there is not approximately calculated.
As (\ref{eq:x1}) can be solved efficiently with Conjugate Gradient (CG) method, it is unnecessary to construct $\K$ and $\W_{\z}$  explicitly, as only the ability to evaluate a vector with respect to them is required in the CG algorithm, thus avoiding the intractability of explicitly constructing  the covariance matrix or resorting to approximation to it.

Note that in (\ref{eq:tau2}), the prior knowledge on the noise level is incorporated into the final noise estimation via the shape  and scale parameter $a^o$ and $b^o$ of the hyperprior.
If we have no knowledge available on the noise level, we can simply use a uninformative prior by setting  $a^o=1$ and $b^o=0$.
The overall procedure of the proposed method is summarized in Algorithm~\ref{fig:algo}.
Note that when the downsampling operator $\D$ is an identity matrix, our model reduces naturally to a deblurring algorithm.

\begin{algorithm}[tb]
\caption{(Empirical Bayesian Super-Resolution.).}
\begin{algorithmic}[1]
\STATE {\bf Input:} LR observation $\y$, zooming factor $r$, number of iterations $T$.
\STATE {\bf Initialize:}  set up $\D$ and $\HH$ according to $r$ and set the initial HR estimation  as $\tilde{\x}_{0}  = \HH^{\top}\D^{\top}\y$. The $\gam_l$'s are initialized as $\gam_l= \ 10^{-5} \mathbf{1}$.
\STATE  {\bf For} $i = 1, 2, \cdots, T$, do
        \begin{itemize}
        \item update the HR image $\x$ via (\ref{eq:x1});
        \item update the covariance approximation via (\ref{eq:diag_apprx_W}) and (\ref{eq:diag_apprxo_IV});
        \item update $\gam_l$ via  (\ref{eq:update_gamma_II});
        \item update $\tau$ via (\ref{eq:tau2}) and (\ref{eq:update_w}).
        \end{itemize}
\STATE {\bf End}
\STATE {\bf Output:} SR estimation $\x$.
\end{algorithmic}
\label{fig:algo}
\end{algorithm}

\begin{figure}[ht]
\centering
\begin{overpic}[width=1cm]{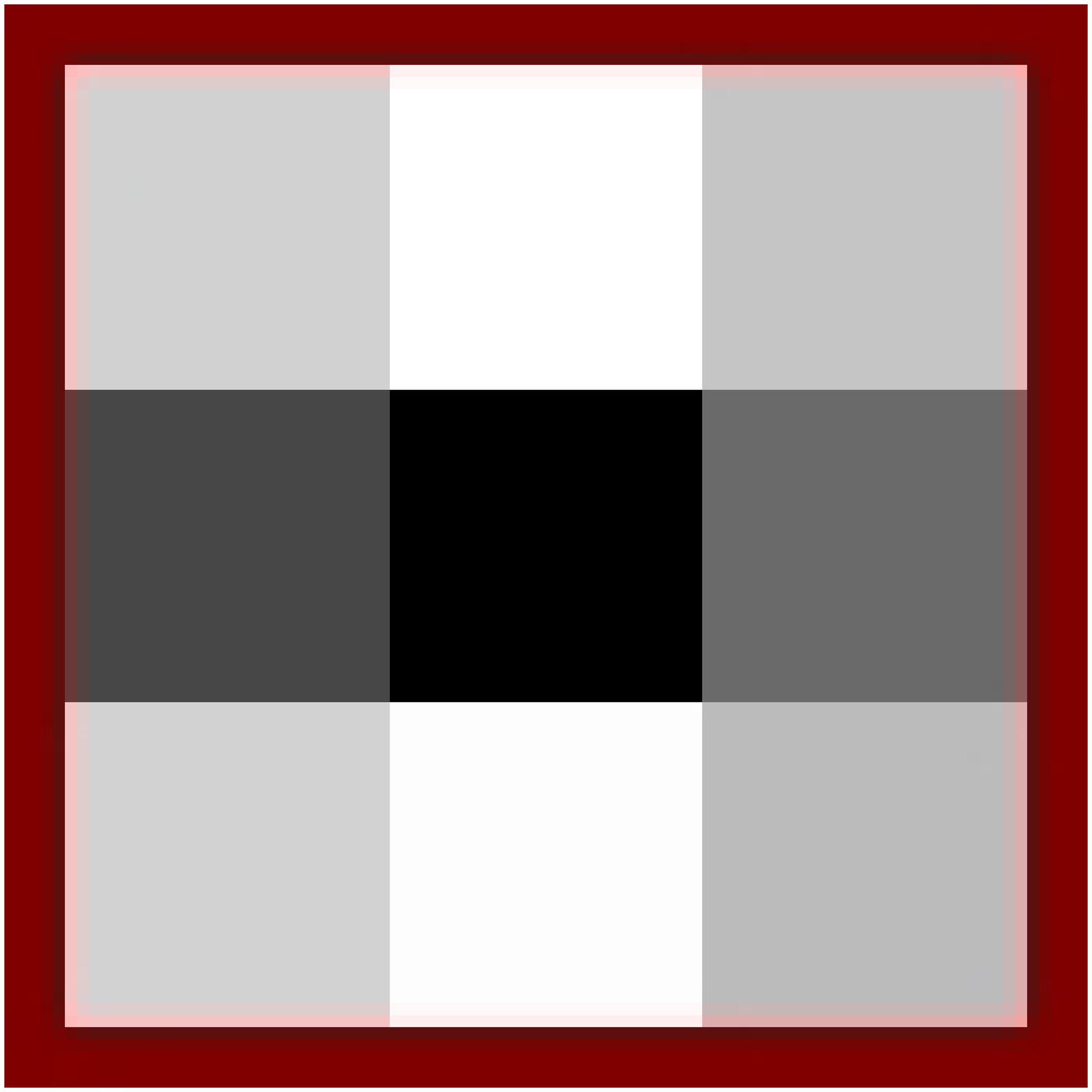}
\end{overpic}
\begin{overpic}[width=1cm]{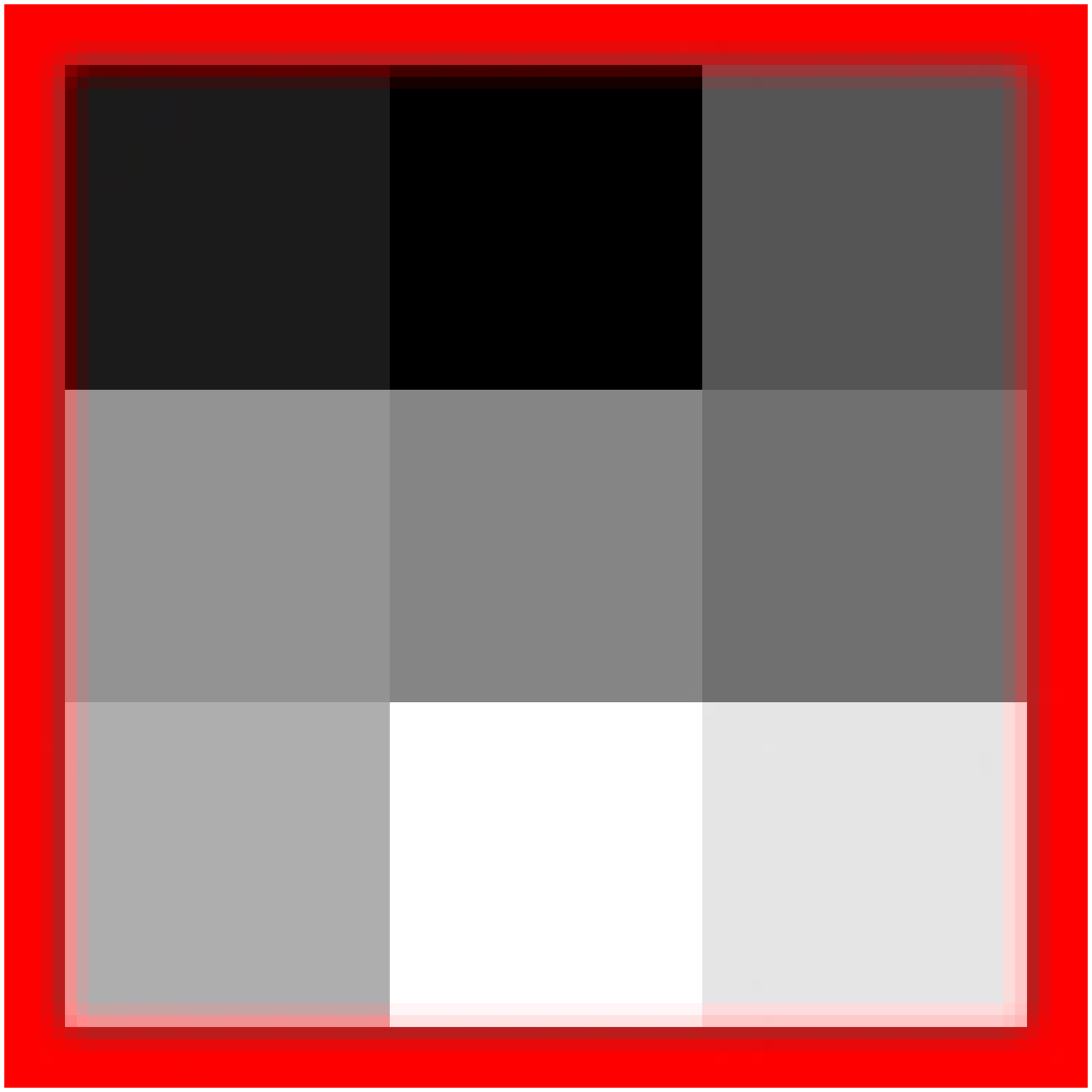}
\end{overpic}
\begin{overpic}[width=1cm]{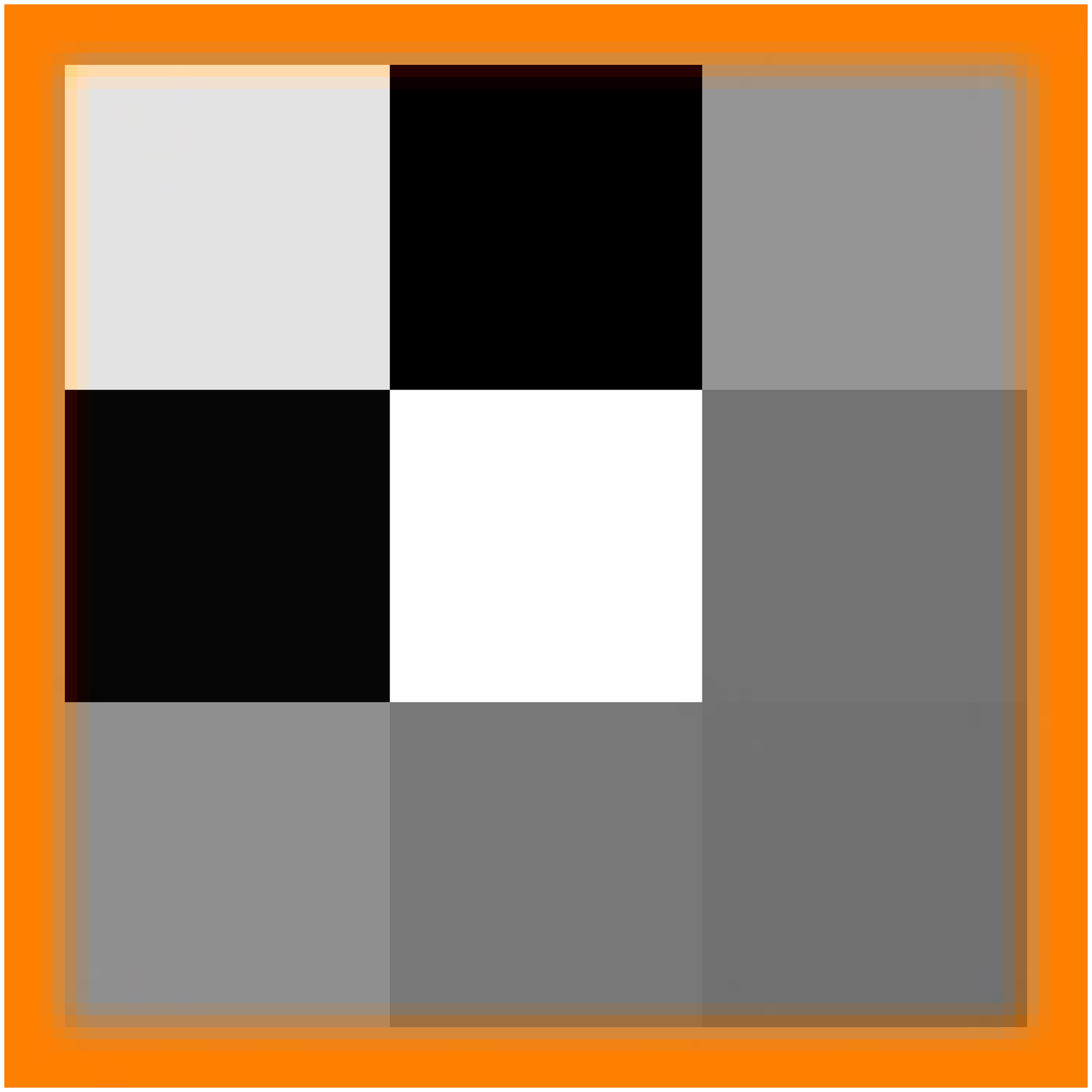}
\end{overpic}
\begin{overpic}[width=1cm]{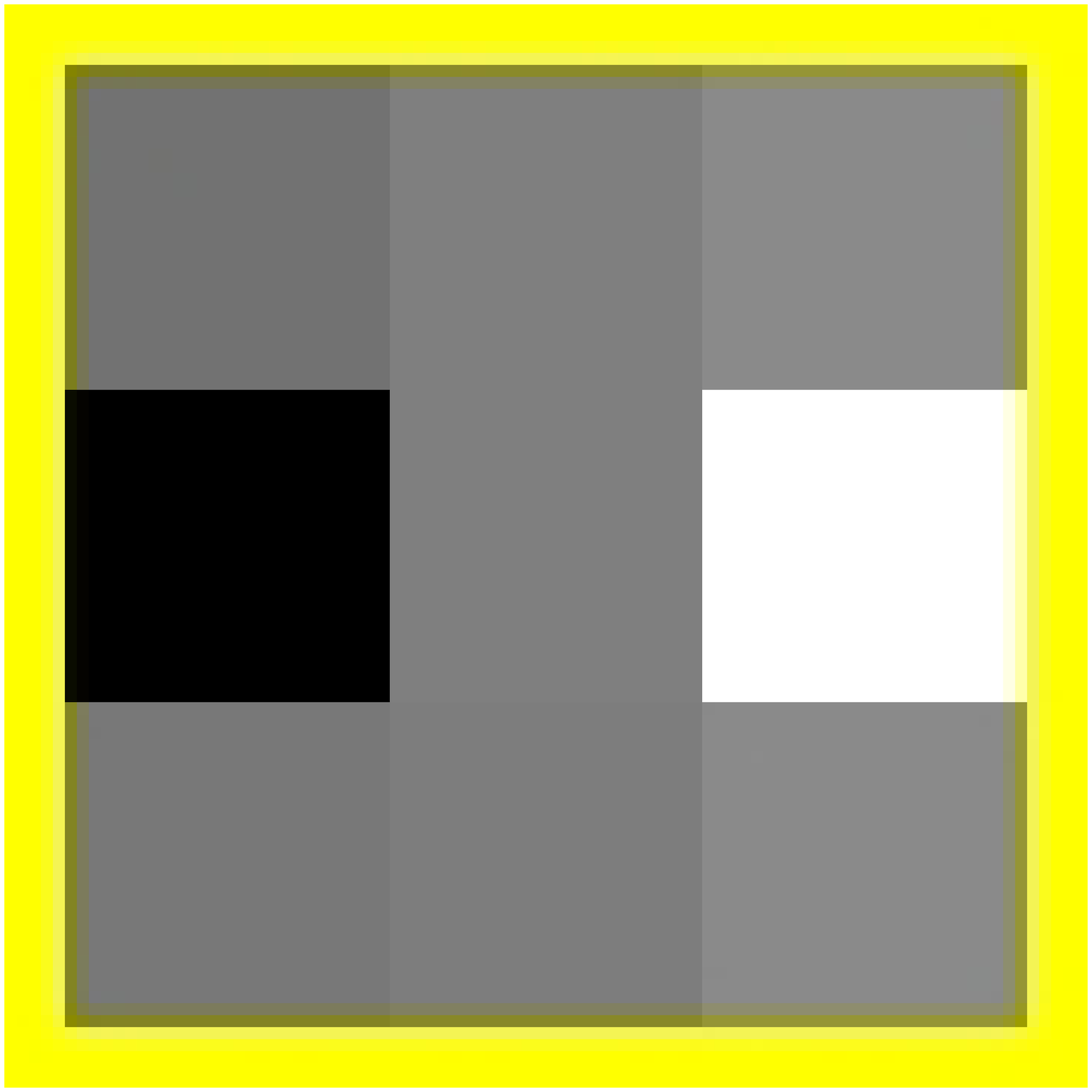}
\end{overpic}
\begin{overpic}[width=1cm]{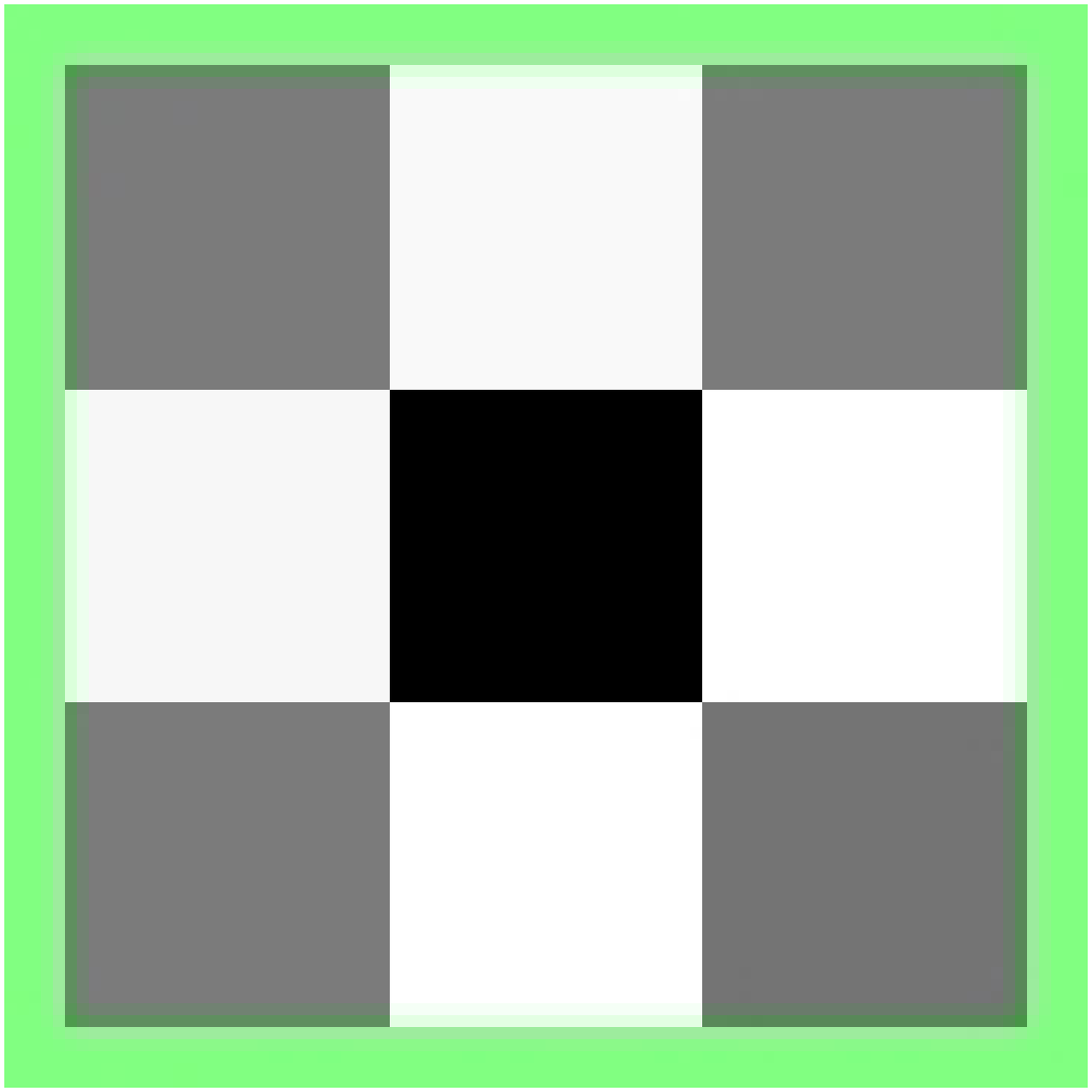}
\end{overpic}
\begin{overpic}[width=1cm]{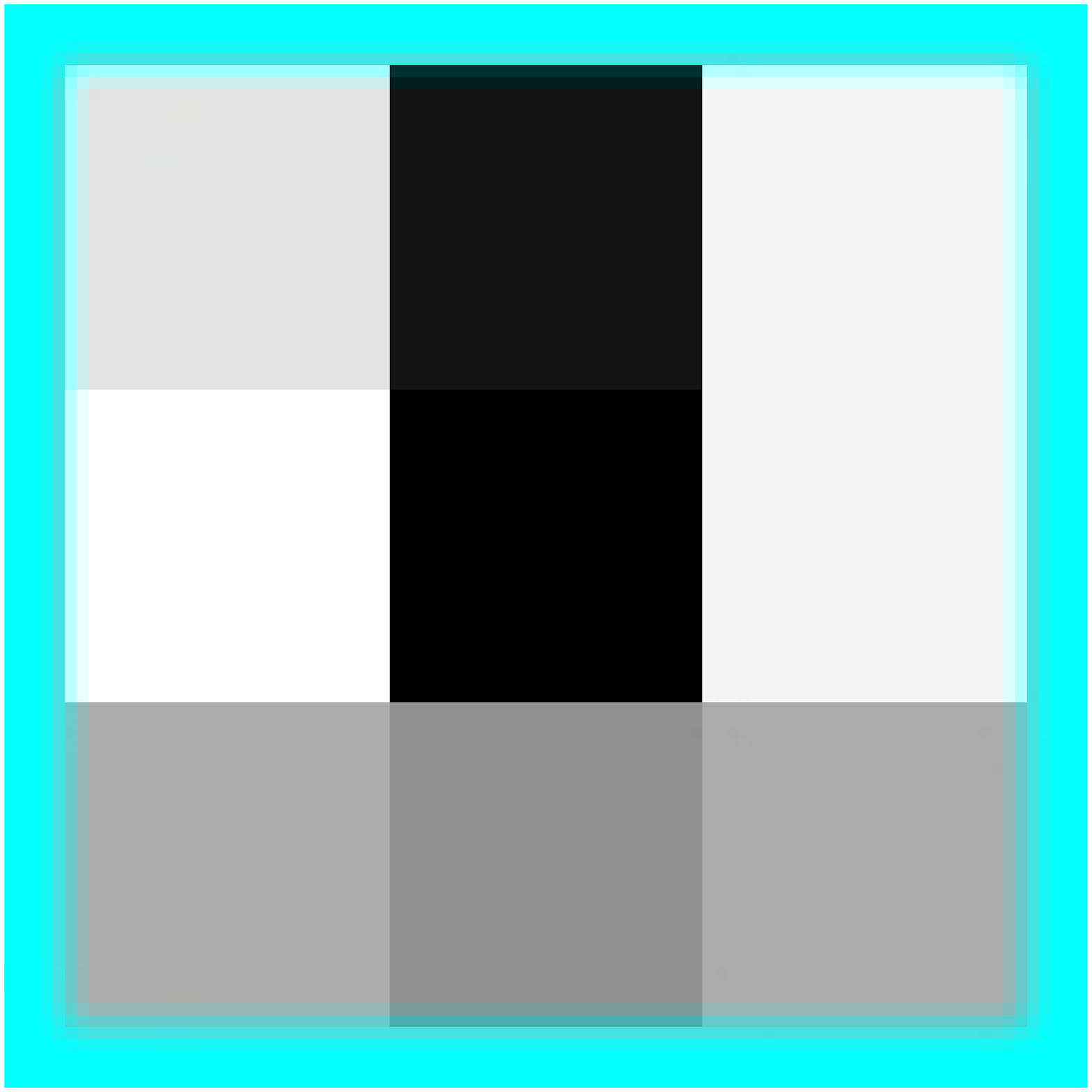}
\end{overpic}
\begin{overpic}[width=1cm]{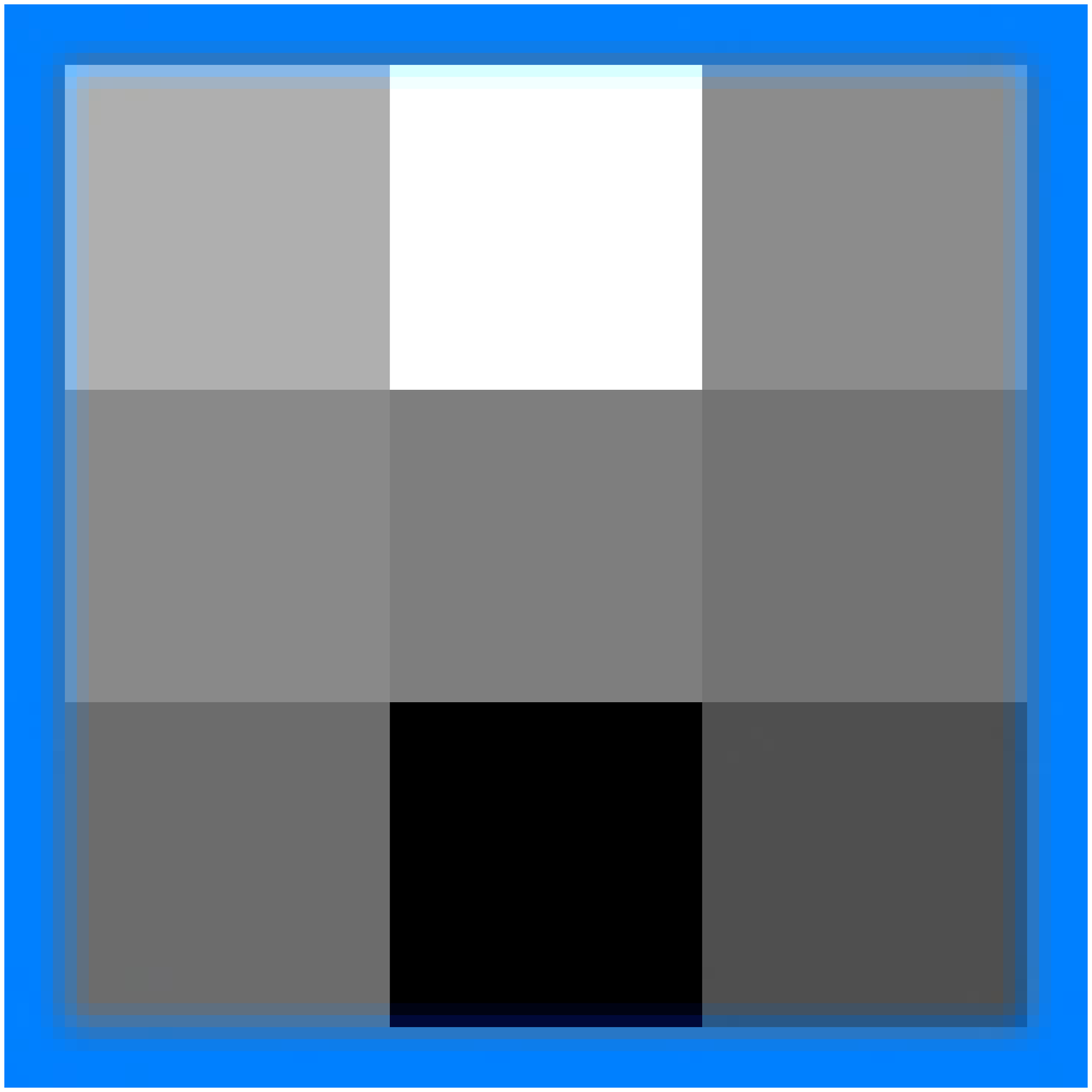}
\end{overpic}
\begin{overpic}[width=1cm]{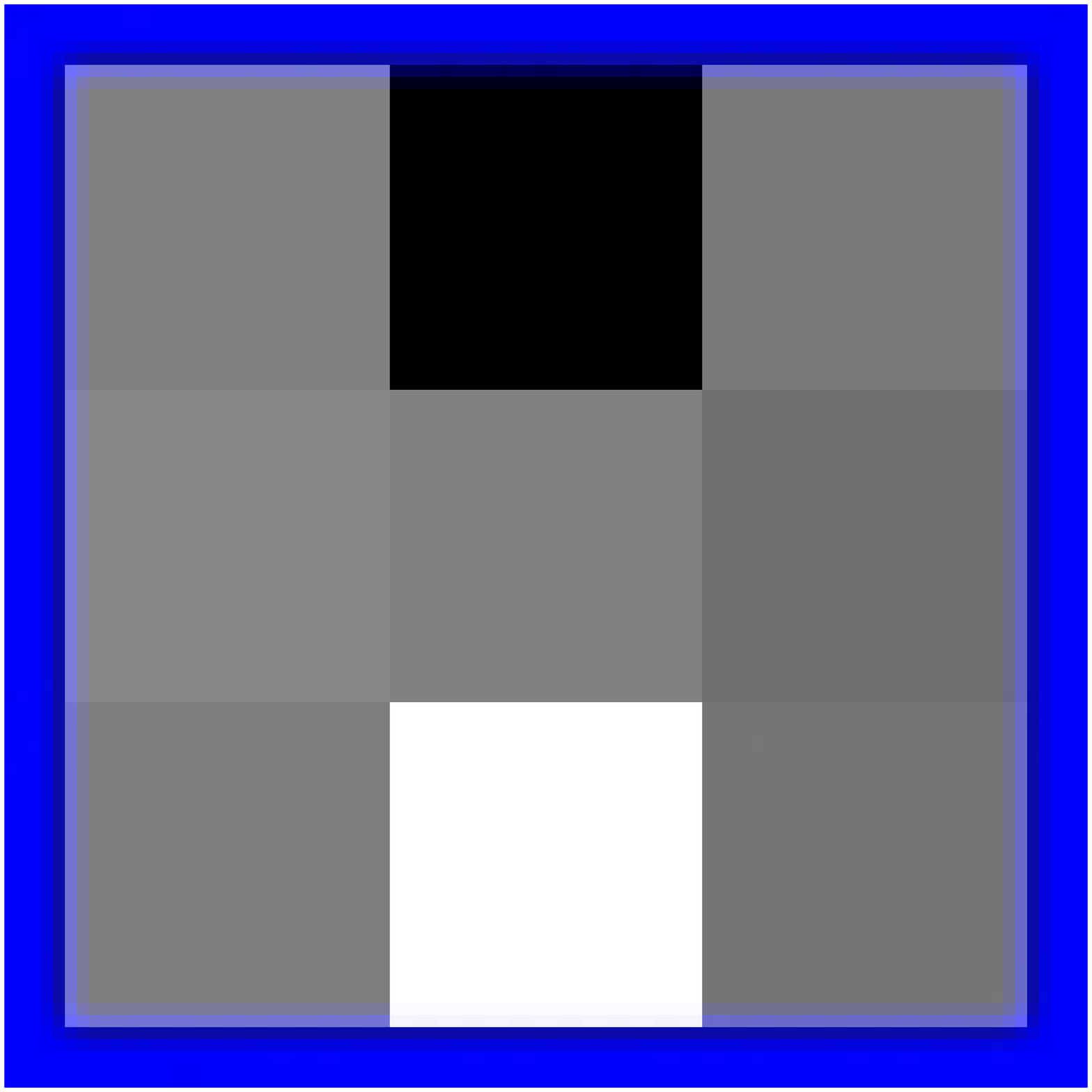}
\end{overpic}
\caption{Illustration of the $8$ learned filters from natural images~\cite{generative_MRF}. These filters are used in the FoE model with an infinite GSM as its expert functions. Note that the shape of this function is not determined beforehand but adaptively learned alongside the estimation process.}
\label{fig:fliter}
\end{figure}

\subsection{Implementation Details}
In this subsection,  we make some comments on the implementation details.
The low-passing filter corresponding to $\HH$  is modeled as a Gaussian filer with standard deviation of~$2$ when the zooming factor is $3$. The size of the Gaussian filer is set as $7$.
The filtering process is implemented  via FFT in the frequency domain.
The down sampling operator $\D$  is implemented via subsampling the image for every $r$ rows and columns for the zooming factor of $r$.
Eight experts (thus $L=8$) are multiplied together to form the energy potential $f(\cdot)$, which is a PoE model.
Each expert is associated with one of the eight different filters as shown in Figure~\ref{fig:fliter}.
These filters are used in the FoE model with a infinite GSM as its expert functions. Note that the shape of this function is not determined beforehand but adaptively learned alongside the estimation process.
In our implementation, we use filters $\{\k_l\}_{l=1}^L$ with size of $3\times 3$.
The filters  $\{\k_l\}_{l=1}^L$  for each GSM  are learned from the training images using the {\rv Contrastive Divergence} algorithm~\cite{PoE}, which are the same as those used in our previous work~\cite{NBSR_Zhang}.
Specifically, we use the implementation given by~\cite{generative_MRF} which is trained on the {\rv Berkeley} Segmentation database~\cite{BSD}.
Note that the training dataset  does not include any test images used for comparison in the experimental section.

\section{Experiments and Results}
\label{sec:Exp}
In this section, we conduct several experiments to evaluate the effectiveness of the proposed method, both qualitatively and quantitatively.
We first give an illustrative example to demonstrate some basic properties of the proposed algorithm.
Then we compare the SR estimation results with several classical as well as \emph{state-of-the-art} SR methods, on several standard test images as well as real-world color images.
Finally, the robustness of the proposed SR method with respect to noise is also investigated.
Peak Signal to Noise Ratio (PSNR) is adopted as an evaluation metric which is defined as:
\begin{eqnarray}\nonumber
{\rm PSNR} = 10 \log_{10} \frac{255^2}{{\frac{1}{m}\sum_i(\hat{\x} - \x)^2_i}},
\end{eqnarray}
where $\hat{\x}$ denotes the estimated SR image and $\x$ denotes the original HR image. $m$ is the total number of pixels in $\x$.
 The Structural SIMilarity index (SSIM) is  used as another objective measure, which aligns better with the human perception than PSNR~\cite{SSIM}.

\begin{table}
\begin{center}
\caption{Super resolution  ($\times 3$) result comparison of estimation quality.}
\addtolength{\tabcolsep}{-2pt}\renewcommand{\arraystretch}{1}
\label{table:SR1}
\begin{tabular}{|c|c|cccccc|}
\hline\noalign{}
\multicolumn{2}{|c|}{Methods} &  NN & BI&  Fast~\cite{Shan_SR} & ScSR~\cite{Yang_SR_TIP} & NBSR~\cite{NBSR_Zhang}& Proposed \\
\hline\hline
\multirow{2}{*}{house}&PSNR &  24.90  & 25.62 & 28.72  & 30.80 &   {32.06}& {32.85}\\
&SSIM  & 0.763  &  0.788  &  0.837 &0.835  & {0.897}& {0.897}\\
\hline
\multirow{2}{*}{peppers}&PSNR &  21.59   &   22.18 &24.38 & 25.52 &    {25.88}& {26.88}\\
&SSIM  & 0.727  & 0.775  &  0.841& 0.863 &{0.910}& {0.918}\\
\hline
\multirow{2}{*}{cameraman}&PSNR &  21.71   &   22.12& 24.64  & 25.79 &  {26.42}& {26.65} \\
&SSIM  & 0.695  & 0.714  &  0.774 & 0.810 & {0.847} & {0.846}\\
\hline
\multirow{2}{*}{barbara}&PSNR &  23.07   &   23.43& 24.58  & 25.24 & 25.62&{25.60} \\
&SSIM  & 0.618  & 0.649  &  0.683 &  0.732 &0.753& {0.749} \\
\hline
\multirow{2}{*}{lena}&PSNR &  25.91 &	26.62& 29.59& 	31.89& {33.25}& {33.40}\\
&SSIM  & 0.769&	0.802&	0.833& 0.888& {0.911}& {0.909}\\
\hline
\multirow{2}{*}{boat}&PSNR &  24.14& 	24.63& 	26.70& 28.25&29.47& {29.34} \\
&SSIM  & 0.659 & 	0.687& 	0.729 & 0.815&	{0.844}& {0.836}\\
\hline
\multirow{2}{*}{hill}&PSNR & 25.88&	26.41&	27.72& 30.15& {30.44}& {30.84} \\
&SSIM  & 0.654&	0.686&	0.714 & 0.811&	{0.831} & {0.823}\\
\hline
\multirow{2}{*}{couple}&PSNR & 24.16&	24.60& 26.34&	27.69& {28.53}& {28.64}	  \\
&SSIM  & 0.621& 	0.649&	0.698& 0.788&	{0.811} & {0.811}\\
\hline
\end{tabular}
\end{center}
\end{table}

\begin{figure*}[ht]
\centering
\begin{overpic}[height=4.cm]{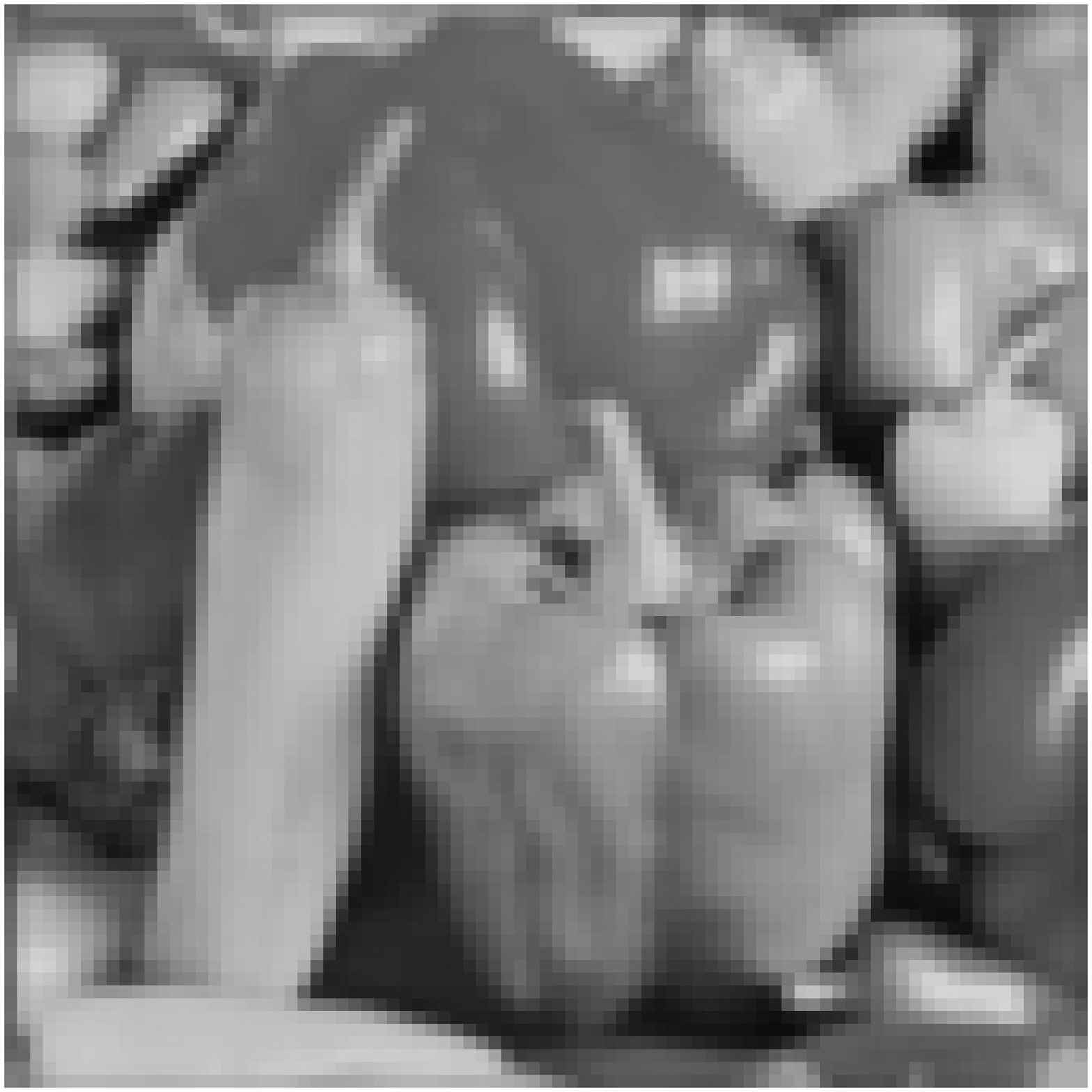}
\put(51,-41){{\includegraphics[viewport = 400 150 500 230, clip, height=1.55cm]{Peppers_NN_PSNR21.588_SSIM0.72713.eps}}}
\put(0,-41){{\includegraphics[viewport = 200 290 300 370, clip, height=1.55cm]{Peppers_NN_PSNR21.588_SSIM0.72713.eps}}}
\put(1,1){\sffamily \footnotesize{\textcolor{white}{\textsc{Nearest Neighbor}}}}
\end{overpic}
\begin{overpic}[height=4.cm]{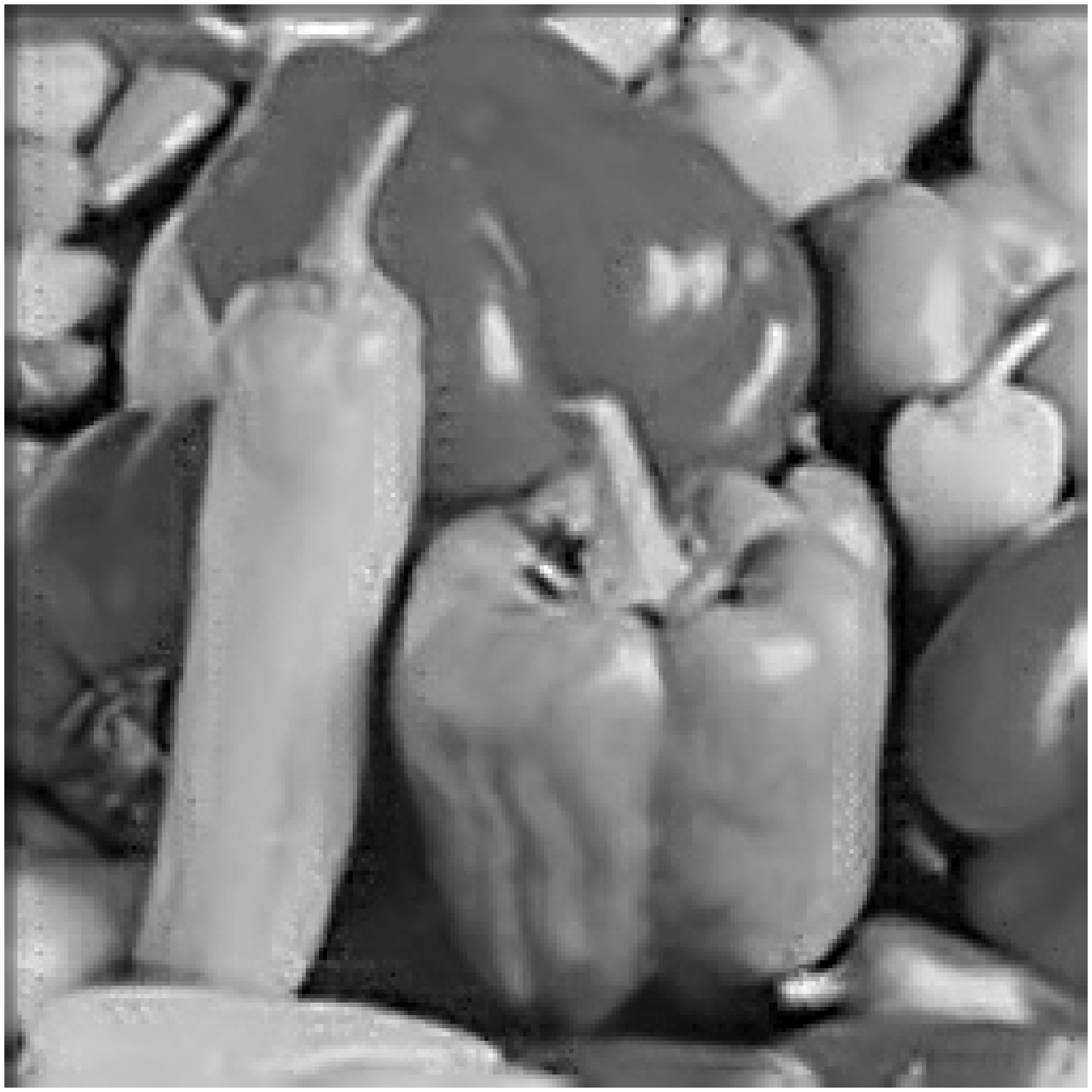}
\put(51,-41){{\includegraphics[viewport = 400 150 500 230, clip, height=1.55cm]{Peppers_SSR_PSNR25.5188_SSIM0.86268.eps}}}
\put(0,-41){{\includegraphics[viewport = 200 290 300 370, clip, height=1.55cm]{Peppers_SSR_PSNR25.5188_SSIM0.86268.eps}}}
\put(1,1){\sffamily \footnotesize{\textcolor{white}{\textsc{ScSR}}}}
\end{overpic}
\begin{overpic}[height=4.cm]{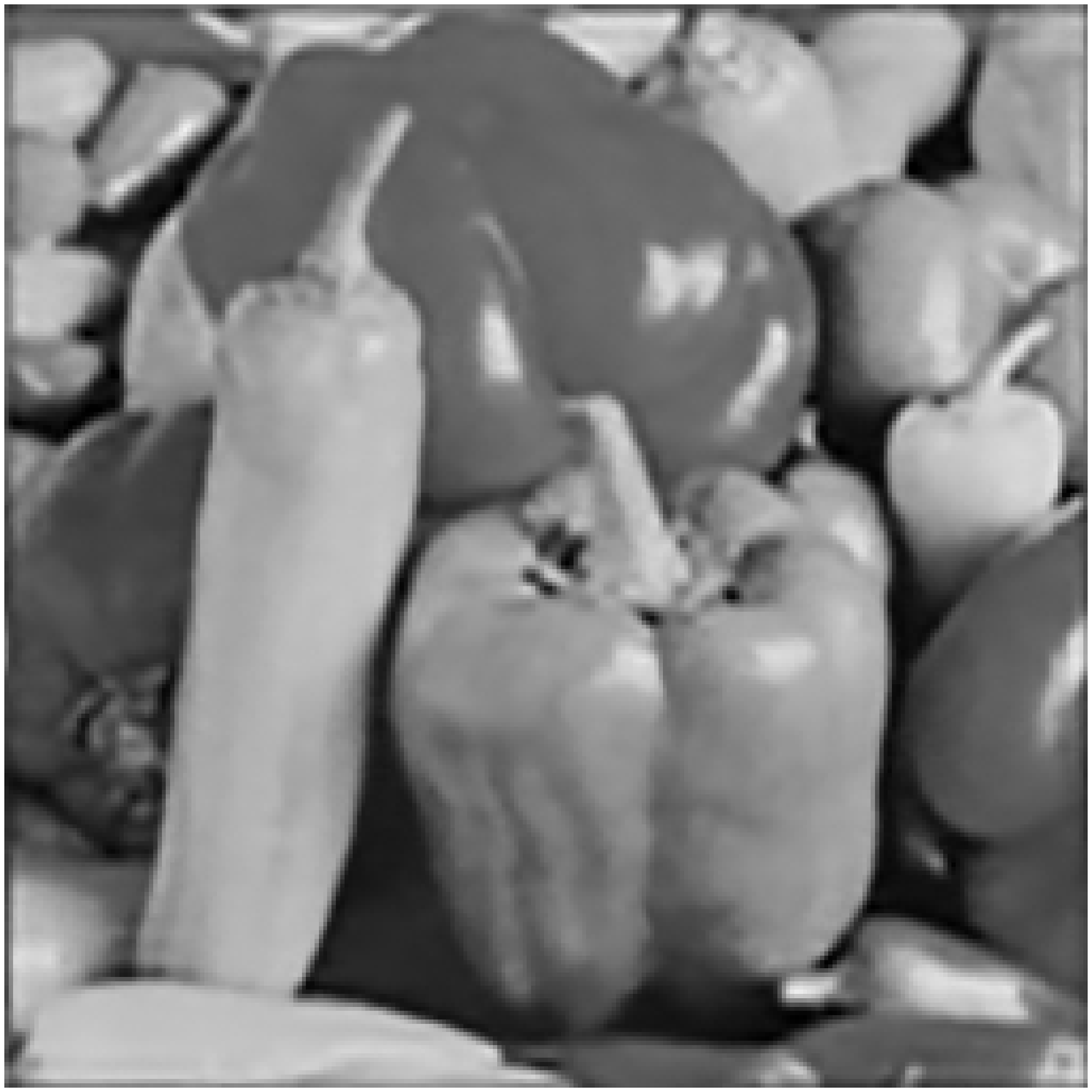}
\put(51,-41){{\includegraphics[viewport = 400 150 500 230, clip, height=1.55cm]{Peppers_HR_PSNR25.8797RMSE_12.9586_SSIM0.90986.eps}}}
\put(0,-41){{\includegraphics[viewport =200 290 300 370, clip, height=1.55cm]{Peppers_HR_PSNR25.8797RMSE_12.9586_SSIM0.90986.eps}}}
\put(1,1){\sffamily \footnotesize{\textcolor{white}{\textsc{NBSR}}}}
\end{overpic}
\begin{overpic}[height=4.cm]{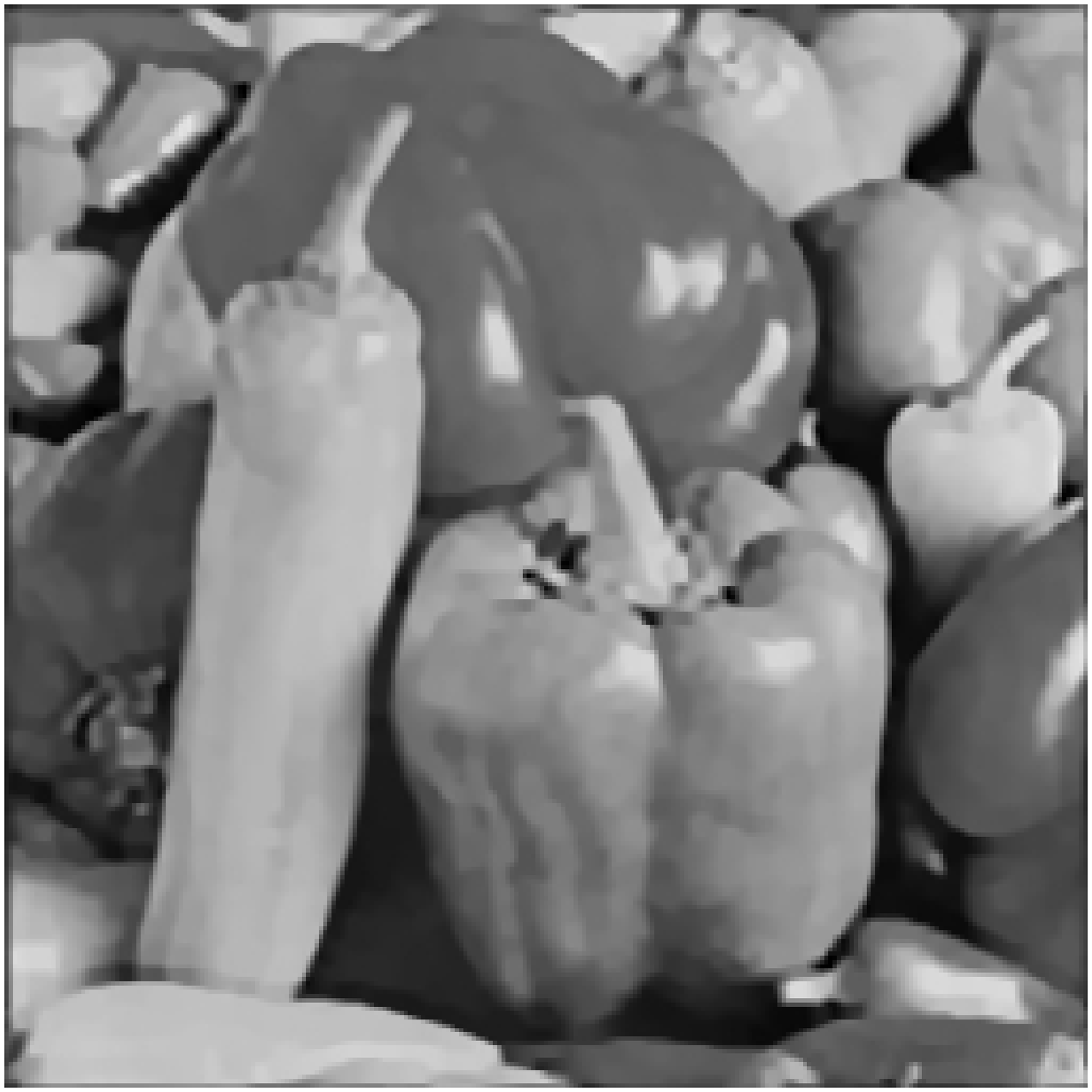}
\put(51,-41){{\includegraphics[viewport = 400 150 500 230, clip, height=1.55cm]{peppers_EBSR_Iter10_PSNR26.877496SSIM0.918278.eps}}}
\put(0,-41){{\includegraphics[viewport = 200 290 300 370, clip, height=1.55cm]{peppers_EBSR_Iter10_PSNR26.877496SSIM0.918278.eps}}}
\put(1,1){\sffamily \footnotesize{\textcolor{white}{\textsc{Proposed}}}}
\end{overpic}
\\
\vspace{0.67in}
\begin{overpic}[height=4.cm]{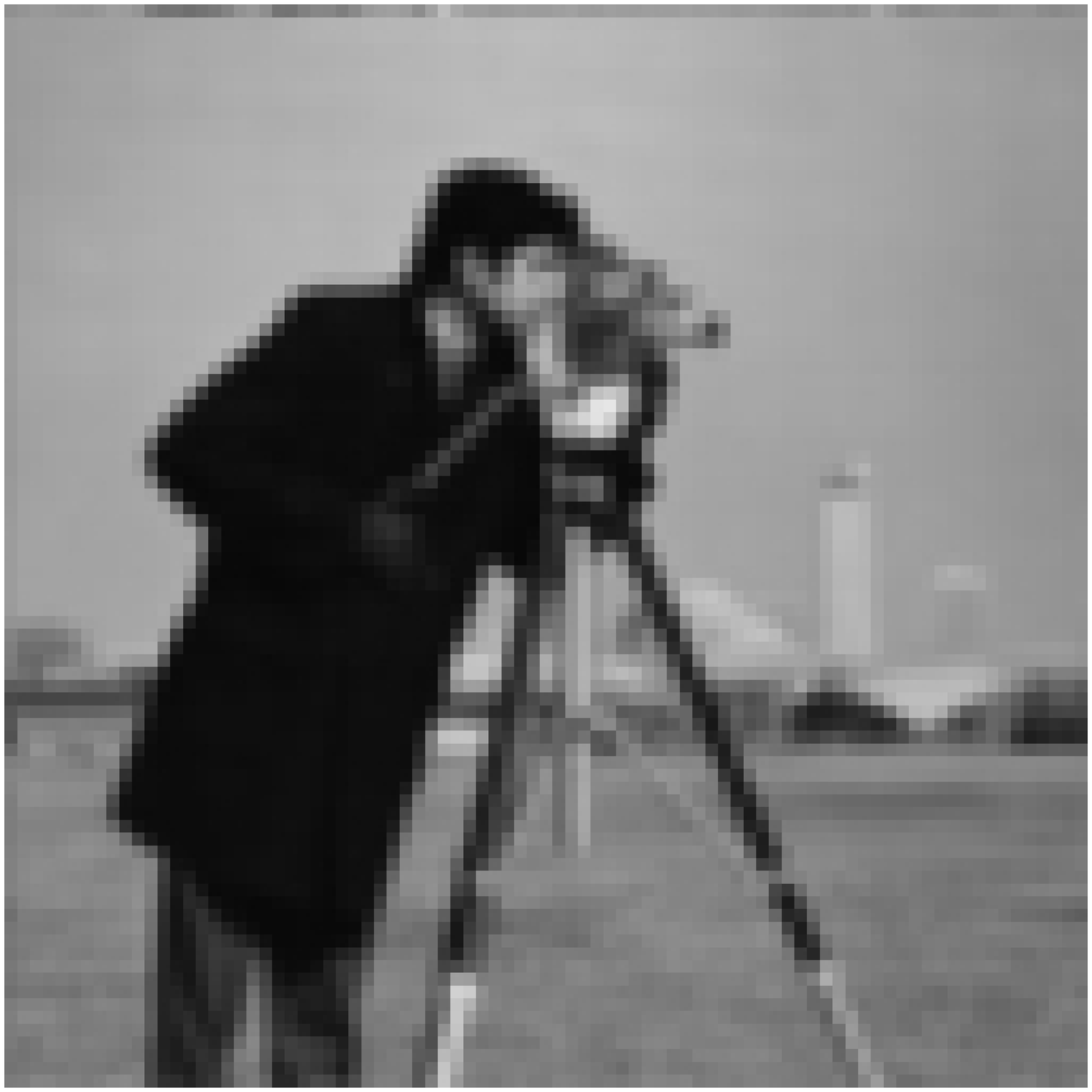}
\put(51,-41){{\includegraphics[viewport = 300 250 400 330, clip, height=1.55cm]{cam_NN_PSNR21.7083_SSIM0.69542.eps}}}
\put(0,-41){{\includegraphics[viewport = 250 430 350 510, clip, height=1.55cm]{cam_NN_PSNR21.7083_SSIM0.69542.eps}}}
\put(1,1){\sffamily \footnotesize{\textcolor{white}{\textsc{Nearest Neighbor}}}}
\end{overpic}
\begin{overpic}[height=4.cm]{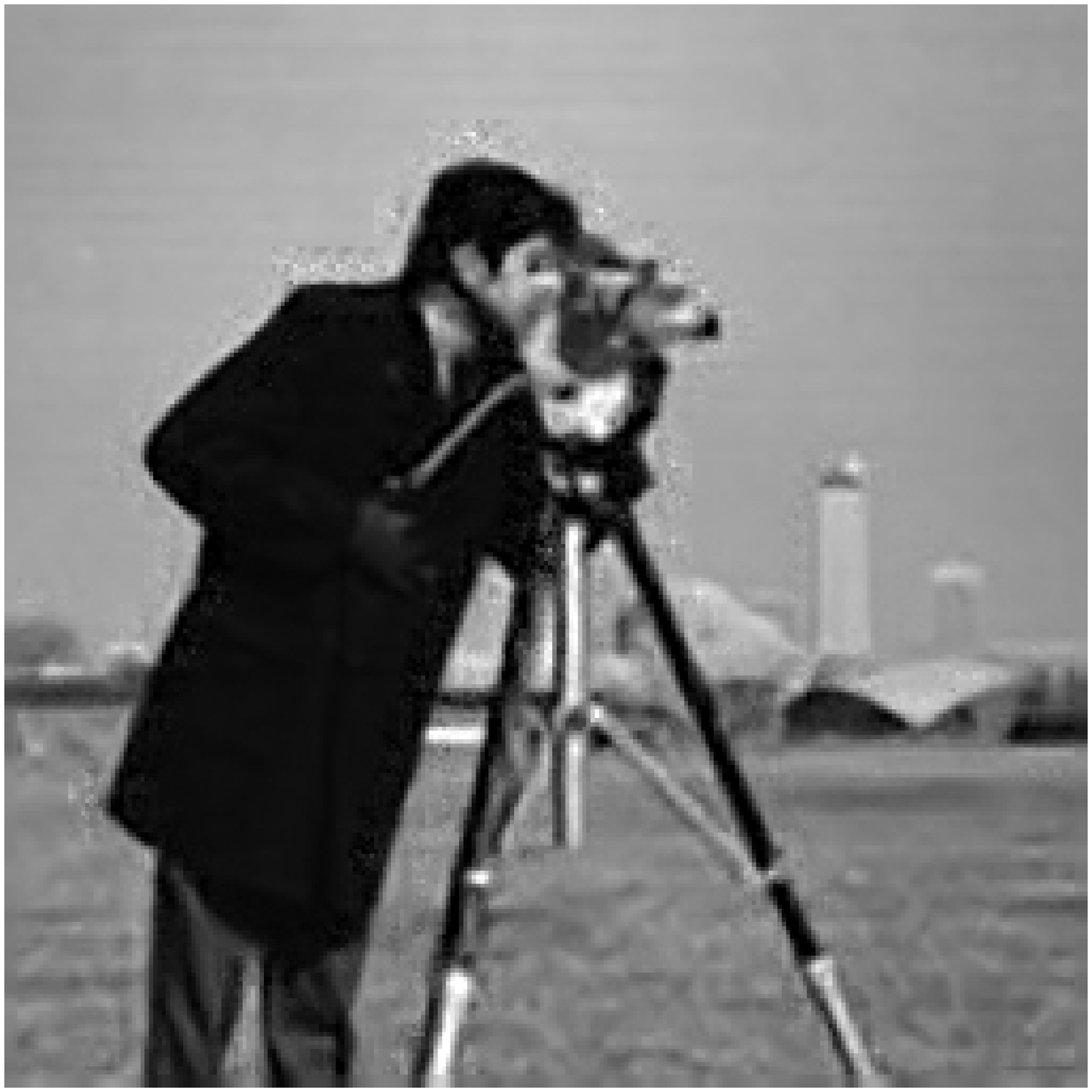}
\put(51,-41){{\includegraphics[viewport = 300 250 400 330, clip, height=1.55cm]{cam_SSR_PSNR25.7871_SSIM0.81008.eps}}}
\put(0,-41){{\includegraphics[viewport = 250 430 350 510, clip, height=1.55cm]{cam_SSR_PSNR25.7871_SSIM0.81008.eps}}}
\put(1,1){\sffamily \footnotesize{\textcolor{white}{\textsc{ScSR}}}}
\end{overpic}
\begin{overpic}[height=4.cm]{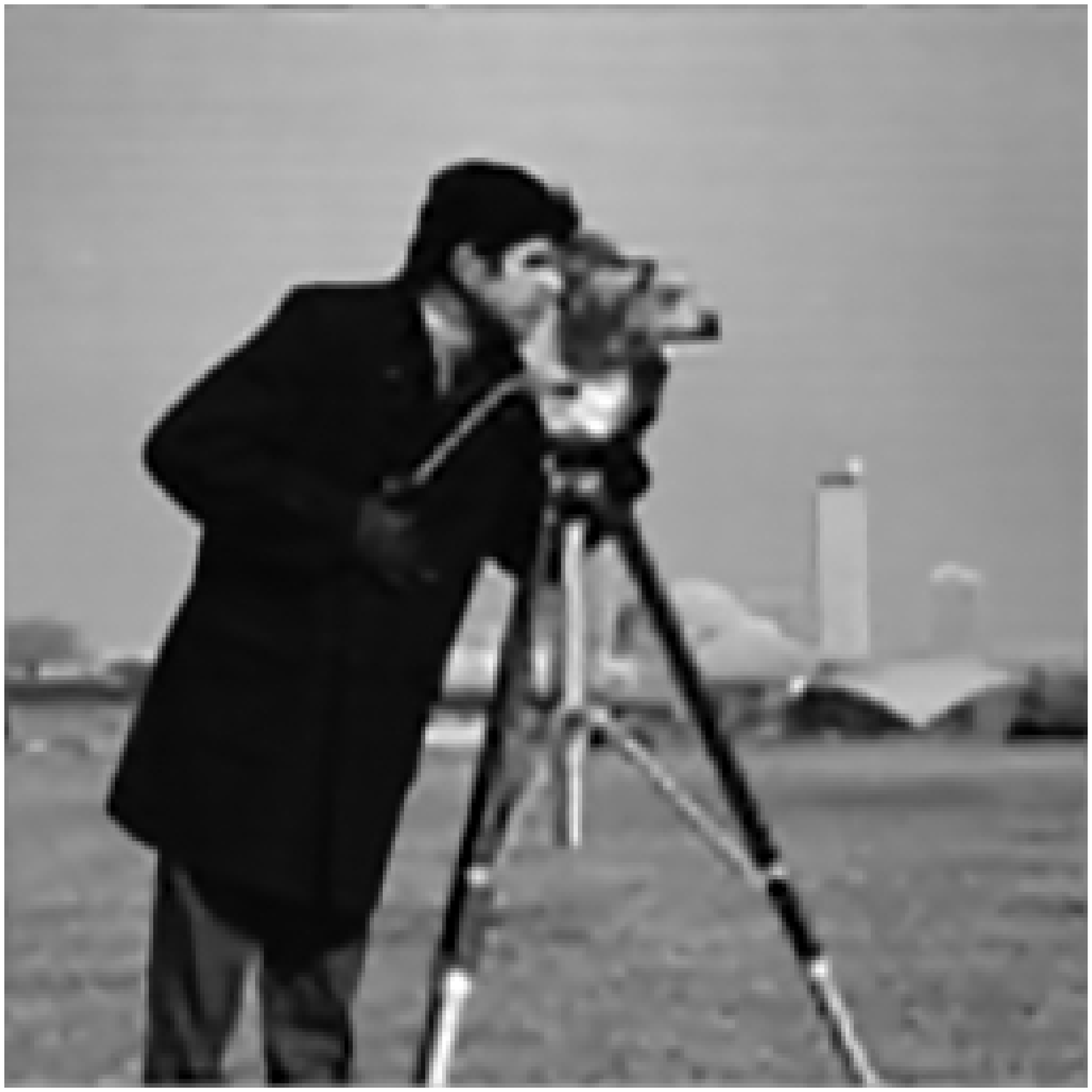}
\put(51,-41){{\includegraphics[viewport = 300 250 400 330, clip, height=1.55cm]{cam_HR_PSNR26.4673RMSE_12.1107_SSIM0.8466.eps}}}
\put(0,-41){{\includegraphics[viewport = 250 430 350 510, clip, height=1.55cm]{cam_HR_PSNR26.4673RMSE_12.1107_SSIM0.8466.eps}}}
\put(1,1){\sffamily \footnotesize{\textcolor{white}{\textsc{NBSR}}}}
\end{overpic}
\begin{overpic}[height=4.cm]{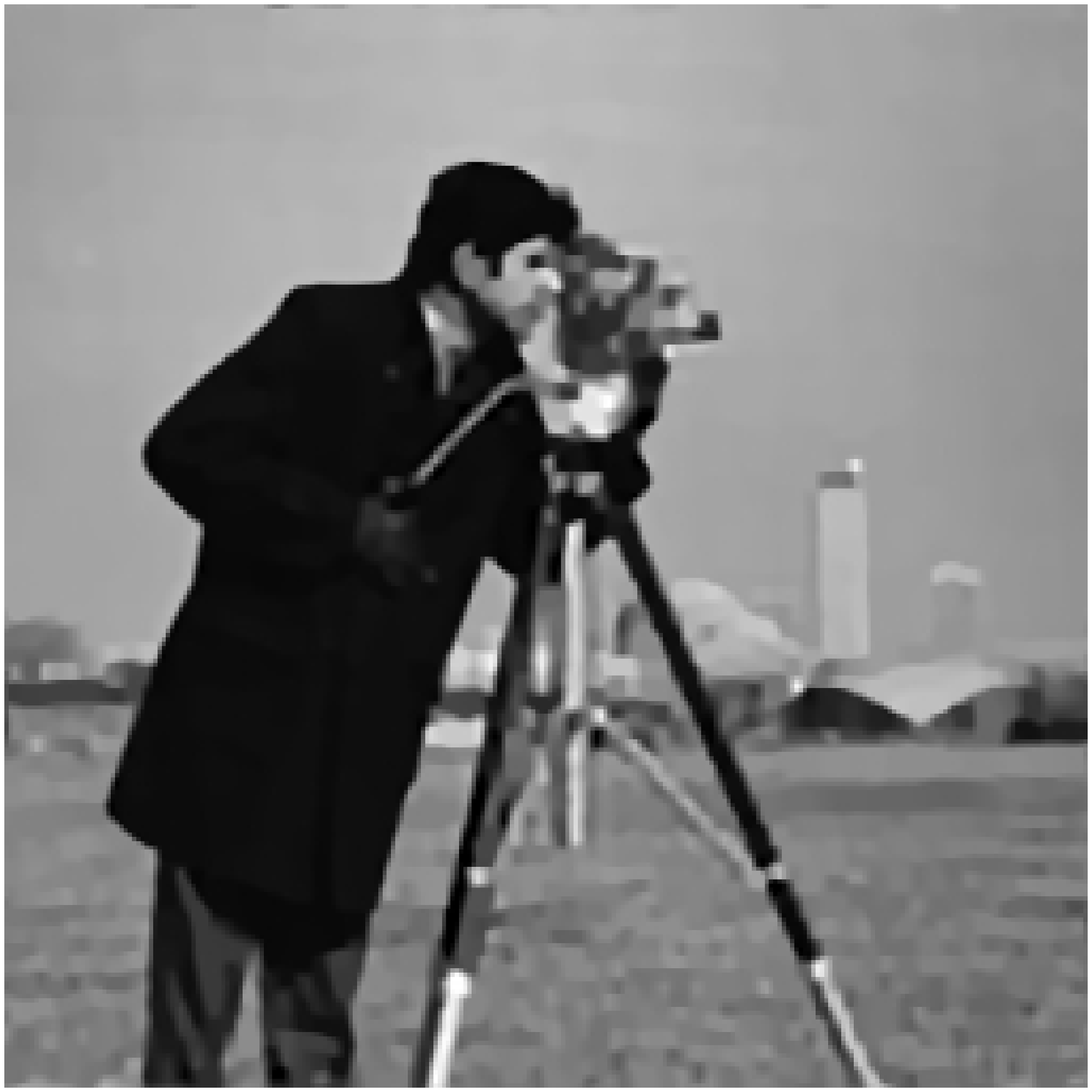}
\put(51,-41){{\includegraphics[viewport = 300 250 400 330, clip, height=1.55cm]{cam_EBSR_Iter10_PSNR26.645617SSIM0.845522.eps}}}
\put(0,-41){{\includegraphics[viewport = 250 430 350 510, clip, height=1.55cm]{cam_EBSR_Iter10_PSNR26.645617SSIM0.845522.eps}}}
\put(1,1){\sffamily \footnotesize{\textcolor{white}{\textsc{Proposed}}}}
\end{overpic}
\vspace{0.5in}
\caption{Single image super resolution results  for `{\ttfamily peppers}', `{\ttfamily cameraman}' and `{\ttfamily lena}' images ($\times 3$) with different algorithms: Nearest Neighbor Interpolation, sparse representation based SR method~\cite{Yang_SR_TIP}, the NBSR method~\cite{NBSR_Zhang} and the proposed EBSR method. The results evaluated in terms of PSNR and SSIM are reported in Table~\ref{table:SR1}.}
\label{fig:Res_test}
\end{figure*}

%
%
%

\subsection{Super Resolution Experiment}\label{eq:SR}
To  verify the effectiveness of the proposed method, we  compare the SR performance of the proposed method with several algorithms: Nearest Neighbor interpolation (NN), bicubic interpolation (BI), fast SR method~\cite{Shan_SR}, ScSR method~\cite{Yang_SR_TIP},\footnote{For the ScSR method, the dictionary size is $1024$ and the regularization parameter is set as $\lambda=0.1$  which is the same setting as in \cite{Yang_SR_TIP}.},  Variational Bayesian based SR (VBSR) method~\cite{VBSR}\footnote{codes are available at {\footnotesize \ttfamily http://decsai.ugr.es/pi} {\ttfamily /superresolution/software.html}}, as well as the Natural image prior based Bayesian SR method using sampling (NBSR)~\cite{NBSR_Zhang}.
In this experiment, we use the standard test images in the noise-free setting  for evaluation  and use PSNR and SSIM as the objective measures.
The zooming factor in this experiment is $r=3$.
The SR results in terms of PSNR and SSIM are summarized  in Table~\ref{table:SR1}.
As can be seen from Table~\ref{table:SR1}, the proposed method can generate SR images with better quality in terms of PSNR and SSIM than the fast SR method~\cite{Shan_SR} and the ScSR method~\cite{Yang_SR_TIP} on all the testing images.
Also, the proposed method performs comparable or better than the NBSR~\cite{NBSR_Zhang} method.
This clearly demonstrates the effectiveness of the proposed method.
To further compare the SR results visually, we show some of the SR results from different algorithms in Figure~\ref{fig:Res_test}.
{\rv As can be seen from Figure~\ref{fig:Res_test}, the fast SR method tends to removing the fine details from the image and the SR results  exhibit cartoon-like artifacts.
Furthermore, it is noticed from Figure~\ref{fig:Res_test} that the SR results from the ScSR method have some artifacts along the edge structures (see the zoomed patches for `{\ttfamily peppers}' and `{\ttfamily cameraman}'  images in  Figure~\ref{fig:Res_test}).} This may result from the inconsistency in recovering the HR patches from the neighboring LR  patches.  On the other hand, the SR image from the proposed method does not suffer from these artifacts, due to the usage of the global statistical model as well as the MMSE estimation criteria for the latent HR images. This clearly demonstrates the benefit of using a global statistical model as a prior based on natural image statistics, combined with empirical Bayesian estimation scheme.

\begin{table*}
\begin{center}
\caption{Super resolution  ($\times 4$) quality comparison ($\sigma=1$).}
\label{table:SR_Comp_VB}
\begin{tabular}{|c|c|cccccccc|}
\hline\noalign{}
\multicolumn{2}{|c|}{Methods} &  NN & BI & Fast~\cite{Shan_SR}& KRR~\cite{Kim_SR_PAMI}  & ScSR~\cite{Yang_SR_TIP} & VBSR~\cite{VBSR} & NBSR~\cite{NBSR_Zhang}& Proposed  \\
\hline\hline
\multirow{2}{*}{house}&PSNR &  22.24   & 23.14 &  26.79 & 27.60& 26.86& 27.00 & 28.20&   \bf{28.81}\\
&SSIM  & 0.627  &  0.671 & 0.771&0.785&  0.742 &0.767  &  0.767 &\bf{0.802}\\
\hline
\multirow{2}{*}{peppers}&PSNR &  21.47   &   22.20 &25.36&25.95 & 25.61 & 25.36 &  26.90  &\bf{27.51}\\
&SSIM  & 0.647  & 0.711  &0.786 &0.821&  0.764 &0.807& 0.837 &\bf{0.841}\\
\hline
\multirow{2}{*}{cameraman}&PSNR & 18.35     &   18.91 &  21.06 &21.35& 21.17 & 21.50 & 21.72 &  \bf{22.43} \\
&SSIM  & 0.695  & 0.714  &  0.742&0.758& 0.695 &0.725 &  0.760 &\bf{0.794}\\
\hline
\multirow{2}{*}{barbara}&PSNR &  21.91   &   21.50 &24.12& 24.75 &  24.62  &24.12 & 24.94 & \bf{25.15}\\
&SSIM  & 0.515  & 0.582   & 0.640&0.696 & 0.654 & 0.699 & \bf{0.703} &{0.701}\\
\hline
\multirow{2}{*}{lena}&PSNR & 21.68  &	22.41& 26.66 & 26.88& 27.17 & 27.62 & 28.32 & \bf{ 29.25}\\
&SSIM  & 0.624 &	0.678 &  0.766& 0.807&  0.760	&0.814&0.825 &\bf{0.829}\\
\hline
\multirow{2}{*}{boat}&PSNR & 19.52 & 	20.17& 22.47& 23.27& 22.94&	23.81& 23.57 &\bf{24.36}\\
&SSIM  & 0.659 & 	0.687& 0.614& 0.652& 0.598	&0.657&	0.640 &\bf{0.686}\\
\hline
\multirow{2}{*}{hill}&PSNR & 24.56&	25.17&  26.94& 27.75& 24.92 &	27.10& {27.94} & \bf{28.44} \\
&SSIM  & 0.540 &	0.581 & 0.623&0.666 & 0.669	&0.675&	{0.680} & \bf{0.682} \\
\hline
\multirow{2}{*}{couple}&PSNR & 21.98 &	22.41 &24.08& 24.79& 25.17 &	25.59& 25.11 &	\bf{25.53}  \\
&SSIM  & 0.524& 	0.564 & 0.655&	0.697& 0.641 &0.707&	{0.663} & \bf{0.719}\\
\hline
\end{tabular}
\end{center}
\end{table*}

\begin{figure}[t]
\centering
\begin{overpic}[width= 7cm]{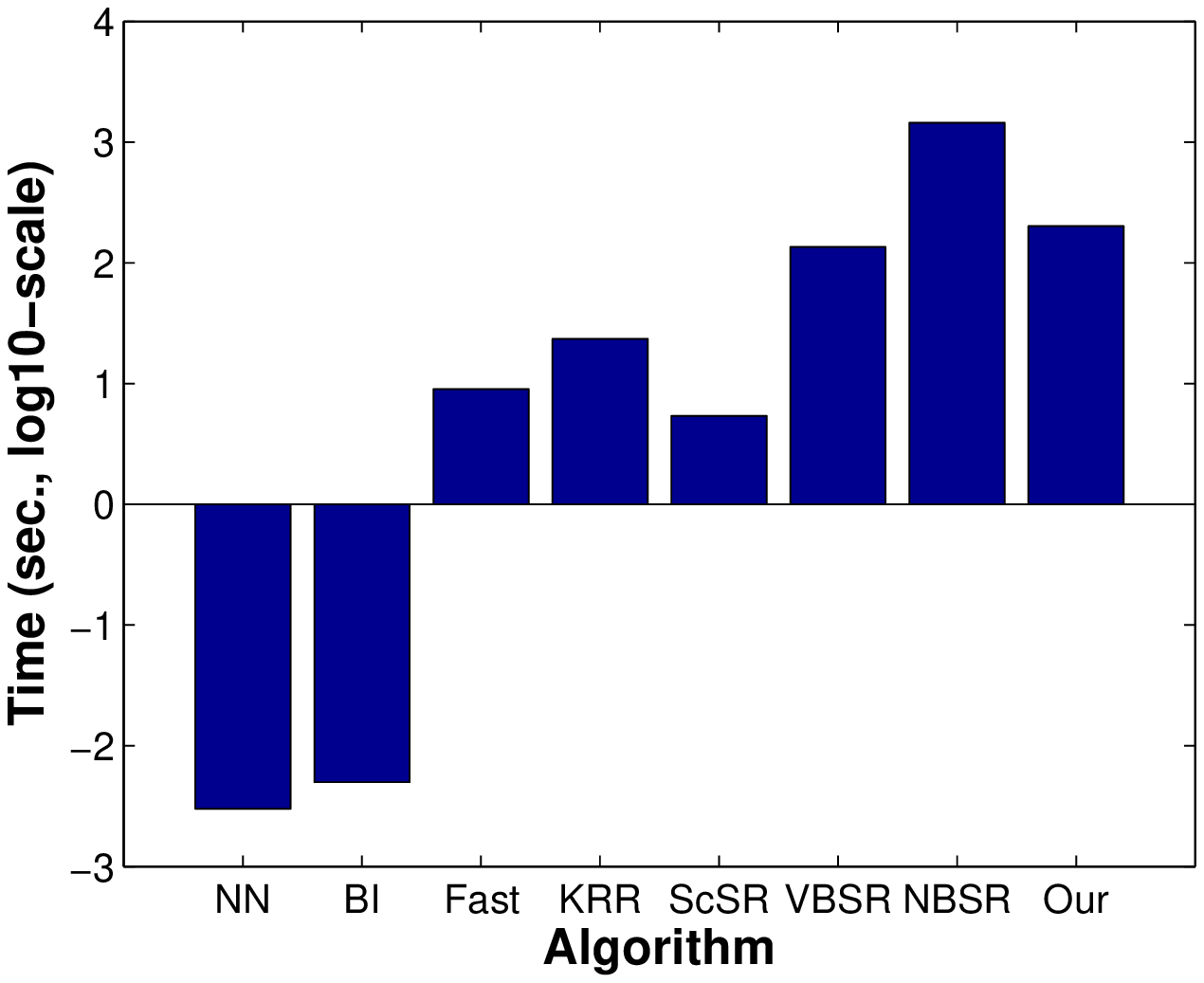}
\put(23, 72){\sffamily \footnotesize{\textcolor{black}{{Time consuming comparison}}}}
\end{overpic}\\
\begin{overpic}[width= 7cm]{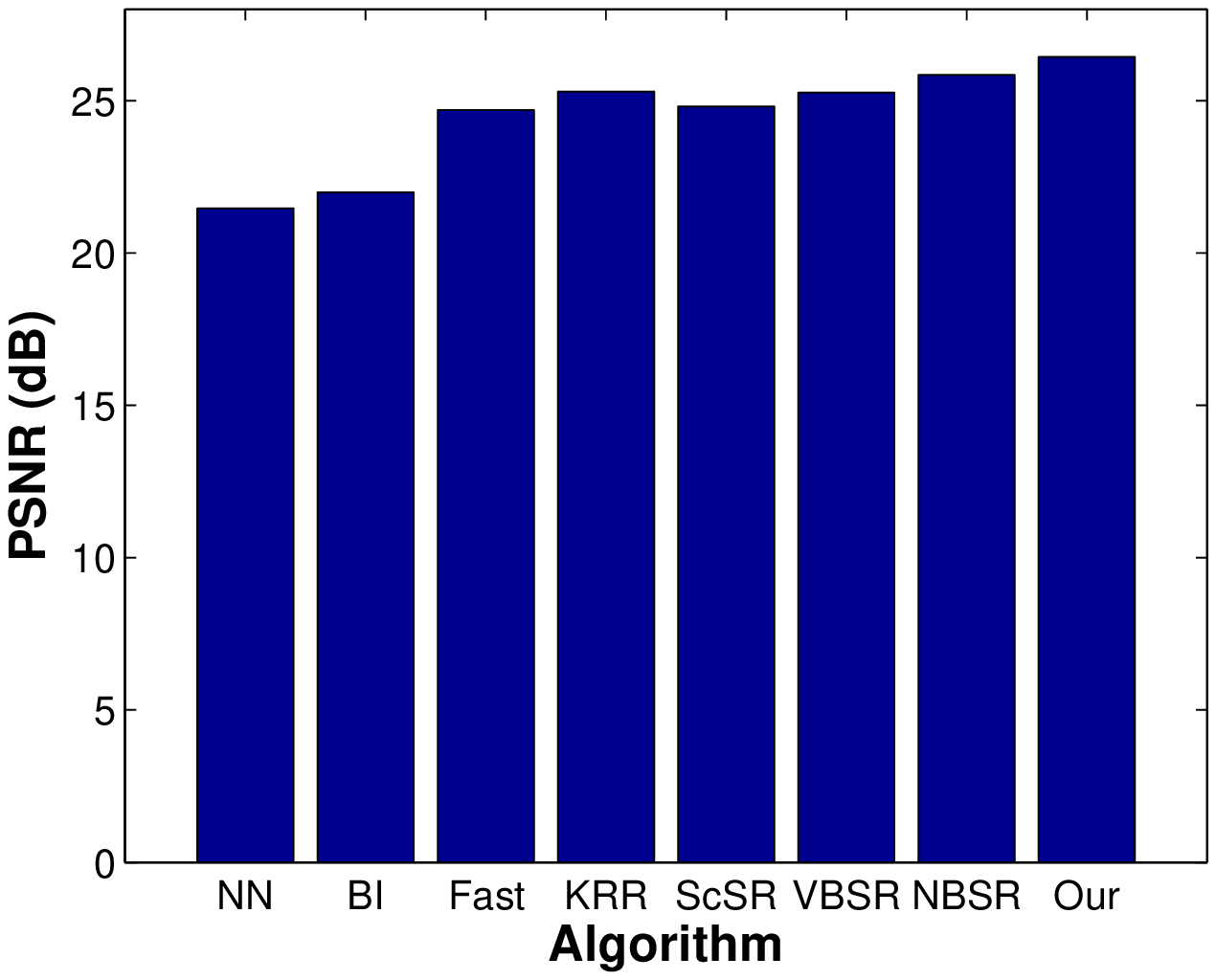}
\put(23,72){\sffamily \footnotesize{\textcolor{black}{{PSNR results comparison}}}}
\end{overpic}
\begin{overpic}[width= 7cm]{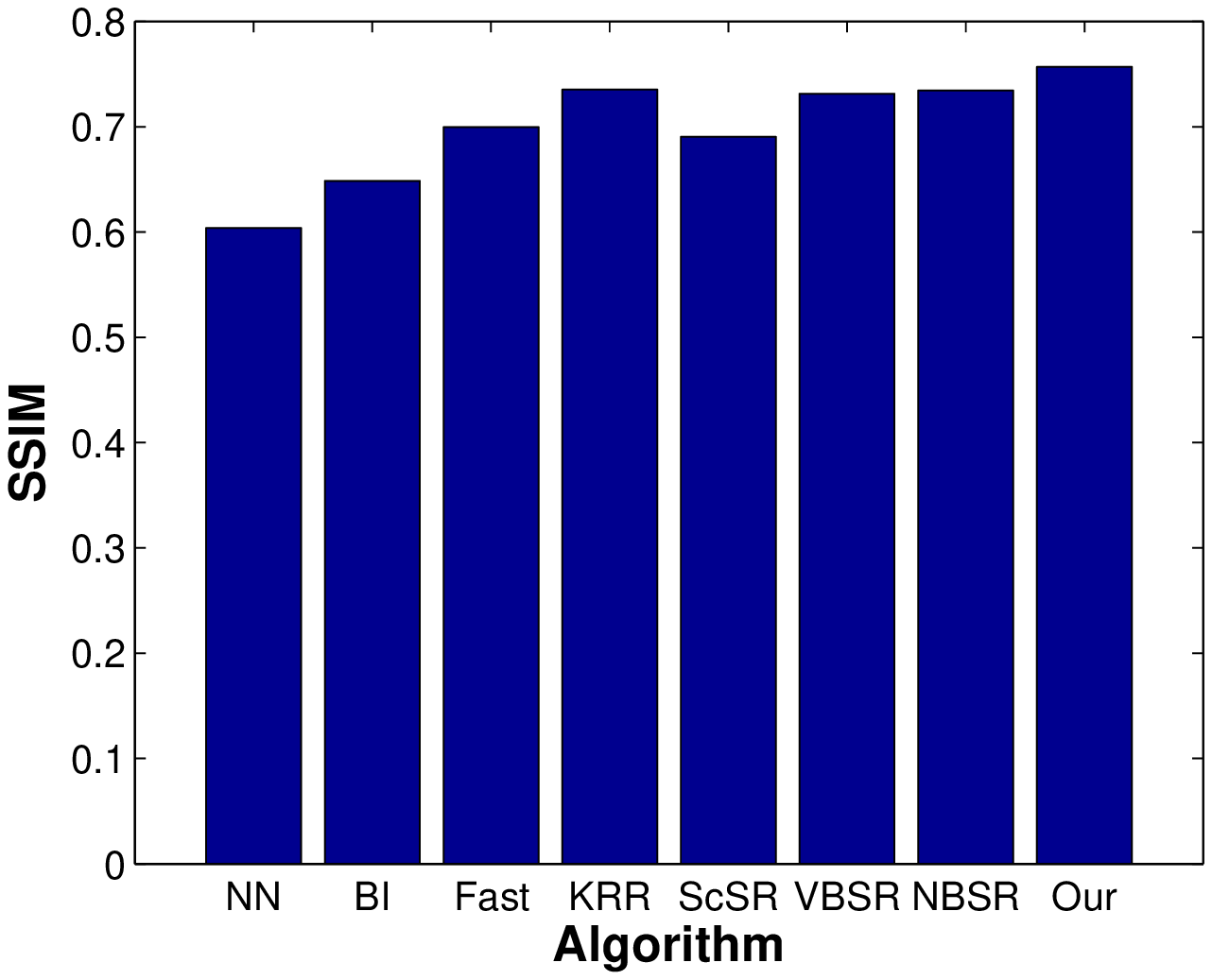}
\put(23,72){\sffamily \footnotesize{\textcolor{black}{{SSIM results comparison}}}}
\end{overpic}
\caption{Computational time consuming comparison and the corresponding SR performs comparison in terms of PSNR and SSIM.}
\label{fig:comp_time}
\end{figure}

 To further verify the effectiveness of the proposed method, we compare its SR performance with more SR algorithms, including the Kernel Ridge Regression SR  (KRR) method~\cite{Kim_SR_PAMI}, the Variational Bayesian SR (VBSR) method~\cite{VBSR} and  NBSR~\cite{NBSR_Zhang}, with zooming factor of $4$.\footnote{As both the KRR and VBSR methods can only handle SR tasks with even zooming factor, they are not compared in Table~\ref{table:SR1}.}  The KRR method achieves SR via patch-based sparse kernel regression for capturing the mapping between low resolution and high resolution patches. The VBSR method uses a simple  TV prior model and estimates the HR image with posterior mean via  variational Bayesian approximation.  As in the current implementation of the VBSR method, the matrices corresponding to the low-pass filtering as well as the down-sampling operator are  constructed explicitly, thus it can only handle small images. Therefore, we crop image parts of size $192\times 192$ from the standard test images and summarize the results in Table~\ref{table:SR_Comp_VB}. As can be seen from Table~\ref{table:SR_Comp_VB}, the KRR method performs better than the fast SR method in general, and the VBSR method overall performs similar to  the ScSR method and  the KRR method.
Finally, the proposed method outperforms the VBSR method and all the other methods constantly  in terms of  both PSNR and SSIM, which  demonstrates the superiority  of the proposed method.
The average computational time for different algorithms as well as their corresponding SR performance in terms of PSNR and SSIM are shown  in Figure~\ref{fig:comp_time}.\footnote{The computational time is measured on a Laptop with Intel Core2 Duo P8600 (2.4GHz) CPU and 1GB RAM.} The typical number of iteration  is 10 for the fast SR method, 30 for  VBSR, 100 for NBSR  and  10 for the proposed method. The other algorithms in Table~\ref{table:SR_Comp_VB} are `one-shot' methods that require no iteration.
As can be observed from Figure~\ref{fig:comp_time} that proposed method outperforms the state-of-art NBSR~\cite{NBSR_Zhang} method in terms of HR estimation quality  while can reduce the computational time by around an order of magnitude.
Besides, the proposed method has similar computational cost  with the VBSR~\cite{VBSR} while the proposed method performs much better.

\begin{figure*}[t]
\centering
\begin{overpic}[height=4.cm]{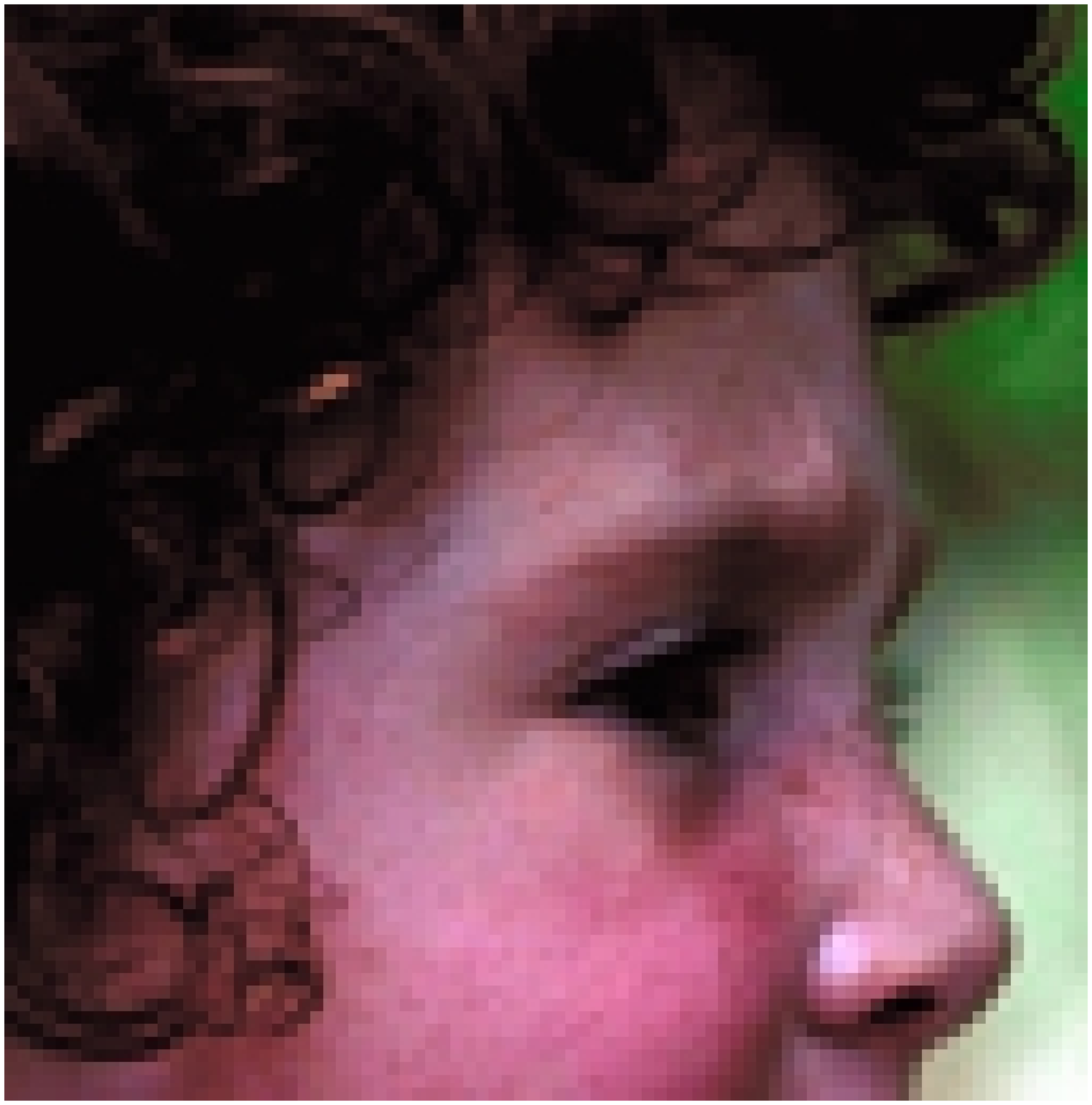}
\put(50,-41){{\includegraphics[viewport = 280 200 380 280, clip, height=1.55cm]{girl_NN.eps}}}
\put(0,-41){{\includegraphics[viewport = 450 30 550 110, clip, height=1.55cm]{girl_NN.eps}}}
\put(1,1){\sffamily \footnotesize{\textcolor{white}{\textsc{Nearest Neighbor}}}}
\end{overpic}
\begin{overpic}[height=4.cm]{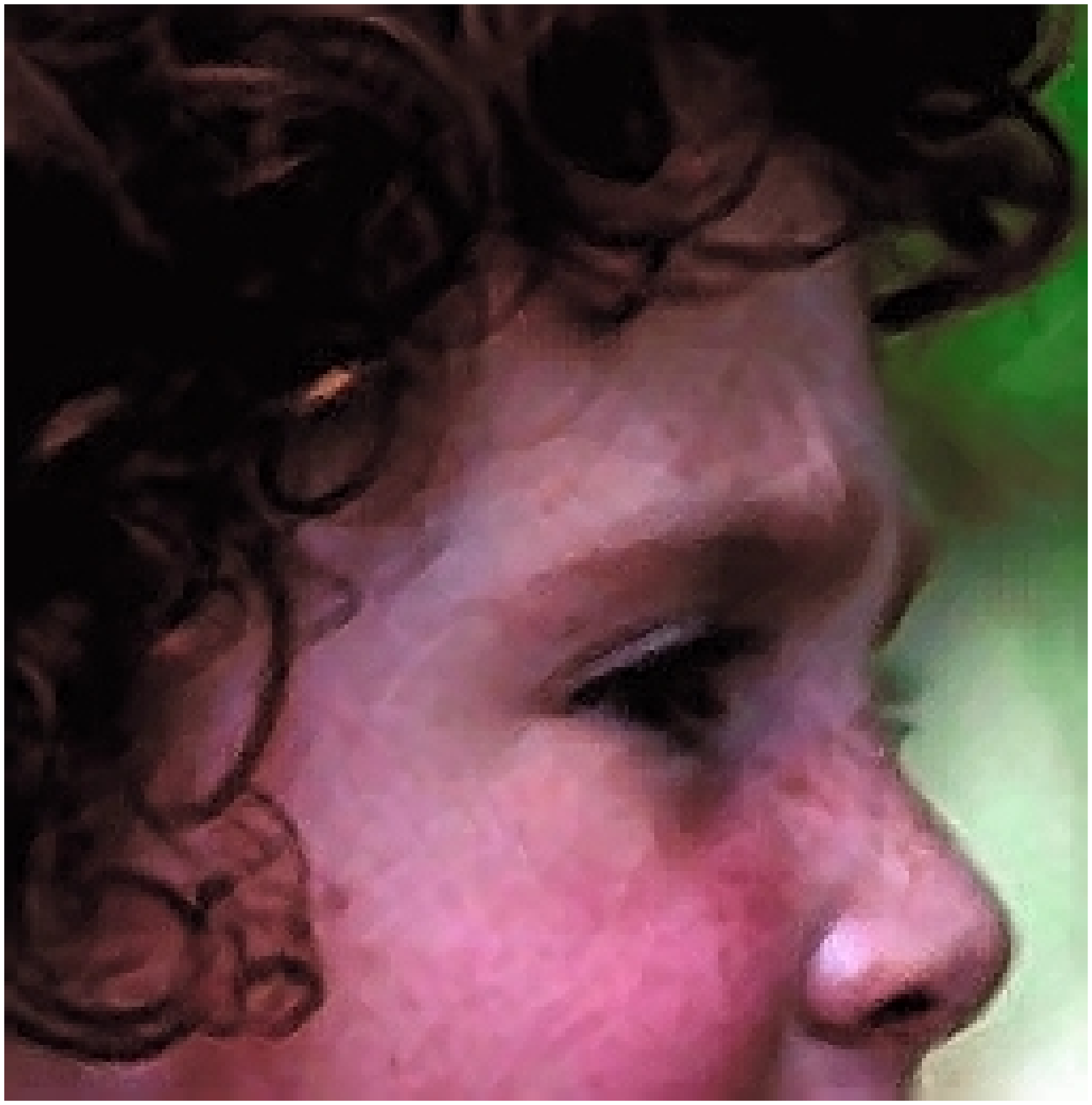}
\put(50,-41){{\includegraphics[viewport = 280 200 380 280, clip, height=1.55cm]{girl_L1SR.eps}}}
\put(0,-41){{\includegraphics[viewport = 450 30 550 110, clip, height=1.55cm]{girl_L1SR.eps}}}
\put(1,1){\sffamily \footnotesize{\textcolor{white}{\textsc{ScSR}}}}
\end{overpic}
\begin{overpic}[height=4.cm]{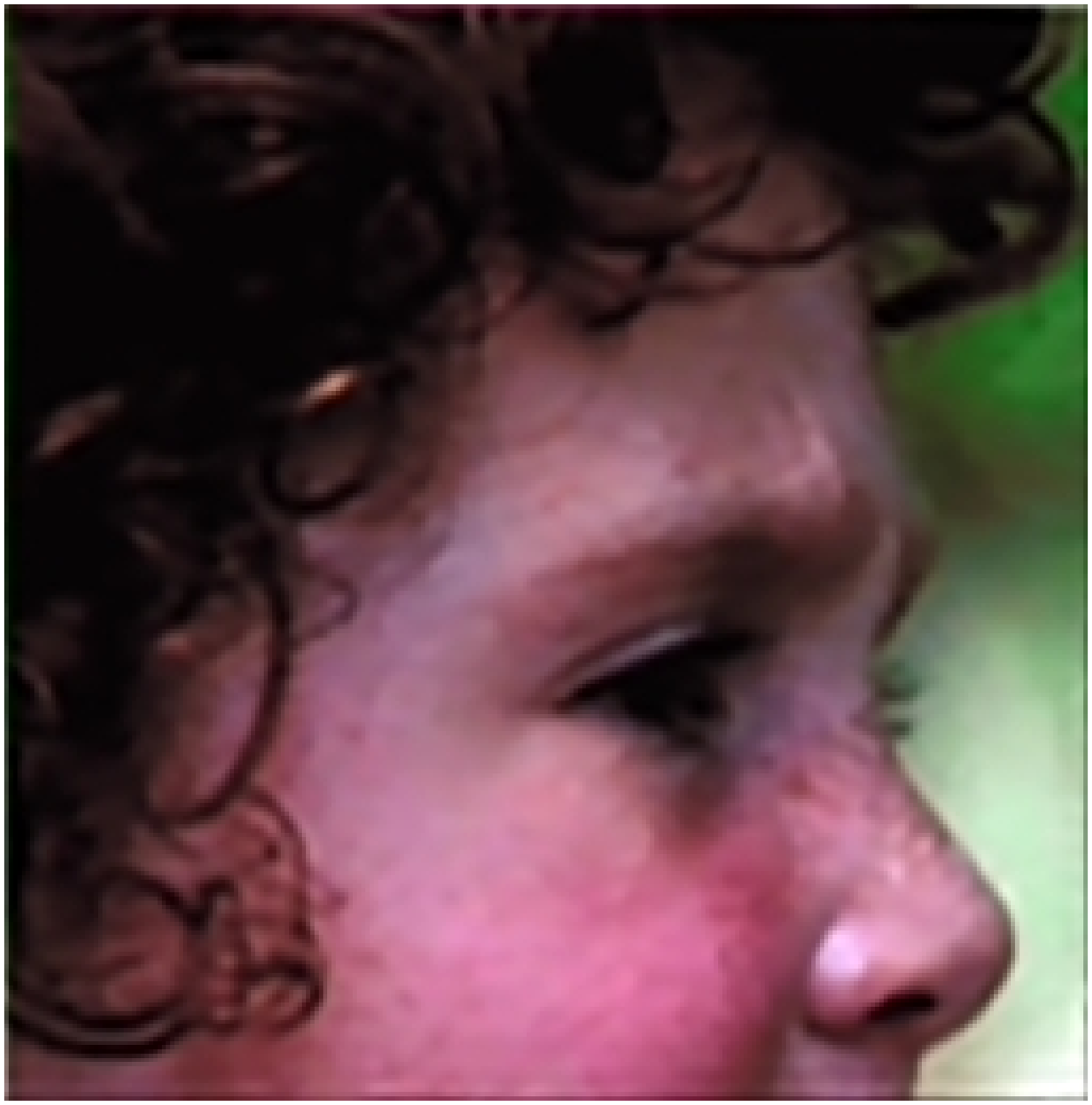}
\put(50,-41){{\includegraphics[viewport = 280 200 380 280, clip, height=1.55cm]{girl_HR_PSNR31.4014_SSIM0.80425.eps}}}
\put(0,-41){{\includegraphics[viewport =450 30 550 110, clip, height=1.55cm]{girl_HR_PSNR31.4014_SSIM0.80425.eps}}}
\put(1,1){\sffamily \footnotesize{\textcolor{white}{\textsc{NBSR}}}}
\end{overpic}
\begin{overpic}[height=4.cm]{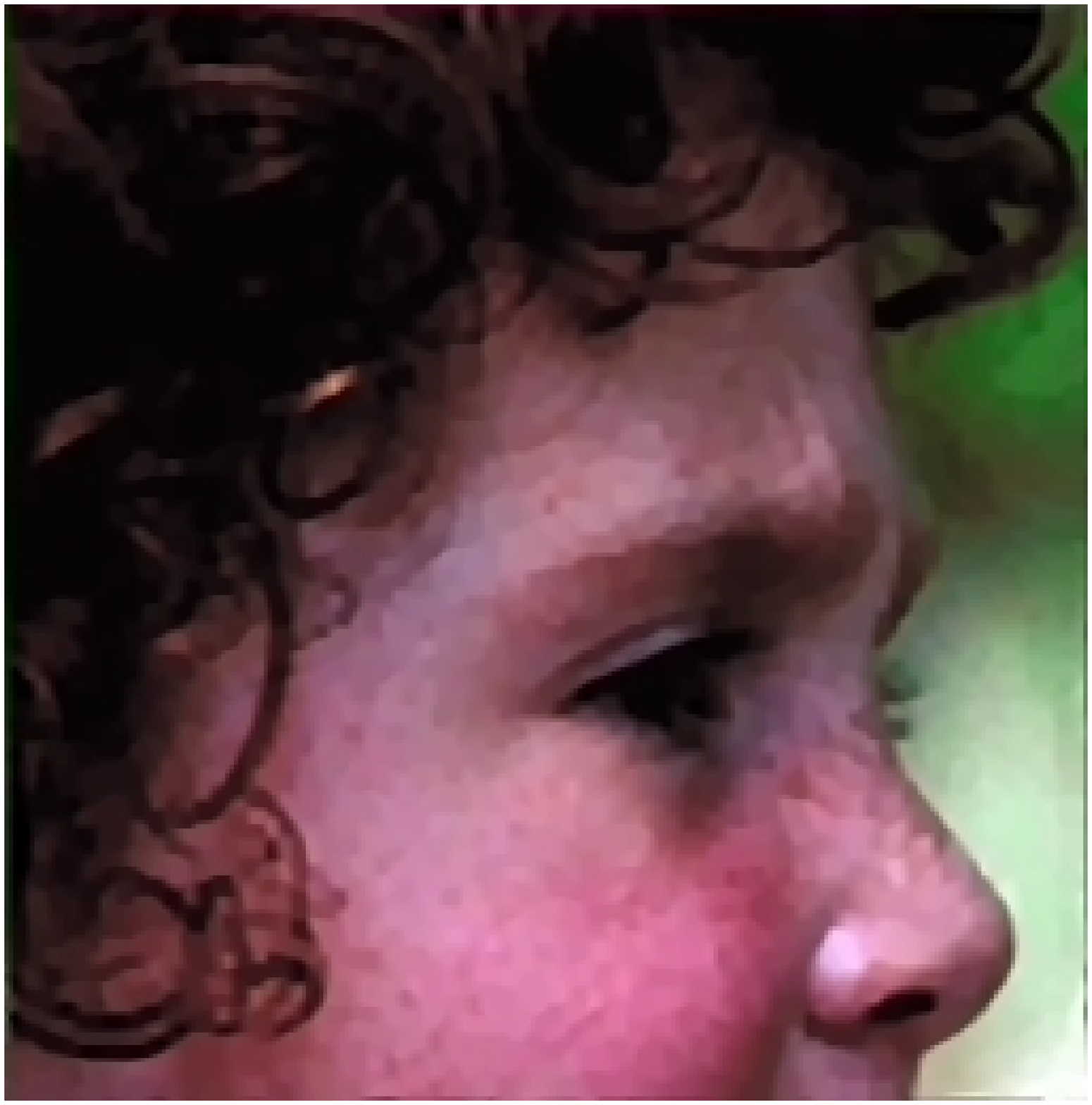}
\put(50,-41){{\includegraphics[viewport = 280 200 380 280, clip, height=1.55cm]{girl_EBSR_HR_PSNR32.467_SSIM0.80907.eps}}}
\put(0,-41){{\includegraphics[viewport = 450 30 550 110, clip, height=1.55cm]{girl_EBSR_HR_PSNR32.467_SSIM0.80907.eps}}}
\put(1,1){\sffamily \footnotesize{\textcolor{white}{\textsc{Proposed}}}}
\end{overpic}\\
\vspace{0.7in}
\begin{overpic}[width=3.96cm]{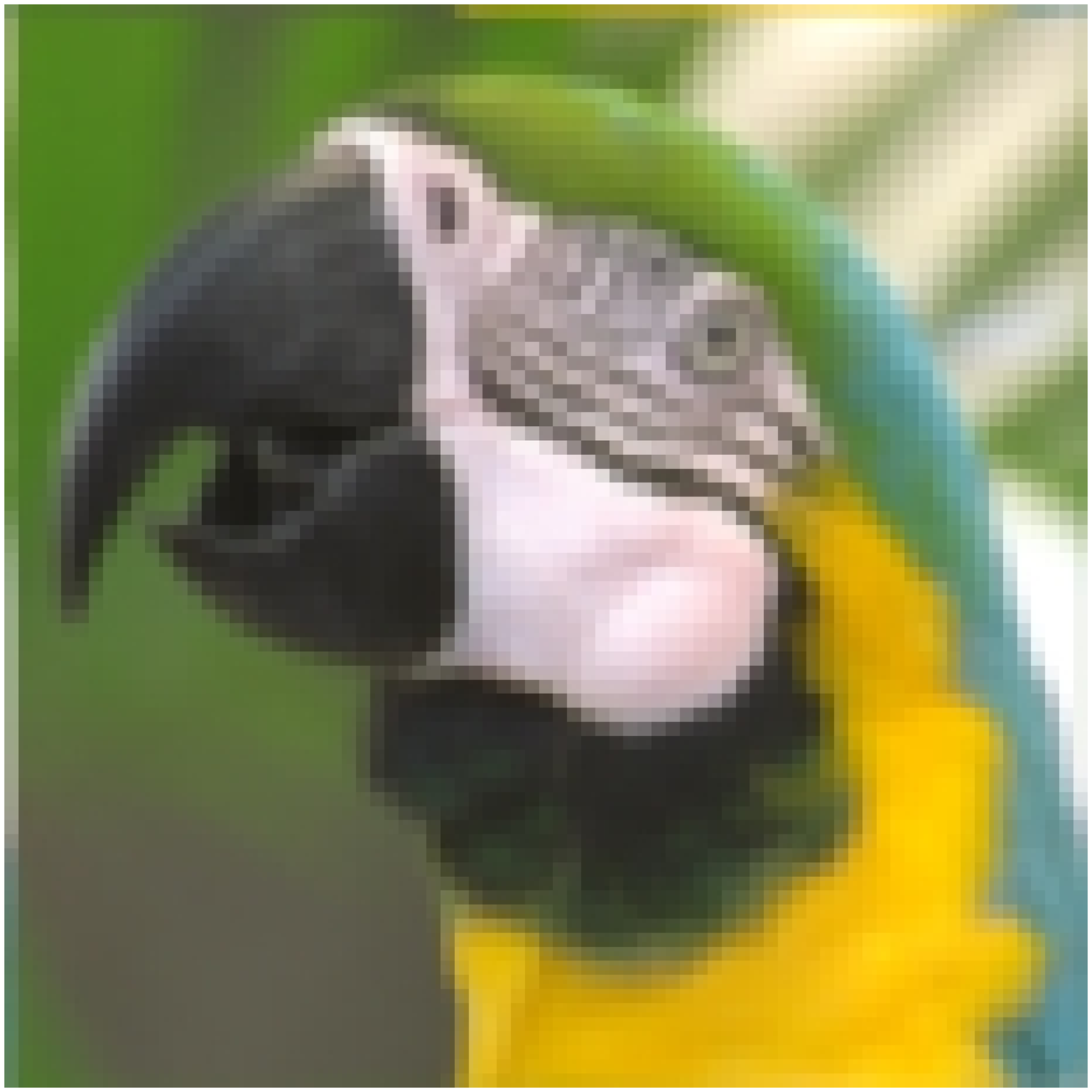}
\put(50,-38){{\includegraphics[viewport = 250 350 410 470, clip, width=1.9cm]{parot_NN_PSNR24.0971_SSIM0.76885.eps}}}
\put(0,-38){{\includegraphics[viewport = 20 270 180 390, clip, width=1.9cm]{parot_NN_PSNR24.0971_SSIM0.76885.eps}}}
\put(1,1){\sffamily \footnotesize{\textcolor{white}{\textsc{Nearest Neighbor}}}}
\end{overpic}
\begin{overpic}[width=3.96cm]{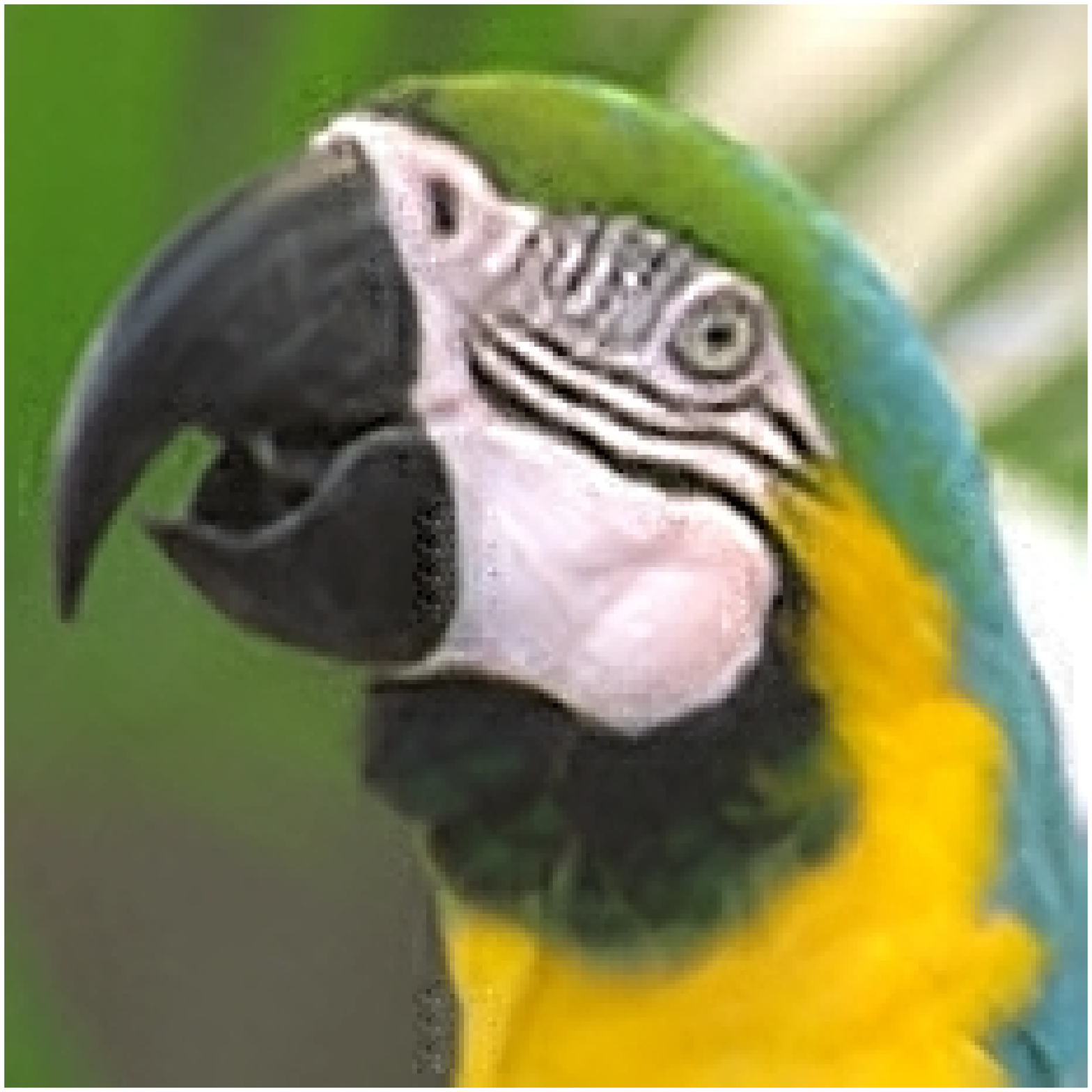}
\put(50,-38){{\includegraphics[viewport = 250 350 410 470, clip, width=1.9cm]{parot_L1SR_PSNR27.885SSIM0.87063.eps}}}
\put(0,-38){{\includegraphics[viewport = 20 270 180 390, clip, width=1.9cm]{parot_L1SR_PSNR27.885SSIM0.87063.eps}}}
\put(1,1){\sffamily \footnotesize{\textcolor{white}{\textsc{ScSR}}}}
\end{overpic}
\begin{overpic}[width=3.96cm]{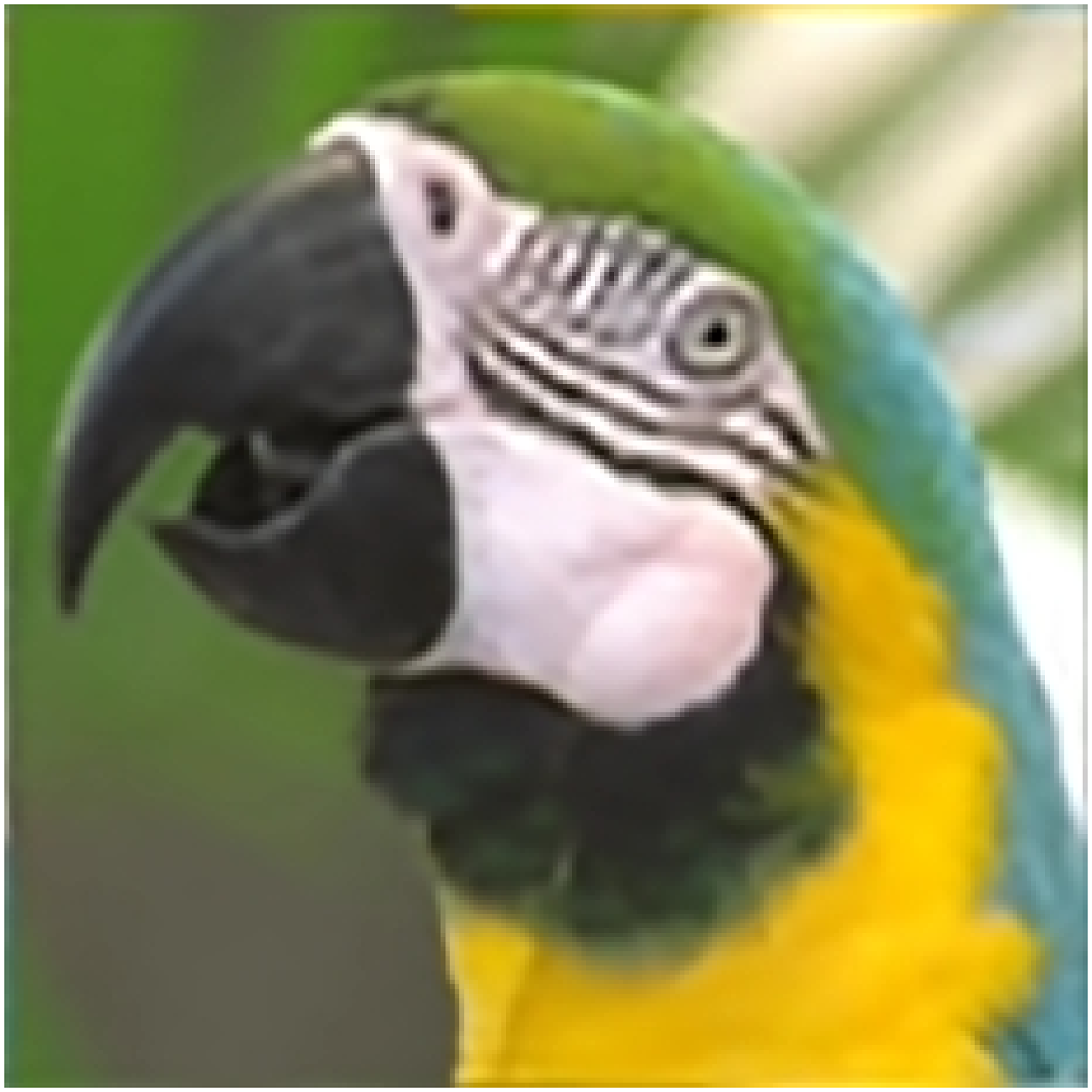}
\put(50,-38){{\includegraphics[viewport = 250 350 410 470, clip, width=1.9cm]{parot_HR_PSNR29.2137_SSIM0.90511.eps}}}
\put(0,-38){{\includegraphics[viewport = 20 270 180 390, clip, width=1.9cm]{parot_HR_PSNR29.2137_SSIM0.90511.eps}}}
\put(1,1){\sffamily \footnotesize{\textcolor{white}{\textsc{NBSR}}}}
\end{overpic}
\begin{overpic}[width=3.96cm]{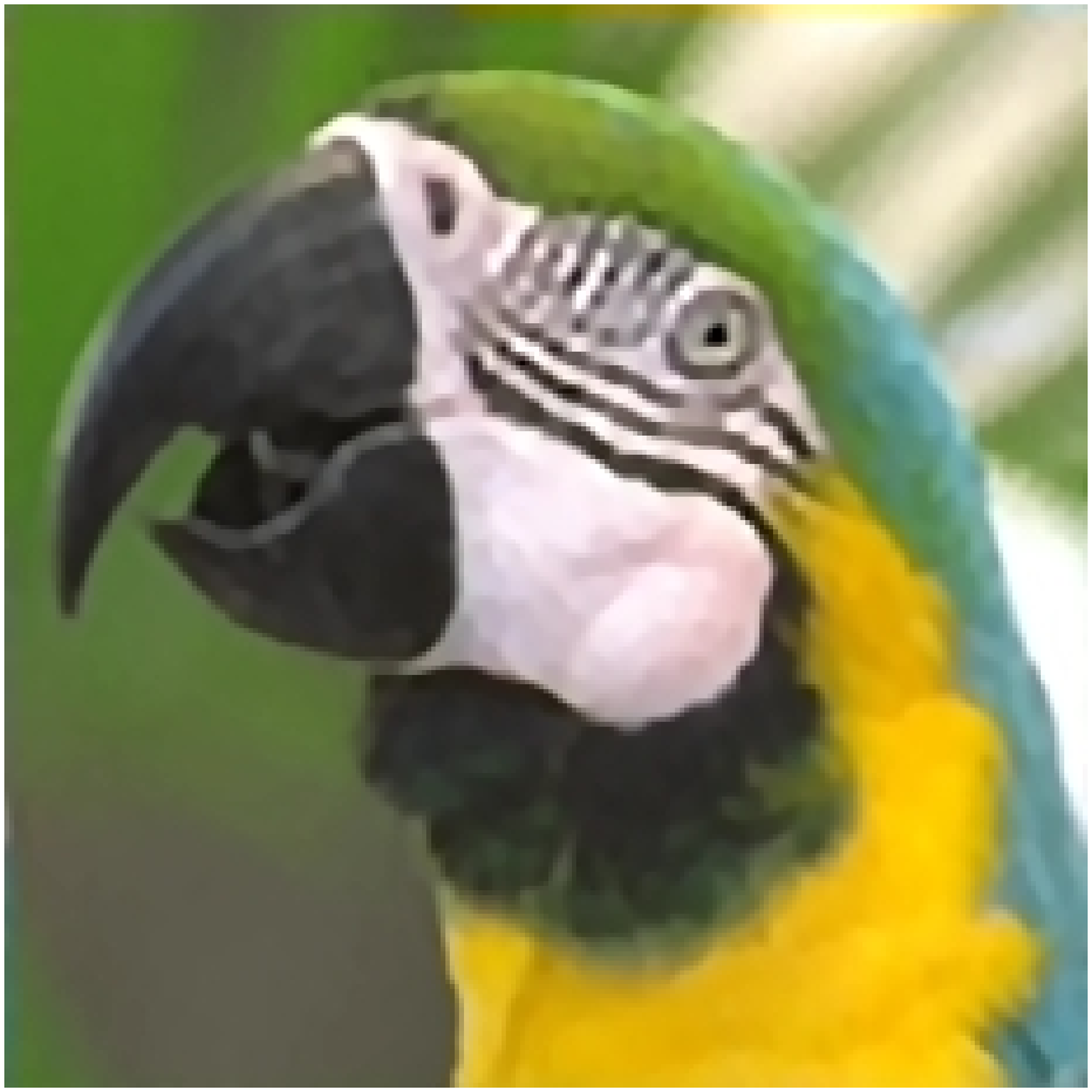}
\put(50,-38){{\includegraphics[viewport = 250 350 410 470, clip, width=1.9cm]{parrot_EBSR_HR_29.477104SSIM0.907581.eps}}}
\put(0,-38){{\includegraphics[viewport = 20 270 180 390, clip, width=1.9cm]{parrot_EBSR_HR_29.477104SSIM0.907581.eps}}}
\put(1,1){\sffamily \footnotesize{\textcolor{white}{\textsc{Proposed}}}}
\end{overpic}\\
\vspace{0.65in}
\begin{overpic}[height=2.05cm]{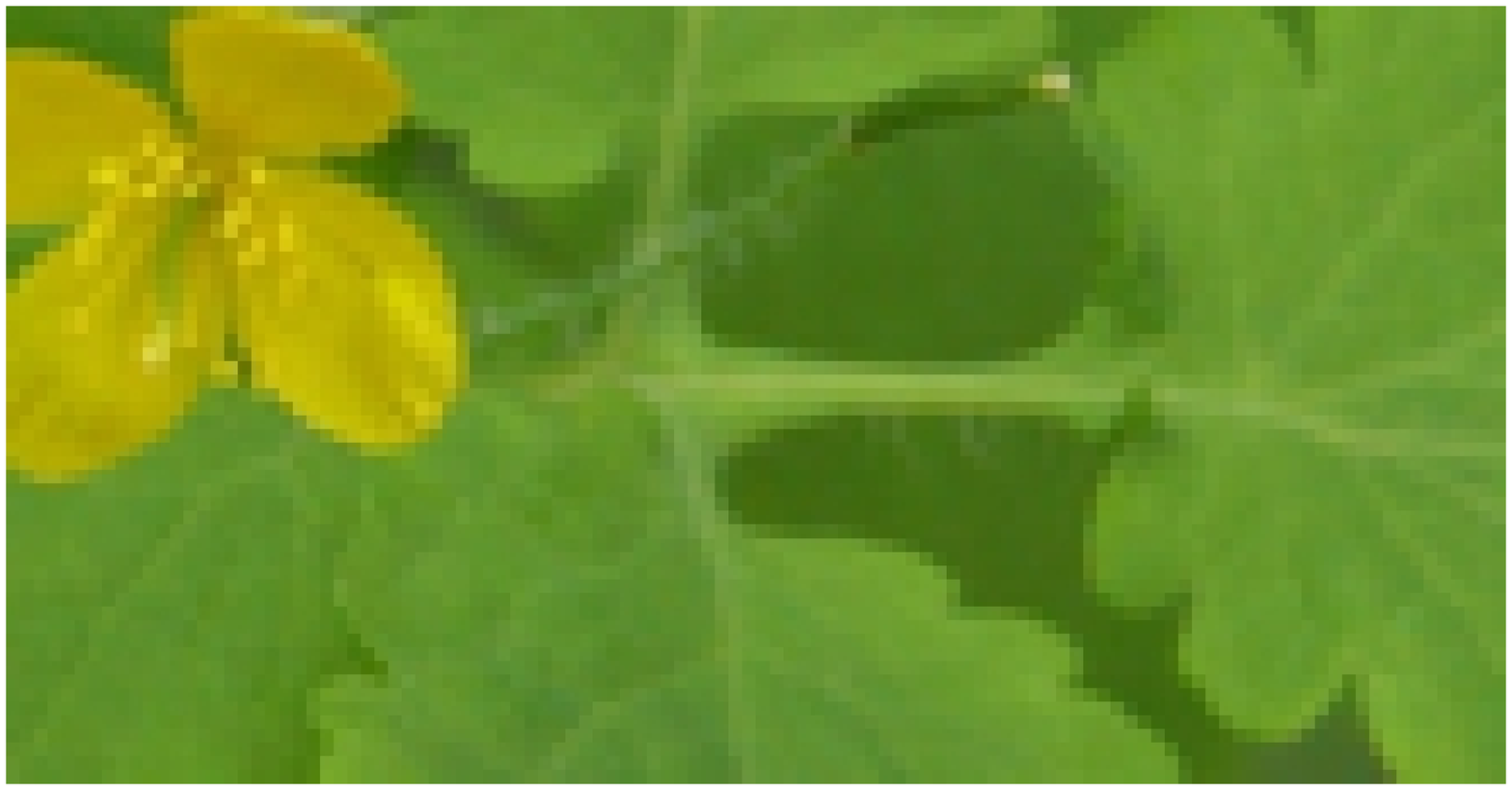}
\put(50,-41){{\includegraphics[viewport = 10 150 110 230, clip, height=1.55cm]{flower_NN.eps}}}
\put(0,-41){{\includegraphics[viewport = 250 100 350 180, clip, height=1.55cm]{flower_NN.eps}}}
\put(1,1){\sffamily \footnotesize{\textcolor{white}{\textsc{Nearest Neighbor}}}}
\end{overpic}
\begin{overpic}[height=2.05cm]{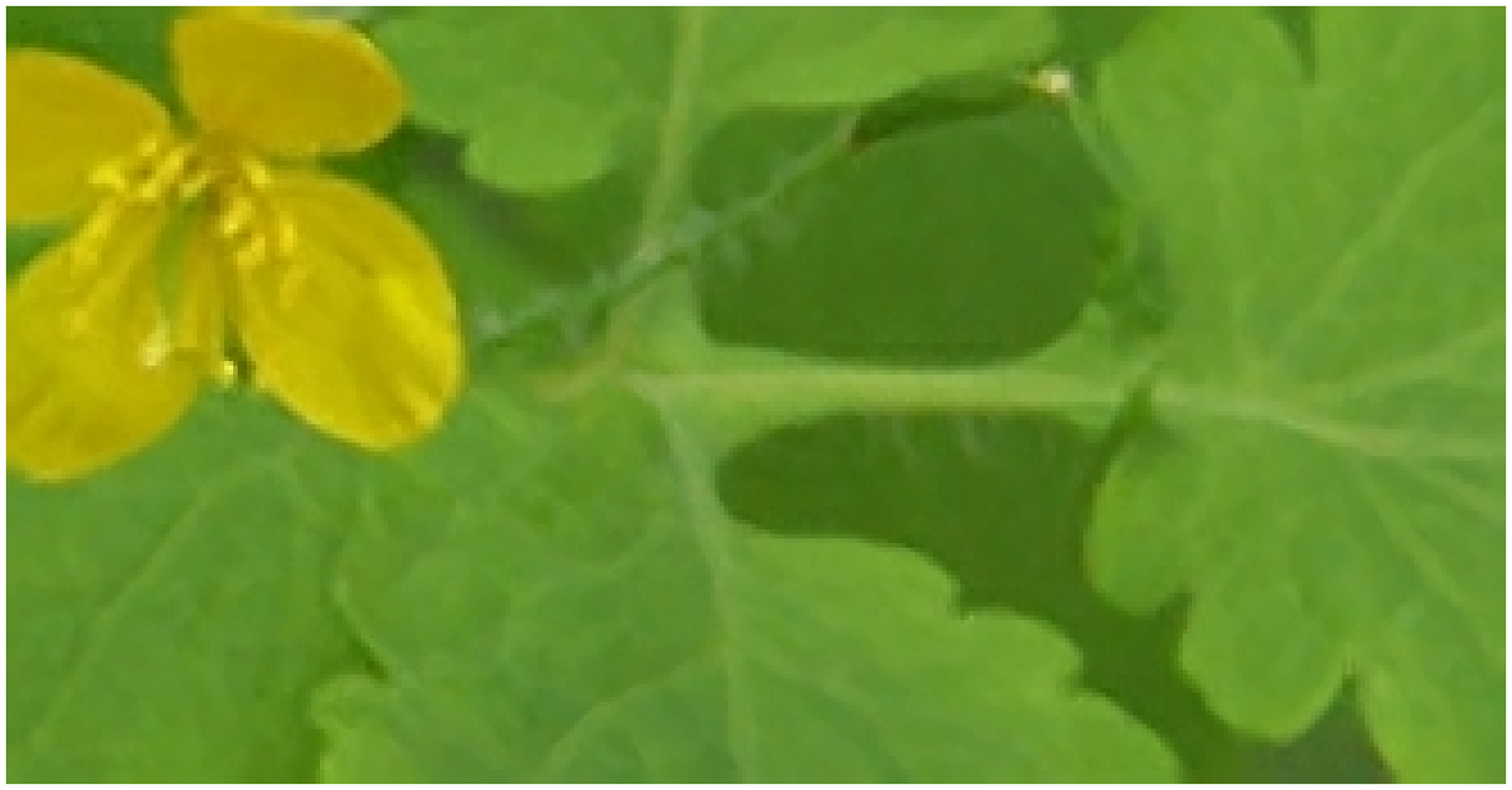}
\put(50,-41){{\includegraphics[viewport = 10 150 110 230, clip, height=1.55cm]{flower_L1SR.eps}}}
\put(0,-41){{\includegraphics[viewport = 250 100 350 180, clip, height=1.5cm]{flower_L1SR.eps}}}
\put(1,1){\sffamily \footnotesize{\textcolor{white}{\textsc{ScSR}}}}
\end{overpic}
\begin{overpic}[height=2.05cm]{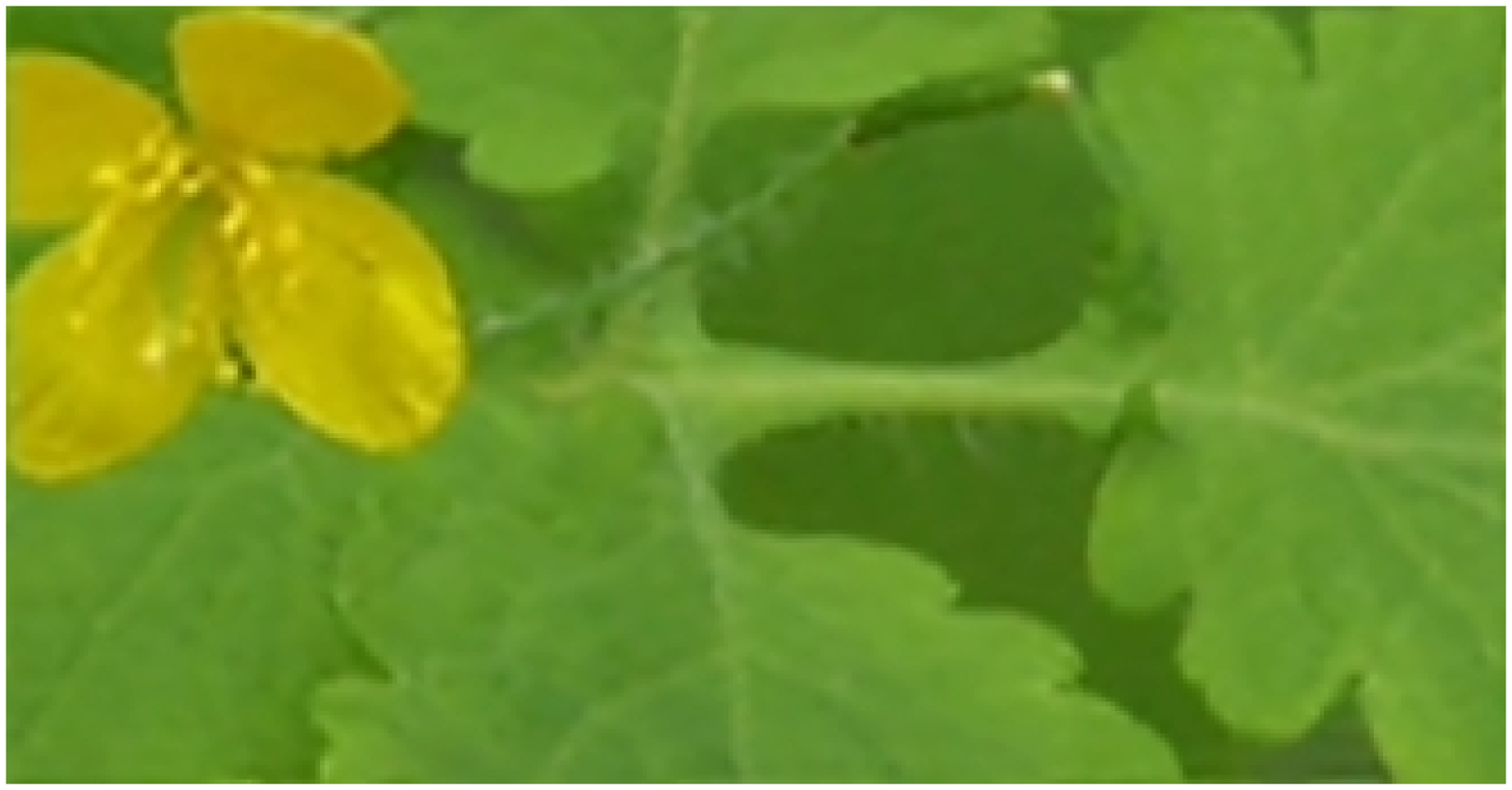}
\put(50,-41){{\includegraphics[viewport = 10 150 110 230, clip, height=1.55cm]{flower_HR_PSNR38.0747_SSIM0.92742.eps}}}
\put(0,-41){{\includegraphics[viewport = 250 100 350 180, clip, height=1.55cm]{flower_HR_PSNR38.0747_SSIM0.92742.eps}}}
\put(1,1){\sffamily \footnotesize{\textcolor{white}{\textsc{NBSR}}}}
\end{overpic}
\begin{overpic}[height=2.05cm]{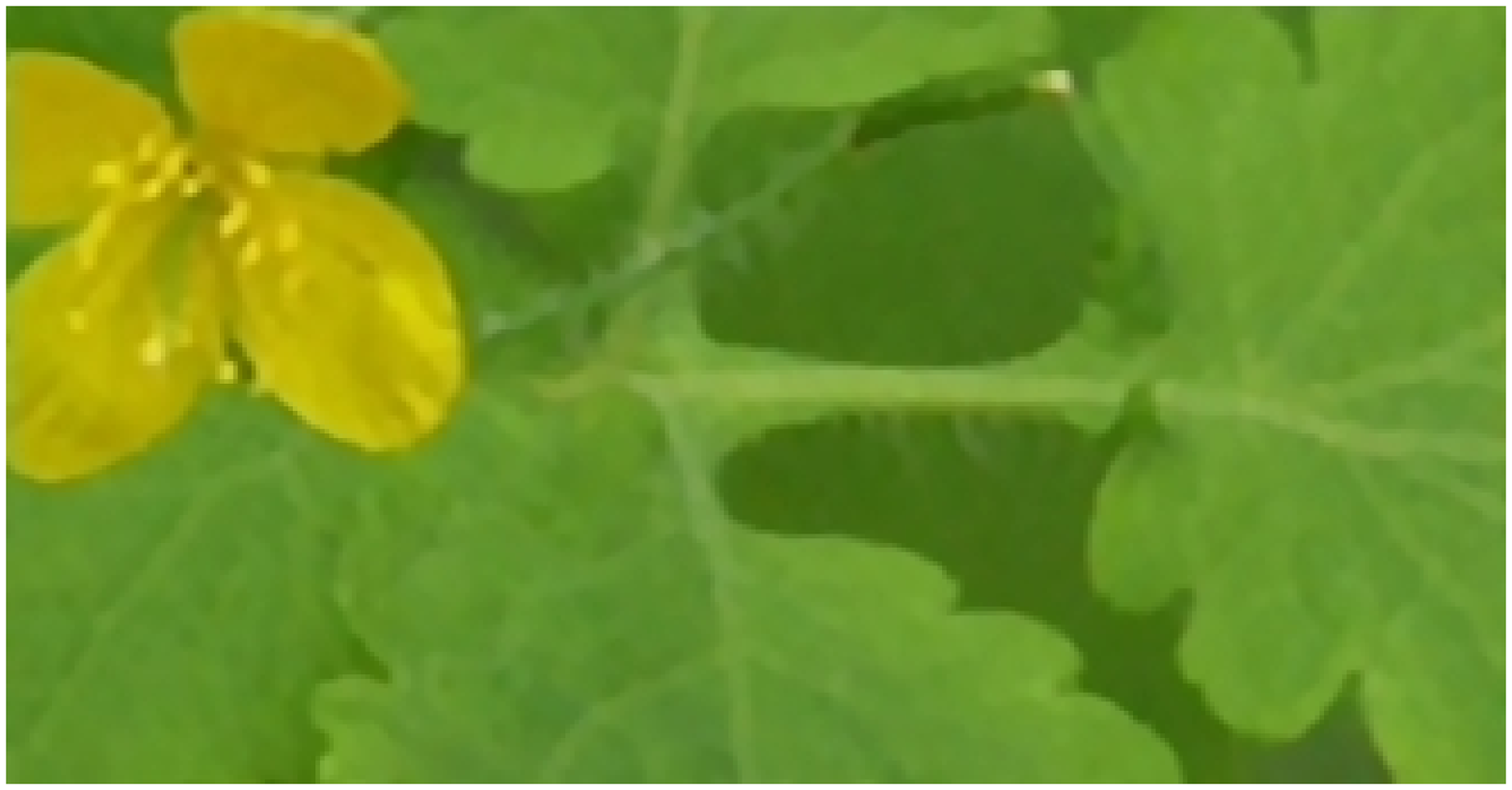}
\put(50,-41){{\includegraphics[viewport = 10 150 110 230, clip, height=1.55cm]{flower_EBSR_HR_PSNR38.302241_SSIM0.927263.eps}}}
\put(0,-41){{\includegraphics[viewport = 250 100 350 180, clip, height=1.55cm]{flower_EBSR_HR_PSNR38.302241_SSIM0.927263.eps}}}
\put(1,1){\sffamily \footnotesize{\textcolor{white}{\textsc{Proposed}}}}
\end{overpic}\\
\vspace{0.5in}
\caption{Single image super resolution ($\times 3$) results  for `{\ttfamily girl}', `{\ttfamily parrot}'  and `{\ttfamily flower}' images ($\times 3$). PSNR and SSIM results for the image `{\ttfamily flower}' are given  in brackets.  Left to right: Nearest Neighbor Interpolation (32.95, 0.849),  Fast super resolution method~\cite{Shan_SR} (35.81, 0.890),  sparse representation based SR method~\cite{Yang_SR_TIP} (37.60, 0.923), the NBSR method~\cite{NBSR_Zhang} ({38.07}, {\bf 0.927}) and the proposed EBSR method (\bf{38.30}, \bf{0.927}).}
\label{fig:Res_flower}
\end{figure*}

We further evaluate the proposed SR method on the task of color image SR.
For color image SR, we only perform SR  for the intensity channel in the YCbCr color space and use bicubic interpolation for the other two channels.
The reason is that  human eyes are more sensitive to the intensity changes of the image than to the changes in other  channels  such as saturation and hue. This scheme has been practised already in the previous SR works~\cite{Yang_SR_TIP}.
The SR  results  on color images  with the zooming factor of $3$  are shown in Figure~\ref{fig:Res_flower}.
The PSNR and SSIM results accompanying these images are also reported.
As can be seen from Figure~\ref{fig:Res_flower}, the proposed method can generate  SR images visually comparable to the sparse representation based method, which are much better than the fast SR method.
Detailed inspections also reveal that the SR results from the proposed method have less artifacts than those from the ScSR method (see the zoomed patches for the \emph{nose} and \emph{eye} parts of the `{\ttfamily girl}' image in Figure~\ref{fig:Res_flower}), and appear sharper than than the results from the NBSR method.
Furthermore, the quantitative evaluation also indicates that the SR results from the proposed method have better quality in terms of  PSNR and SSIM.

\begin{table}
\begin{center}
\caption{Noisy image super resolution  ($\times 3$) result comparison of estimation quality.}
\addtolength{\tabcolsep}{-2pt}\renewcommand{\arraystretch}{1}
\label{table:SR_Noisy}
\begin{tabular}{|c|c|c|cccccc|}
\hline\noalign{}
$\sigma$&\multicolumn{2}{|c|}{Methods} &  NN & BI & Fast~\cite{Shan_SR} & ScSR~\cite{Yang_SR_TIP} & NBSR~\cite{NBSR_Zhang}& Proposed\\
\hline\hline
\multirow{6}{*}{0}&\multirow{2}{*}{house}&PSNR &  24.90  & 25.62 & 28.72 & 30.80 &   {32.06}& 32.85 \\
&&SSIM  & 0.763  &  0.788 &  0.837& 0.863 & {0.897} & 0.897\\
\cline{2-9}
&\multirow{2}{*}{peppers}&PSNR &  21.59   &   22.18 & 24.38 & 25.52 &    {25.88}&  26.88\\
&&SSIM  & 0.727  & 0.775  & 0.841&  0.863 &  {0.910} & 0.918\\
\cline{2-9}
&\multirow{2}{*}{cameraman}&PSNR &  21.71   &   22.12 & 24.64 & 25.79 &26.42  & {26.65}\\
&&SSIM  & 0.695  & 0.714  &  0.774& 0.810 & 0.847 &{0.846} \\
\hline
\multirow{6}{*}{1}&\multirow{2}{*}{house}&PSNR &  24.88  & 25.60 &28.88 & 30.67 &   {31.24} & 32.13\\
&&SSIM  & 0.757  &  0.784 &  0.844 &  0.854 & {0.876} & 0.879\\
\cline{2-9}
&\multirow{2}{*}{peppers}&PSNR &  21.58   &   22.17  & 24.53 & 25.49 &    {25.65} & 26.37\\
&&SSIM  & 0.723  & 0.772  &  0.851 & 0.858 &  {0.886} & 0.894\\
\cline{2-9}
&\multirow{2}{*}{cameraman}&PSNR &  21.70   &   22.15 & 24.73 & 25.75 &   {26.09} & 26.32\\
&&SSIM  & 0.689  & 0.710  &  0.778&  0.801 &  {0.823} & 0.828\\
\hline
\multirow{6}{*}{2}&\multirow{2}{*}{house}&PSNR &  24.81  & 25.55  & 28.76 & {30.47} &   {30.47}& 31.33\\
&&SSIM  & 0.738  &  0.773 & 0.840 &  0.834 & {0.858}& 0.864\\
\cline{2-9}
&\multirow{2}{*}{peppers}&PSNR &  21.55   &   22.15 & 24.49 & {25.36} &    25.23 & 25.84\\
&&SSIM  & 0.710  & 0.764  &  0.847 & 0.844 &  {0.860} &0.871\\
\cline{2-9}
&\multirow{2}{*}{cameraman}&PSNR & 21.66    & 22.13 & 24.70   & 25.63  & {25.67} & 25.92 \\
&&SSIM  & 0.671  &  0.700  &0.775 &   0.779   &  {0.802} & 0.809\\
\hline
\multirow{6}{*}{4}&\multirow{2}{*}{house}&PSNR &  24.57  & 25.36 & 28.41 & {29.38} &   {29.33} & 30.07 \\
&&SSIM  & 0.678  &  0.734 & 0.823 &  0.798 & {0.832} & 0.842\\
\cline{2-9}
&\multirow{2}{*}{peppers}&PSNR &  21.44   &   22.06& 24.29  & {24.65} &   24.54 &25.01 \\
&&SSIM  & 0.667  & 0.736  & {0.830} &  0.803 & {0.823}& 0.834\\
\cline{2-9}
&\multirow{2}{*}{cameraman}&PSNR &  21.54   &   22.03& 24.52  & 24.88 &  {24.94} &25.18 \\
&&SSIM  & 0.612  & 0.662  &  0.757 &  0.734 & {0.770} &0.778\\
\hline
\end{tabular}
\end{center}
\end{table}

\begin{figure*}[t]
\centering
\begin{overpic}[height=5.3cm]{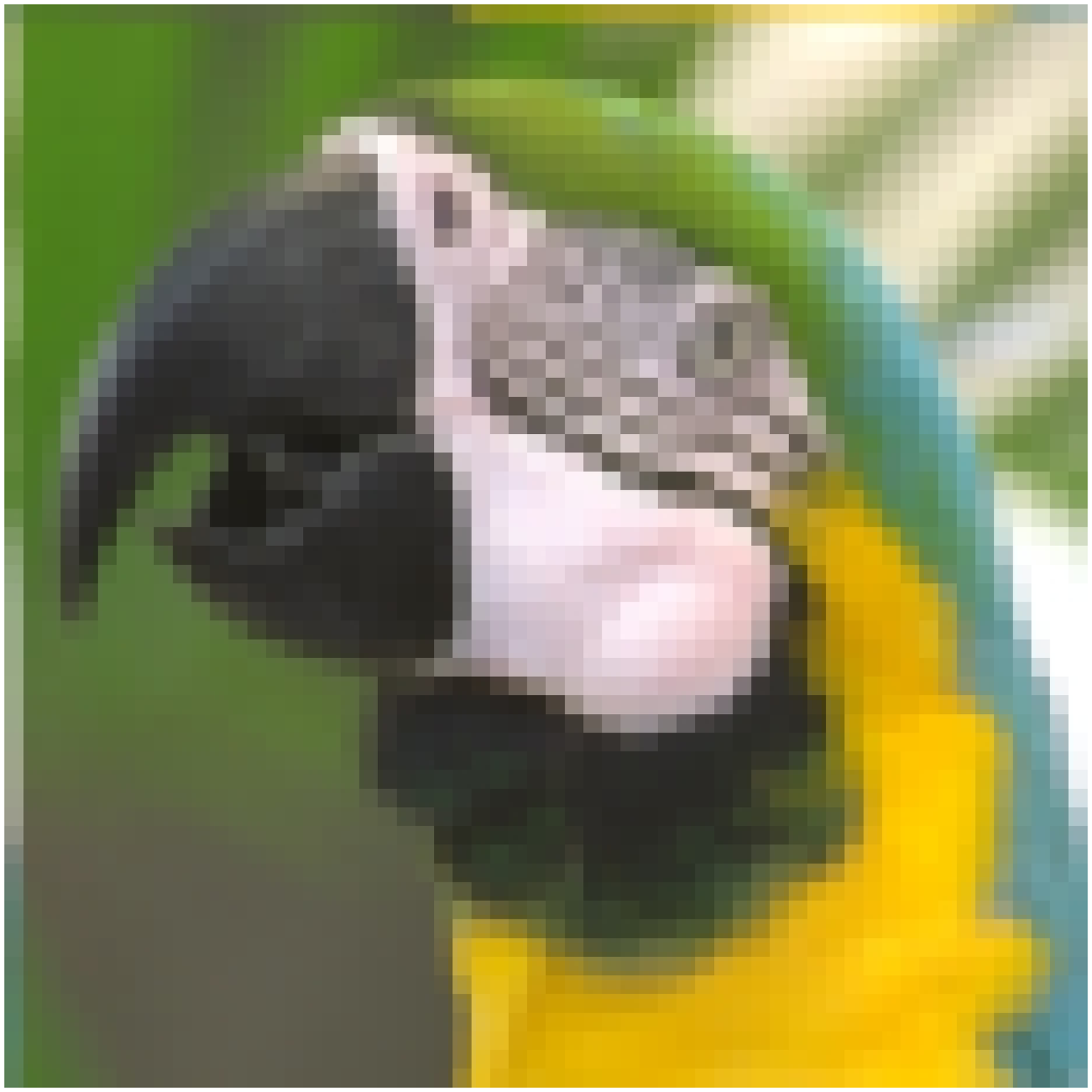}
\put(51,-31){{\includegraphics[viewport = 200 350 400 470, clip, height=1.55cm]{R4_Parot_NN_PSNR22.6365_SSIM0.69921.eps}}}
\put(0,-31){{\includegraphics[viewport = 70 250 270 370, clip, height=1.55cm]{R4_Parot_NN_PSNR22.6365_SSIM0.69921.eps}}}
\put(1,1){ \sffamily \footnotesize{\textcolor{white}{\textsc{Nearest Neighbor}}}}
\end{overpic}
\begin{overpic}[height=5.3cm]{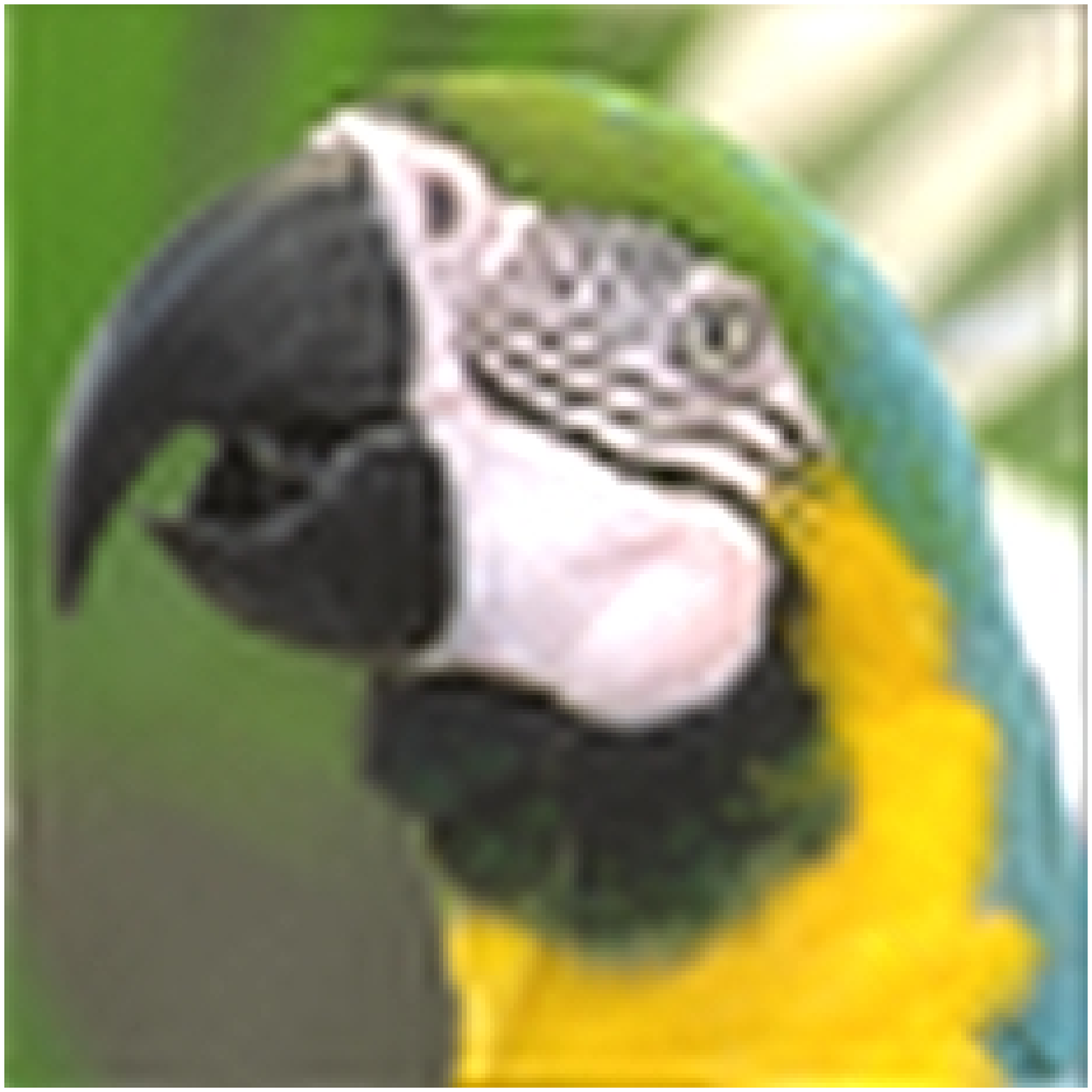}
\put(51,-31){{\includegraphics[viewport = 200 350 400 470, clip, height=1.55cm]{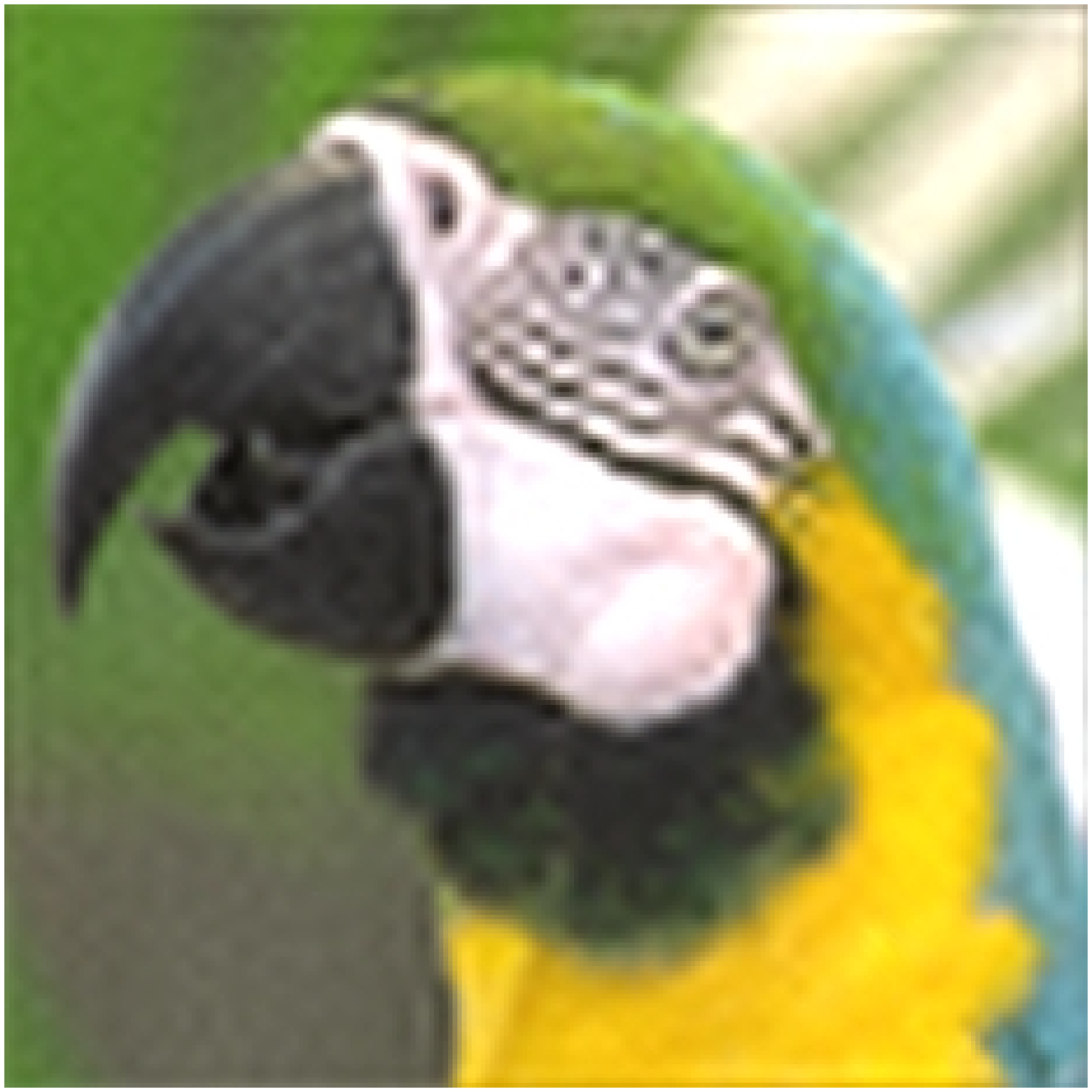}}}
\put(0,-31){{\includegraphics[viewport = 70 250 270 370, clip, height=1.55cm]{R4_Parot_BSR_parot_25.35_SSIM0.8186.eps}}}
\put(1,1){\sffamily \footnotesize{\textcolor{white}{\textsc{VBSR}}}}
\end{overpic}
\vspace{0.70in}
\begin{overpic}[height=5.3cm]{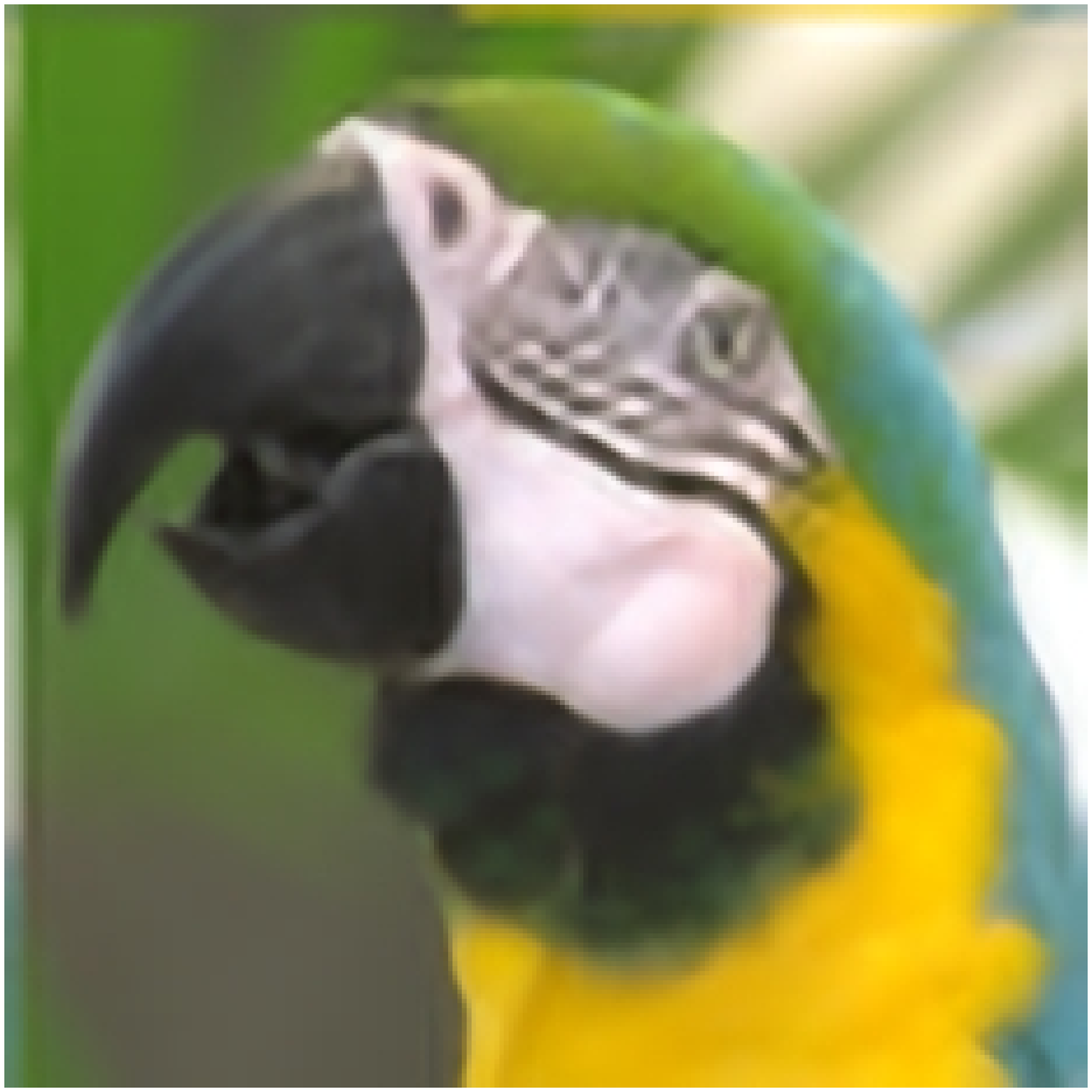}
\put(51,-31){{\includegraphics[viewport = 200 350 400 470, clip, height=1.55cm]{R4_Parot_Kim_PSNR24.223_SSIM0.81824.eps}}}
\put(0,-31){{\includegraphics[viewport = 70 250 270 370, clip, height=1.55cm]{R4_Parot_Kim_PSNR24.223_SSIM0.81824.eps}}}
\put(1,1){\sffamily \footnotesize{\textcolor{white}{\textsc{KRR}}}}
\end{overpic}\\
\begin{overpic}[height=5.3cm]{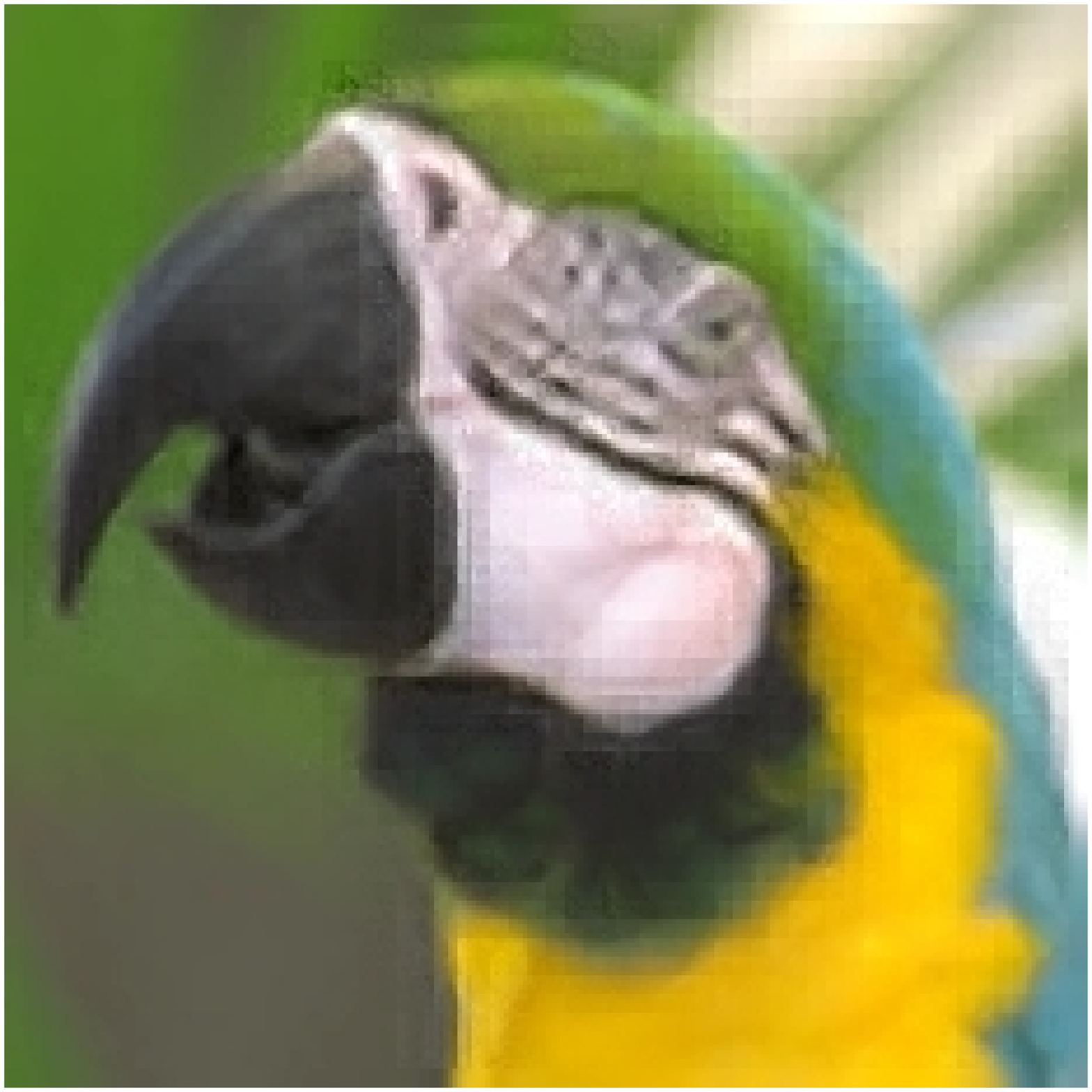}
\put(51,-31){{\includegraphics[viewport = 200 350 400 470, clip, height=1.55cm]{R4_Parot_R4_L1SR_PSNR25.4727SSIM0.79789.eps}}}
\put(0,-31){{\includegraphics[viewport = 70 250 270 370, clip, height=1.55cm]{R4_Parot_R4_L1SR_PSNR25.4727SSIM0.79789.eps}}}
\put(1,1){\sffamily \footnotesize{\textcolor{white}{\textsc{ScSR}}}}
\end{overpic}
\begin{overpic}[height=5.3cm]{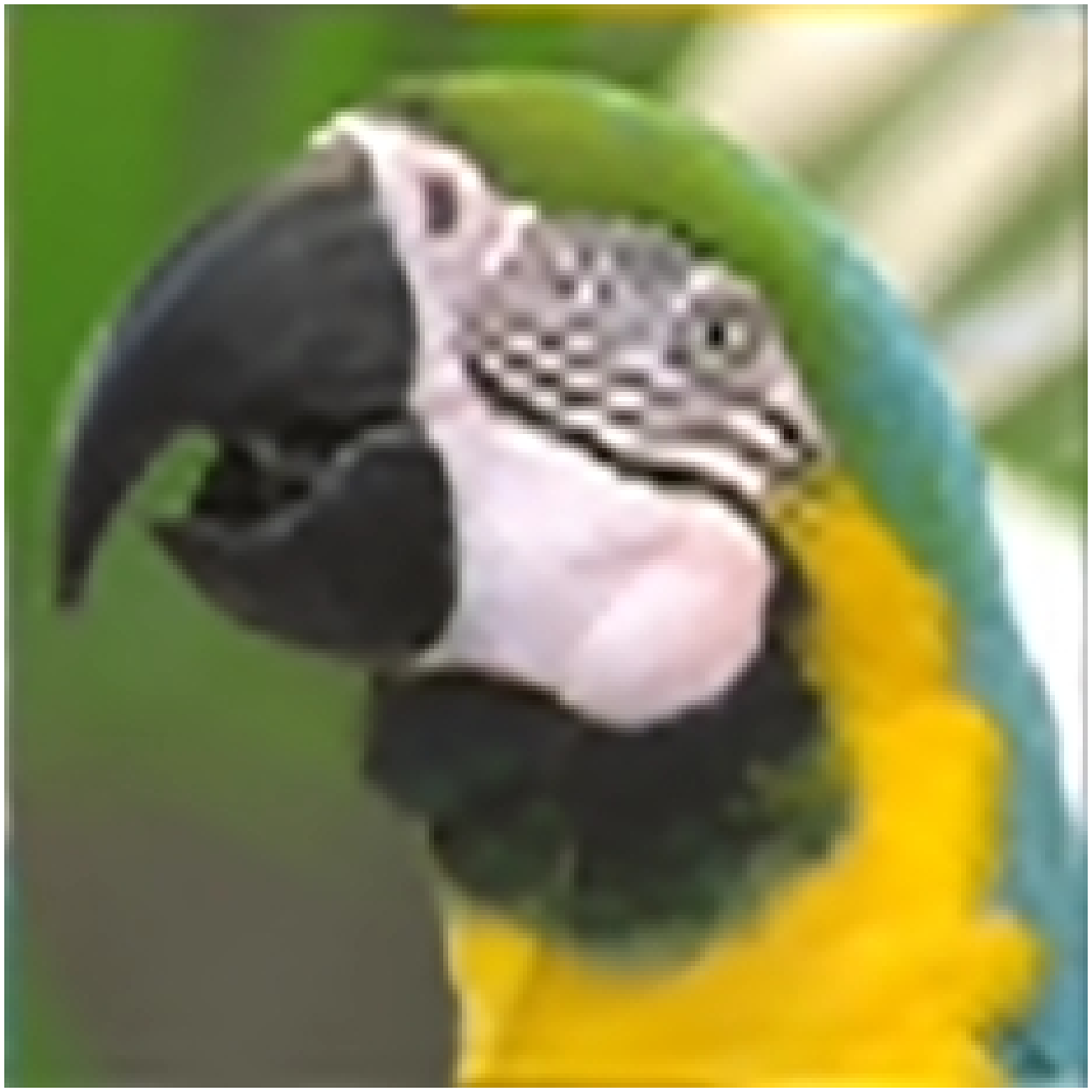}
\put(51,-31){{\includegraphics[viewport = 200 350 400 470, clip, height=1.55cm]{R4_Parot_HR_PSNR26.2238_SSIM0.84708.eps}}}
\put(0,-31){{\includegraphics[viewport = 70 250 270 370, clip, height=1.55cm]{R4_Parot_HR_PSNR26.2238_SSIM0.84708.eps}}}
\put(1,1){\sffamily \footnotesize{\textcolor{white}{\textsc{NBSR}}}}
\end{overpic}
\begin{overpic}[height=5.3cm]{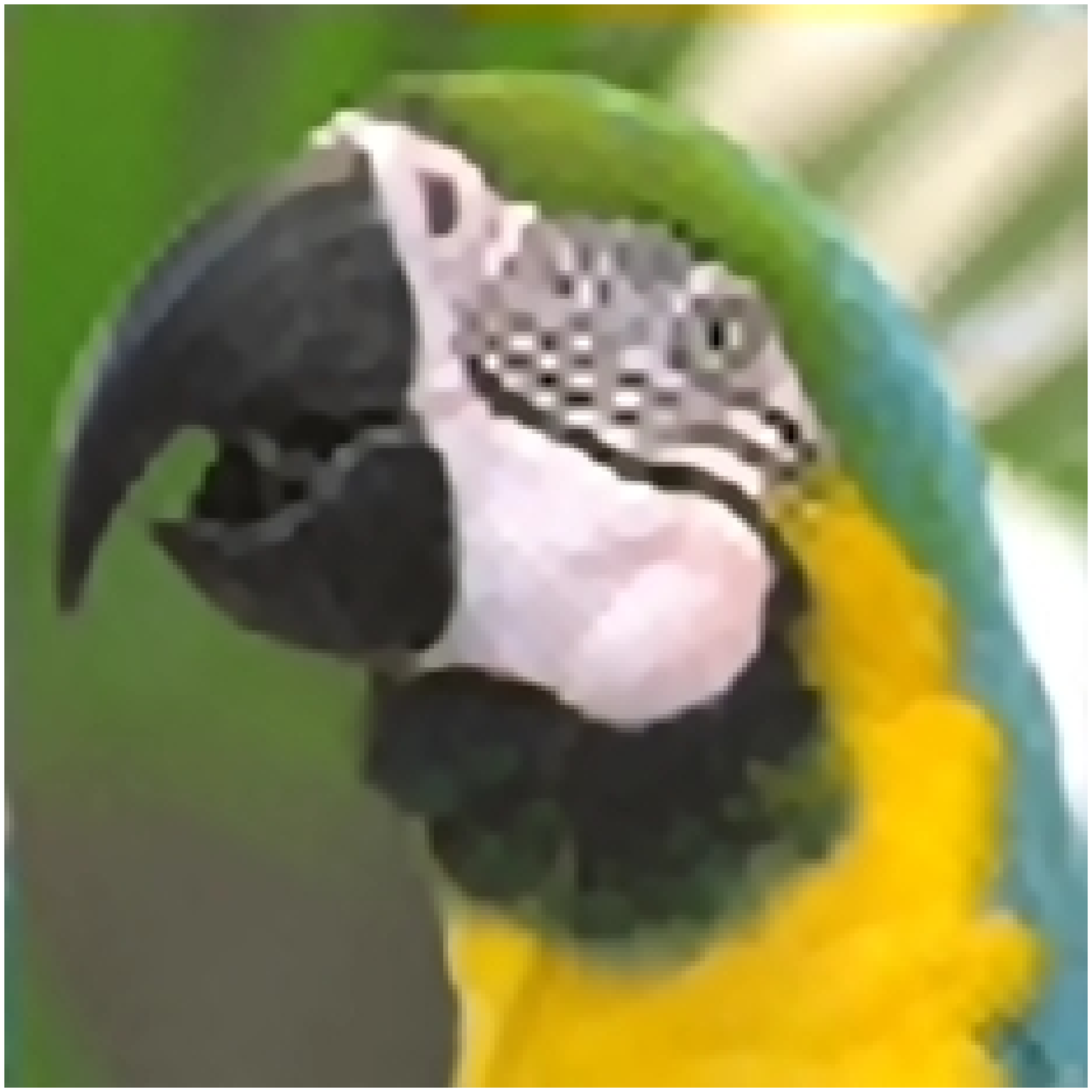} 
\put(51,-31){{\includegraphics[viewport = 200 350 400 470, clip, height=1.55cm]{parrot_EBSR_r4_HR_PSNR26.467522SSIM0.84877.eps}}}
\put(0,-31){{\includegraphics[viewport = 70 250 270 370, clip, height=1.55cm]{parrot_EBSR_r4_HR_PSNR26.467522SSIM0.84877.eps}}}
\put(1,1){\sffamily \footnotesize{\textcolor{white}{Proposed}}}
\end{overpic}
\vspace{0.5in}
\caption{Single image super resolution results with different algorithms ($\times 4$).
The results are generated with: Nearest Neighbor interpolation (22.64, 0.699), Shan \emph{et al.}'s Fast SR method~\cite{Shan_SR} (24.09, 0.775), VBSR~\cite{VBSR} (23.33, 0.823), Kim \emph{et al.}'s kernel regression based method (24.22, 0.818)~\cite{Kim_SR_PAMI}, Yang \emph{et al.}'s ScSR method (25.47, 0.798)~\cite{Yang_SR_TIP}, Zhang \emph{et al.}'s NBSR method~\cite{NBSR_Zhang} ({26.22}, {0.847}) and the proposed EBSR method~(\bf{26.47}, \bf{0.849}).}
\label{fig:Res_SR_color_VBSR}
\end{figure*}

\begin{figure*}[ht]
\centering
\begin{overpic}[height=4.cm]{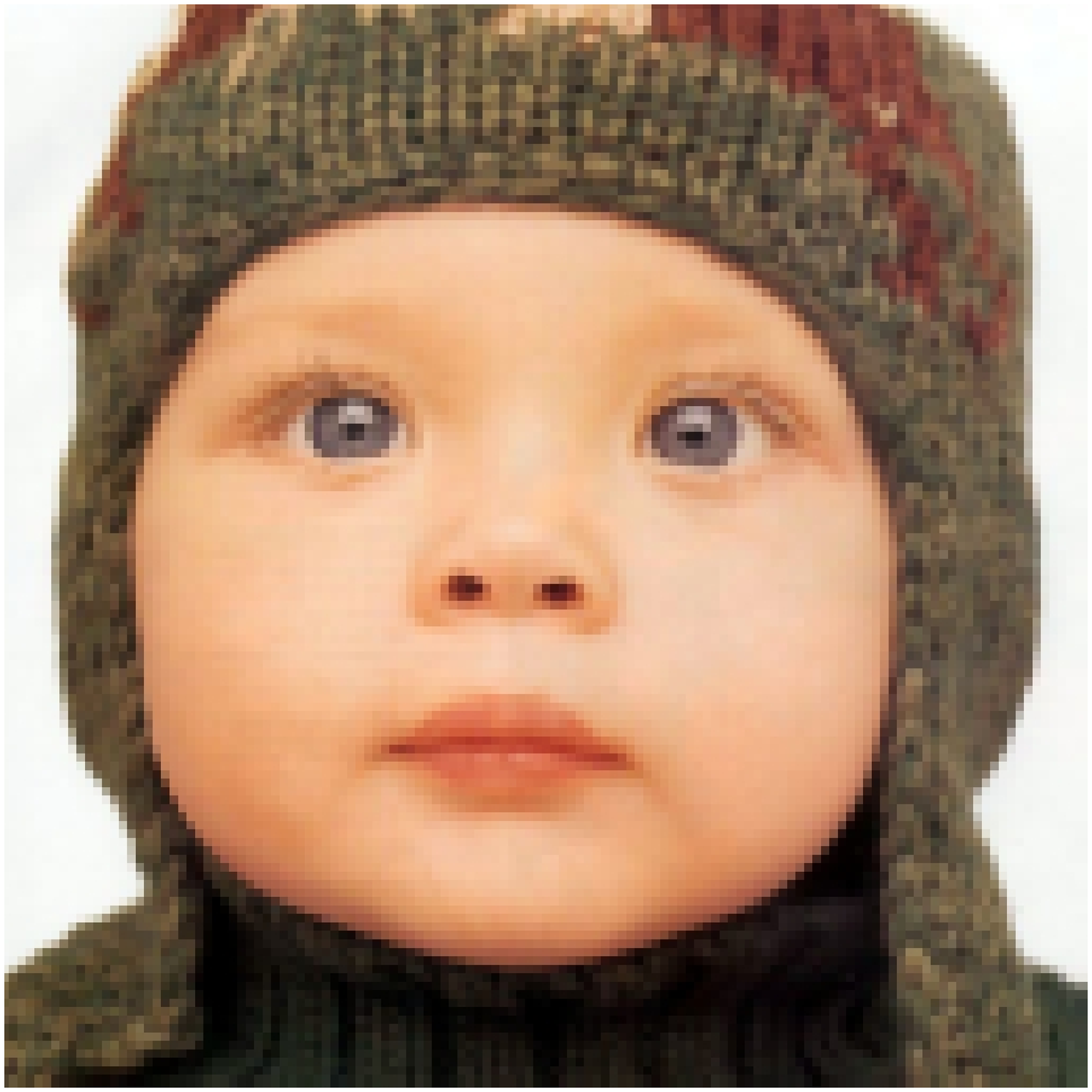}
\put(51,-41){{\includegraphics[viewport = 120 320 220 400, clip, height=1.55cm]{Child_NN.eps}}}
\put(0,-41){{\includegraphics[viewport = 220 450 320 530, clip, height=1.55cm]{Child_NN.eps}}}
\put(1,1){ \sffamily \footnotesize{\textcolor{white}{\textsc{Nearest Neighbor}}}}
\end{overpic}
\begin{overpic}[height=4.cm]{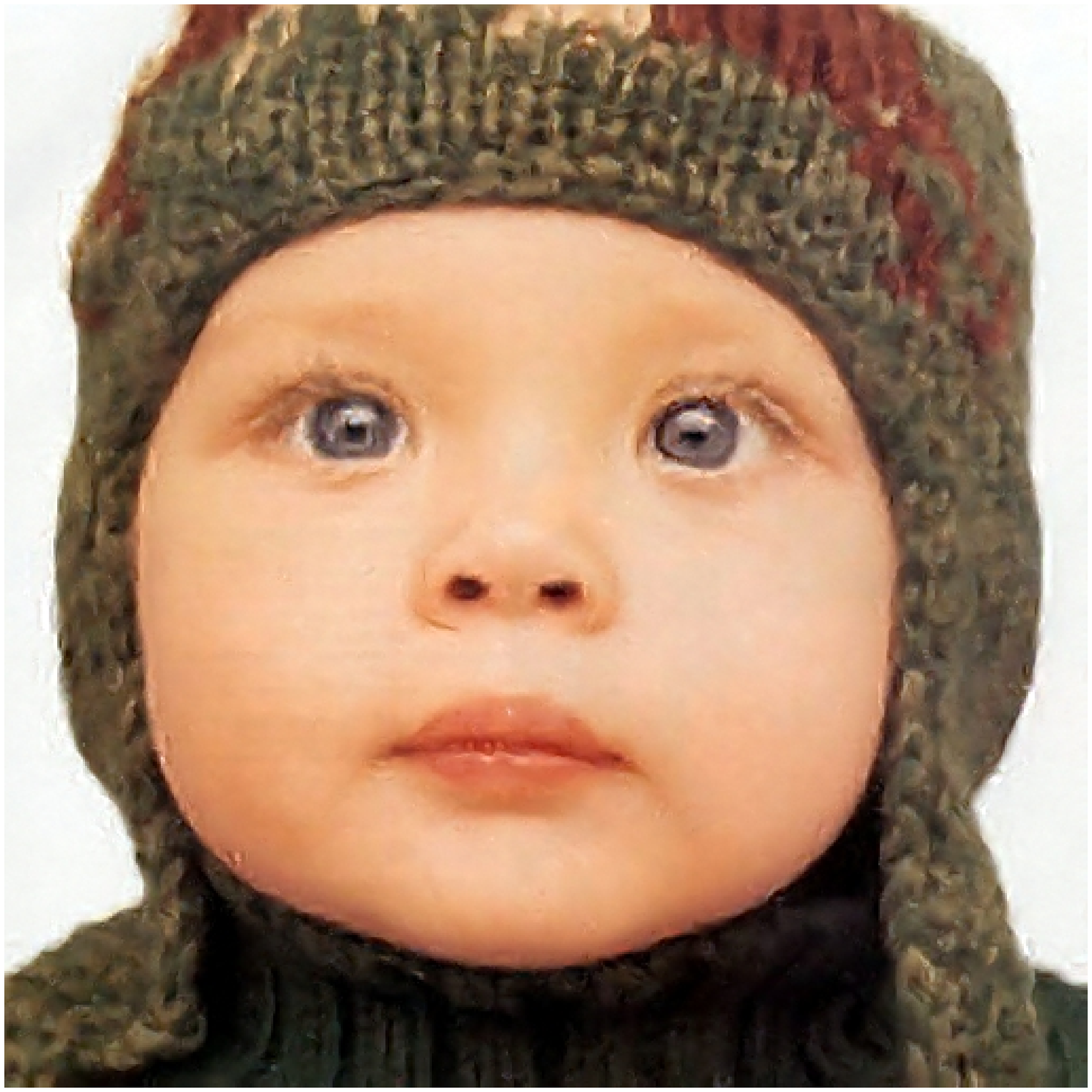}
\put(51,-41){{\includegraphics[viewport = 120 320 220 400, clip, height=1.55cm]{Child_Freeman.eps}}}
\put(0,-41){{\includegraphics[viewport = 220 450 320 530, clip, height=1.55cm]{Child_Freeman.eps}}}
\put(1,1){\sffamily \footnotesize{\textcolor{white}{Freeman \emph{et al.}'s}}}
\end{overpic}
\begin{overpic}[height=4.cm]{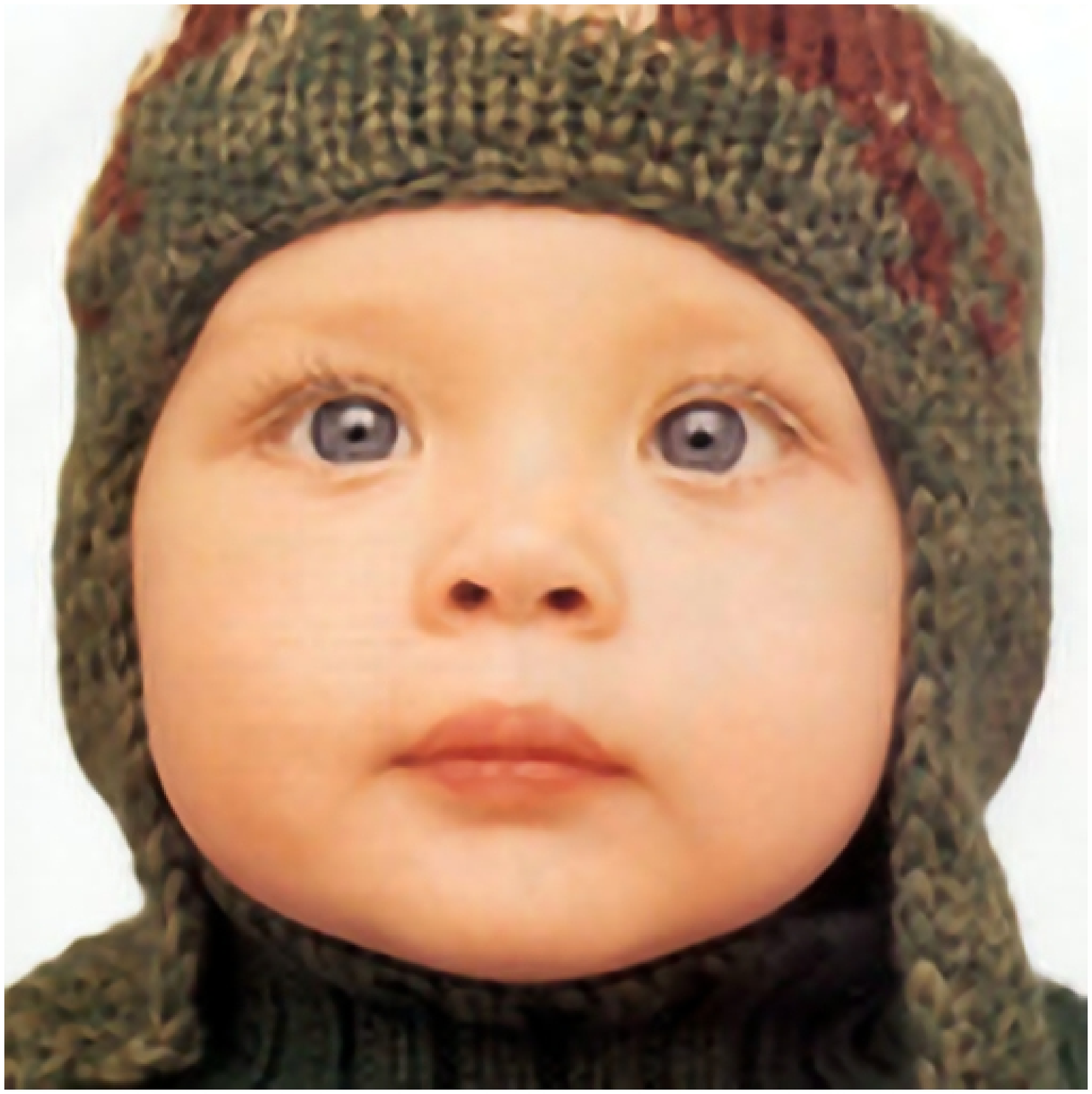}
\put(51,-41){{\includegraphics[viewport = 120 320 220 400, clip, height=1.55cm]{Child_KRR.eps}}}
\put(0,-41){{\includegraphics[viewport = 220 450 320 530, clip, height=1.55cm]{Child_KRR.eps}}}
\put(1,1){\sffamily \footnotesize{\textcolor{white}{{Kim \emph{et al.}'s}}}}
\end{overpic}
\begin{overpic}[height=4.cm]{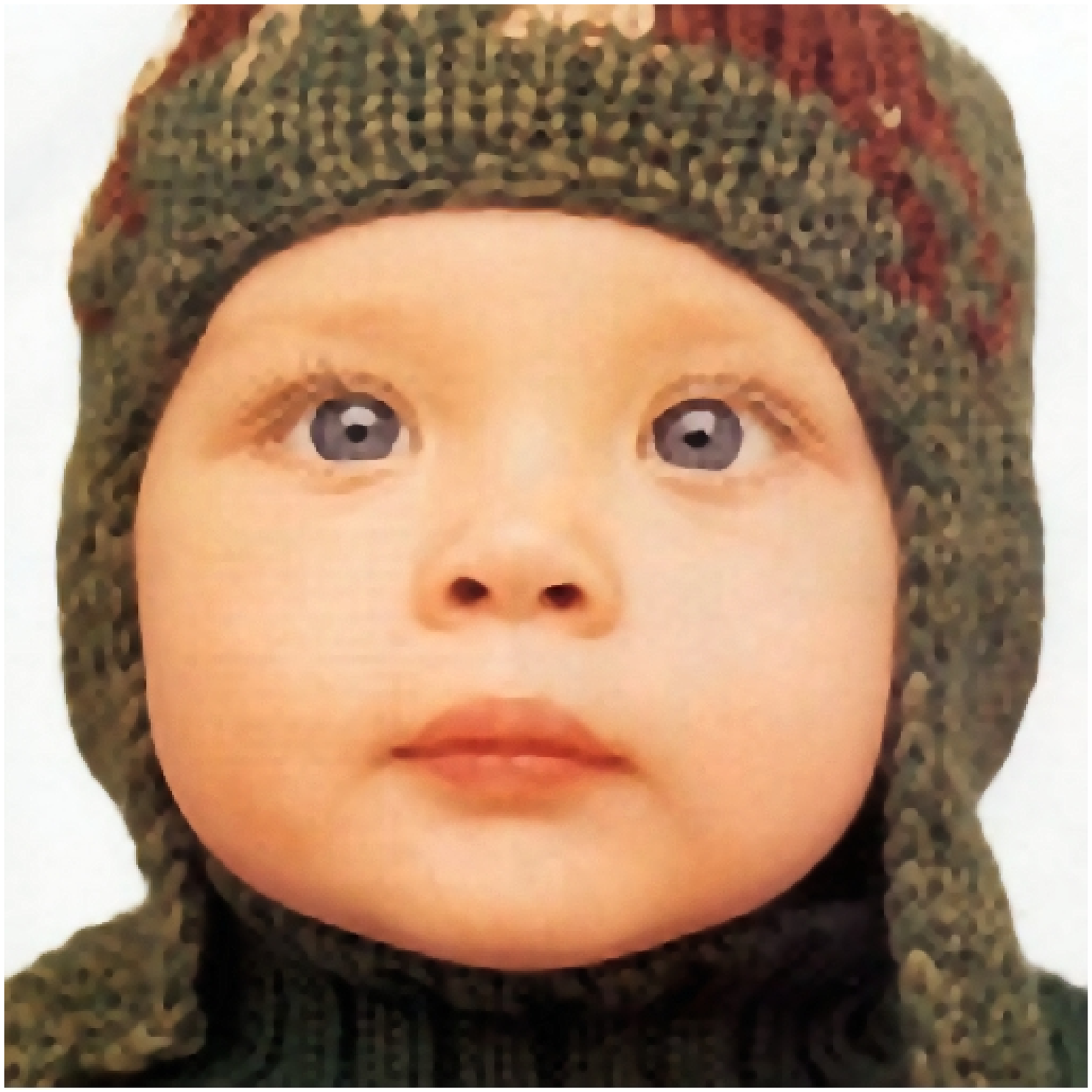}
\put(51,-41){{\includegraphics[viewport = 120 320 220 400, clip, height=1.55cm]{Child_Fattal.eps}}}
\put(0,-41){{\includegraphics[viewport = 220 450 320 530, clip, height=1.55cm]{Child_Fattal.eps}}}
\put(1,1){\sffamily \footnotesize{\textcolor{white}{{Fattal's}}}}
\end{overpic}\\
\vspace{0.8in}
\begin{overpic}[height=4.cm]{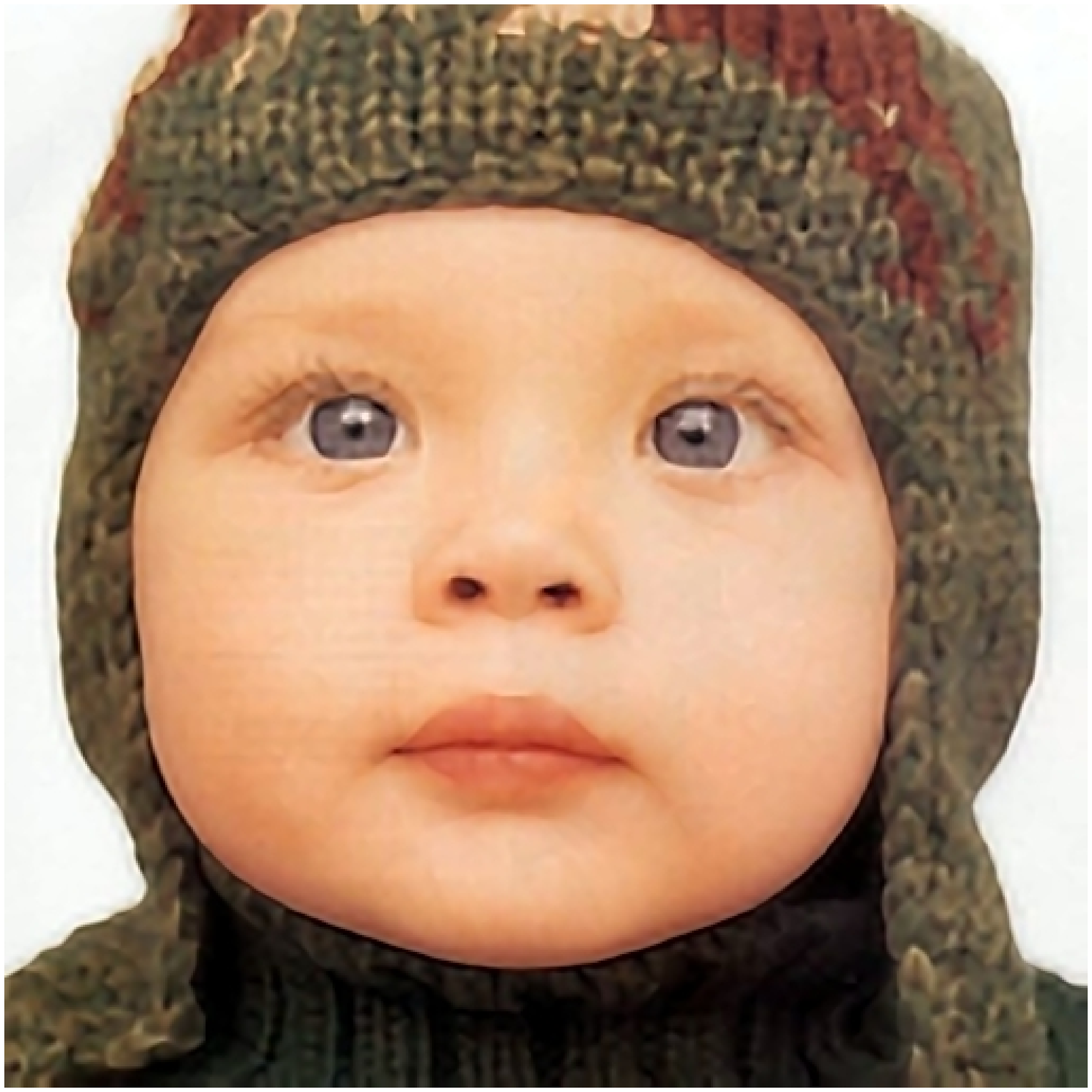} 
\put(51,-41){{\includegraphics[viewport = 120 320 220 400, clip, height=1.55cm]{Child_Irani.eps}}}
\put(0,-41){{\includegraphics[viewport = 220 450 320 530, clip, height=1.55cm]{Child_Irani.eps}}}
\put(1,1){\sffamily \footnotesize{\textcolor{white}{Glasner \emph{et al.}'s}}}
\end{overpic}
\begin{overpic}[height=4.cm]{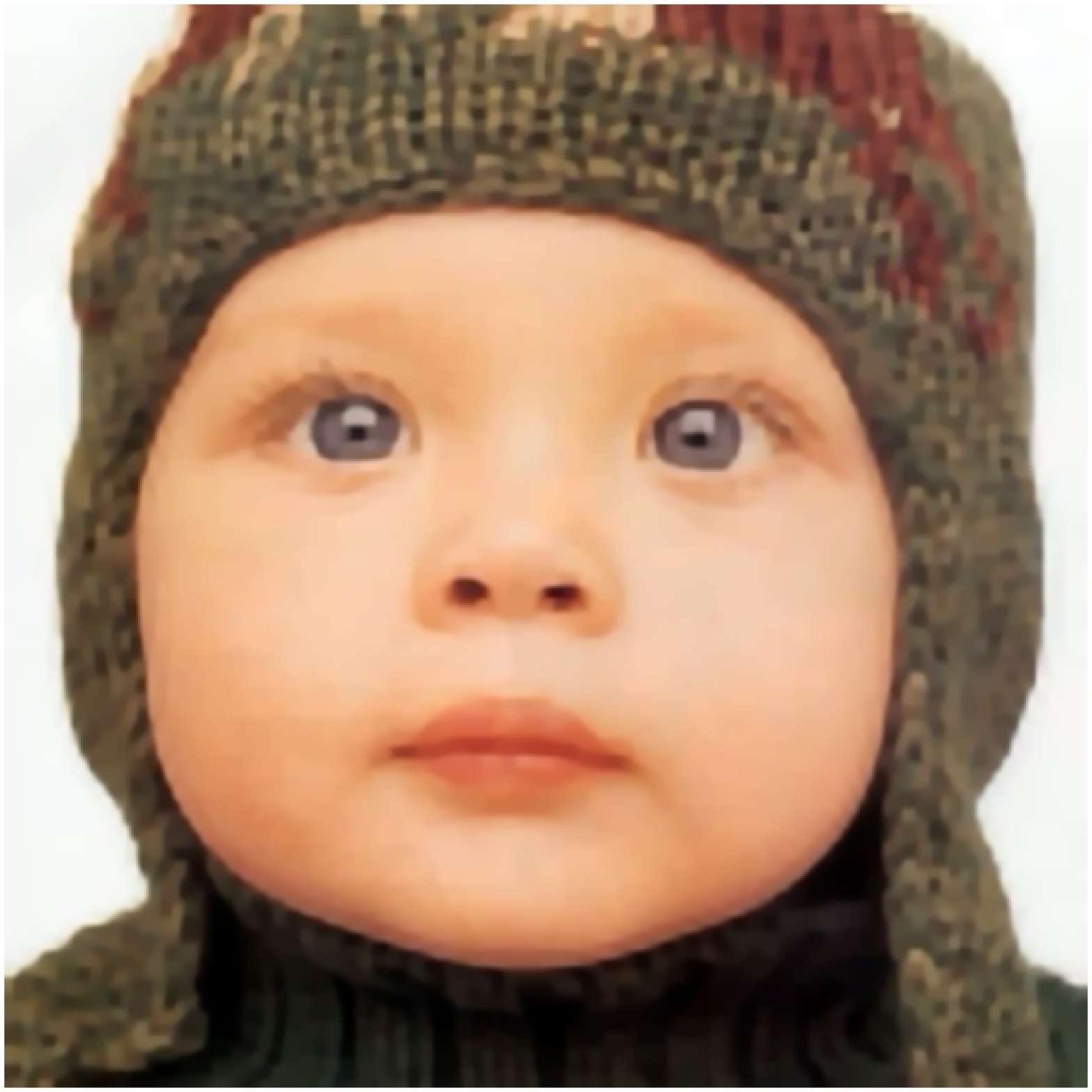} 
\put(51,-41){{\includegraphics[viewport = 120 320 220 400, clip, height=1.55cm]{Child_Shan.eps}}}
\put(0,-41){{\includegraphics[viewport = 220 450 320 530, clip, height=1.55cm]{Child_Shan.eps}}}
\put(1,1){\sffamily \footnotesize{\textcolor{white}{Shan \emph{et al.}'s}}}
\end{overpic}
\begin{overpic}[height=4.cm]{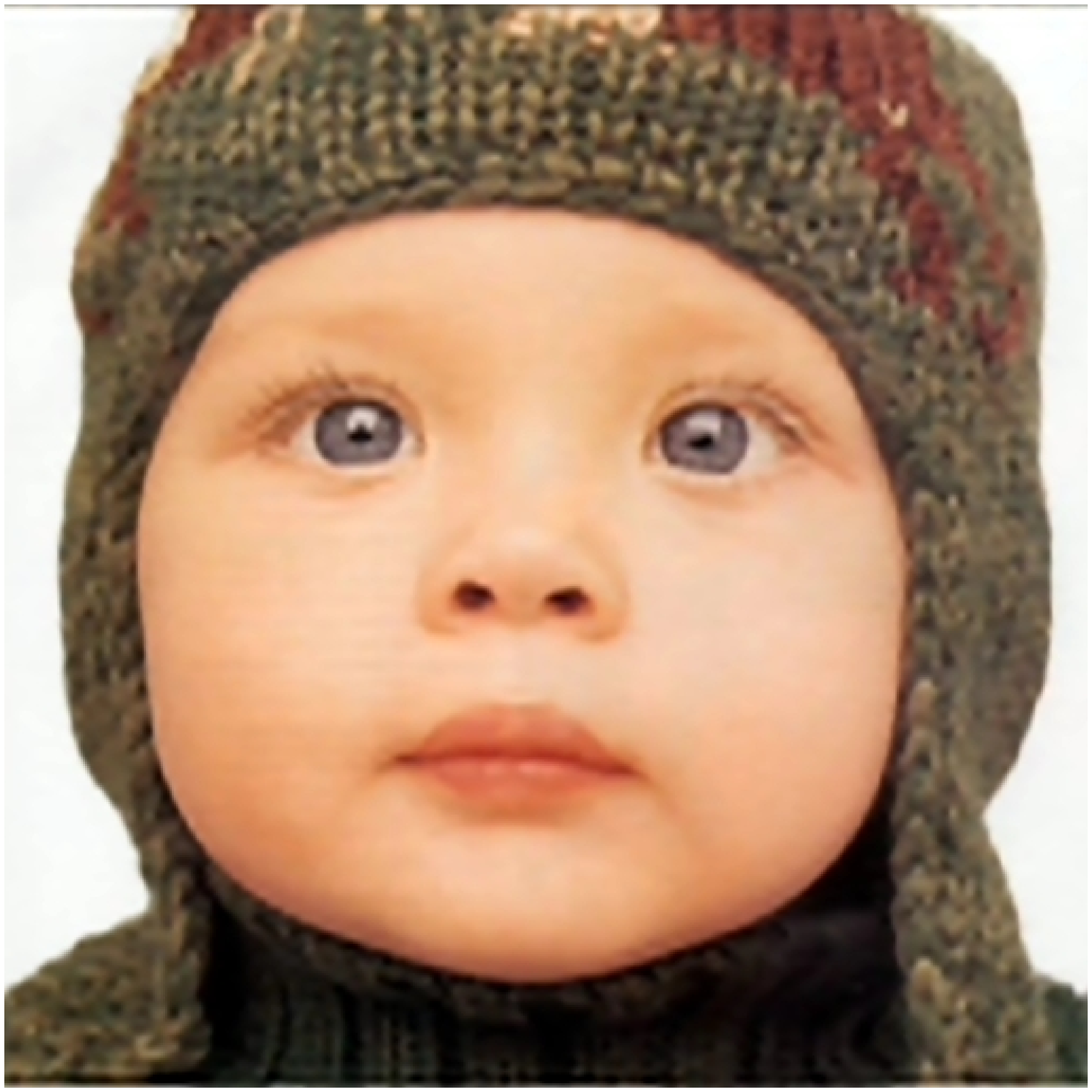} 
\put(51,-41){{\includegraphics[viewport = 120 320 220 400, clip, height=1.55cm]{child_HR_PSNR31.8001_SSIM0.8846.eps}}}
\put(0,-41){{\includegraphics[viewport = 220 450 320 530, clip, height=1.55cm]{child_HR_PSNR31.8001_SSIM0.8846.eps}}}
\put(1,1){\sffamily \footnotesize{\textcolor{white}{\textsc{NBSR}}}}
\end{overpic}
\begin{overpic}[height=4.cm]{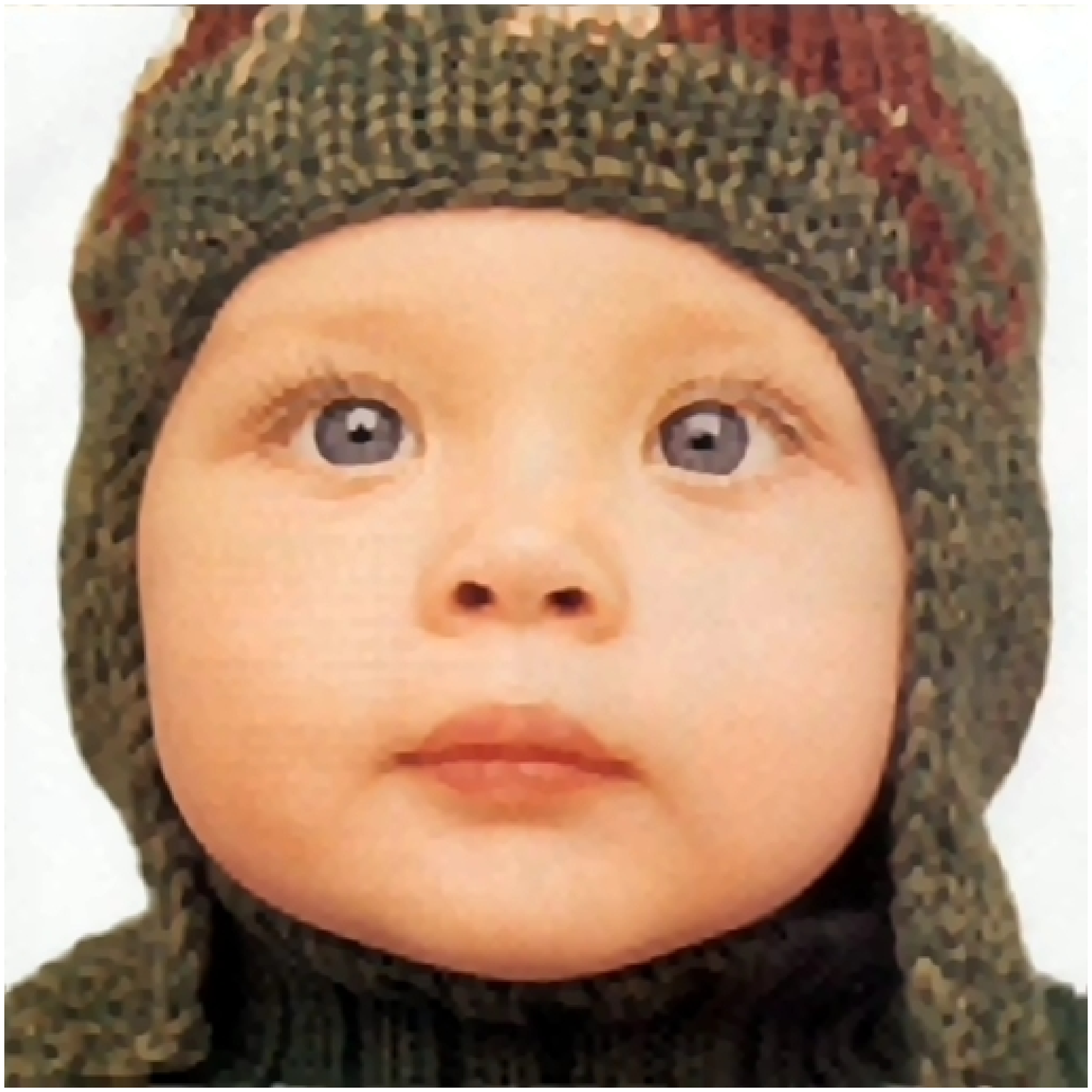}
\put(51,-41){{\includegraphics[viewport = 120 320 220 400, clip, height=1.55cm]{child_EBSR_r4_HR_PSNR33.002738SSIM0.885398.eps}}}
\put(0,-41){{\includegraphics[viewport = 220 450 320 530, clip, height=1.55cm]{child_EBSR_r4_HR_PSNR33.002738SSIM0.885398.eps}}}
\put(1,1){\sffamily \footnotesize{\textcolor{white}{\textsc{Proposed}}}}
\end{overpic}
\vspace{0.5in}
\caption{Single image super resolution results with different algorithms ($\times 4$). The results are generated with: Nearest Neighbor interpolation (26.05, 0.741), Bicubic interpolation (26.92, 0.785), Freeman \emph{et al.}'s example based SR method (25.38, 0.729), Kim \emph{et al.}'s kernel regression based method (31.22, 0.863)~\cite{Kim_SR_PAMI}, Fattal \emph{et al.}'s edge statistics-based SR method (25.55, 0.751)~\cite{Fattal_Edge}, Glasner \emph{et al.}'s method (25.28, 0.741)~\cite{Glasner_super-resolutionfrom}, Shan \emph{et al.}'s method (24.43, 0.727)~\cite{Shan_SR}, NBSR~\cite{NBSR_Zhang} ({31.80}, {0.8846}) and the proposed EBSR method (\bf{33.00}, \bf{0.8854}).}
\label{fig:Res_SR_More}
\end{figure*}

 More color image SR results are shown in Figure~\ref{fig:Res_SR_color_VBSR} and Figure~\ref{fig:Res_SR_More}, with the zooming factor of $4$.
We compare the SR result from our proposed method  with the results from several SR methods in the literature:
 the results from Freeman's example-based  SR method~\cite{Freeman_SR}, Kim's kernel regression method~\cite{Kim_SR_PAMI}, Glasner's method~\cite{Glasner_super-resolutionfrom}, Shan \emph{et al.}'s fast SR method~\cite{Shan_SR}, Yang \emph{et al.}'s sparse representation based SR method~\cite{Yang_SR_TIP} as well as the Variational Bayesian based SR method~\cite{VBSR}.
{\rv From Figure~\ref{fig:Res_SR_color_VBSR}, we can see that the fast SR method again over-smooths the HR image,  and the ScSR method generates HR images with artifacts along the edges. The HR image from the VBSR method suffers from ringing artifacts. The proposed method, on the other hand, generates HR image with less artifacts and with desirable quality both visually and  in terms of PSNR and SSIM, and with comparable (slightly better) quality than the results of NBSR.
More SR algorithms are compared in Figure~\ref{fig:Res_SR_More}.
It can be observed from Figure~\ref{fig:Res_SR_More} that the SR results from
Freeman's example-based  SR method and Glasner's method although have visual resolution improvement, they actually hallcinating some high resolution information that is not fidelity to the original image, thus hindering the improvement in terms of PSNR and SSIM.
Fattal's method and Shan's method appear to suffer from cartoon-like artifacts.
The proposed method is visually much better than results via NN; also, it has less artifacts compared to the result from example-based  SR method~\cite{Freeman_SR},  and is sharper than result from Kim's kernel regression method~\cite{Kim_SR_PAMI} and the NBSR method~\cite{NBSR_Zhang}, and achieves the best performance in terms of PSNR and SSIM.}
The results on both gray and color image SR  all  verified the effectiveness of the proposed  method.

\subsection{Noisy Image Super Resolution}
\begin{figure}[t]
\centering
\begin{overpic}[viewport = 10 5 385 300, clip, width=5cm]{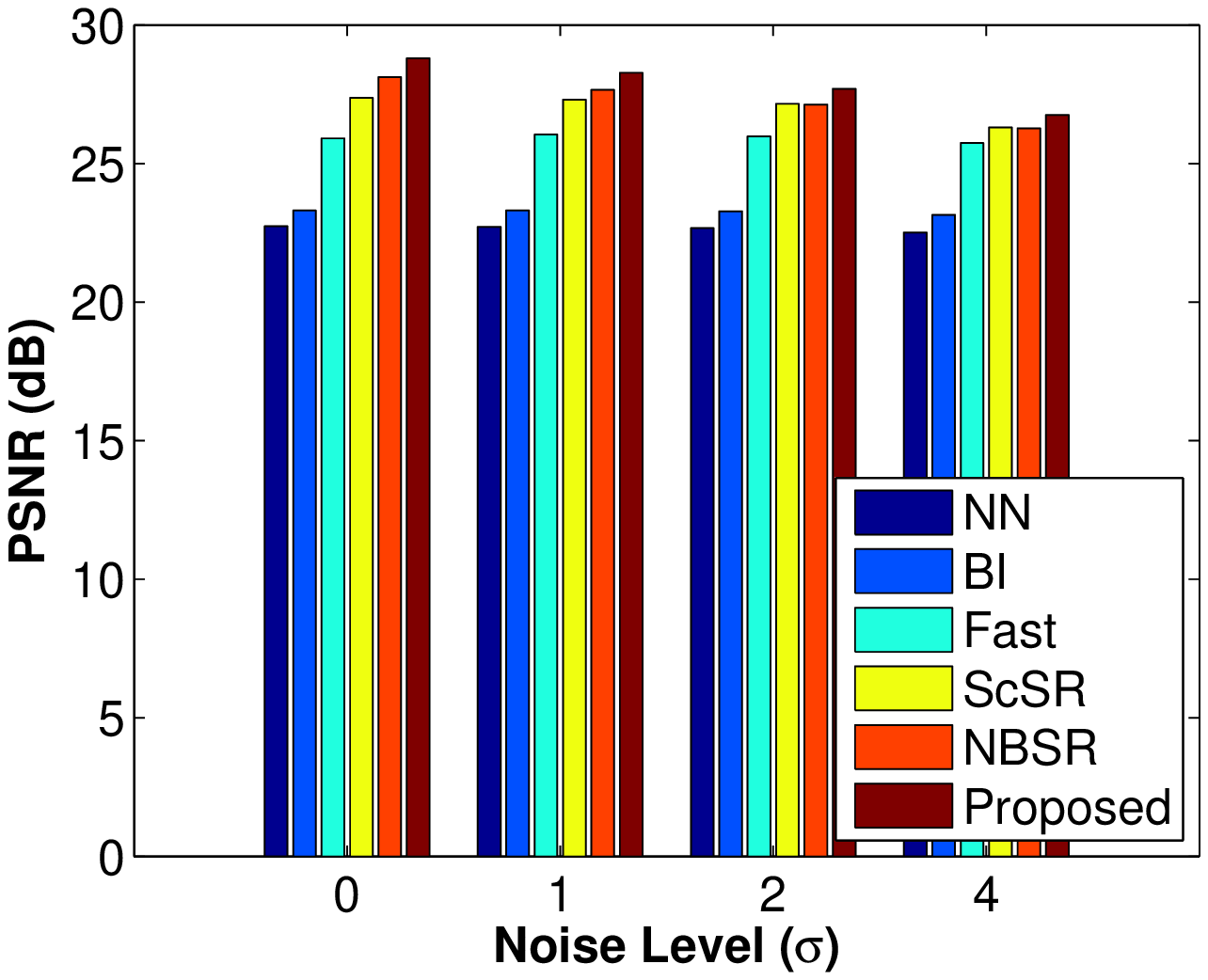}
\put(50,-6){ \sffamily \footnotesize{\textcolor{black}{{(a)}}}}
\end{overpic}
\hspace{0.15in}
\begin{overpic}[viewport = 10 5 385 300, clip, width=5cm]{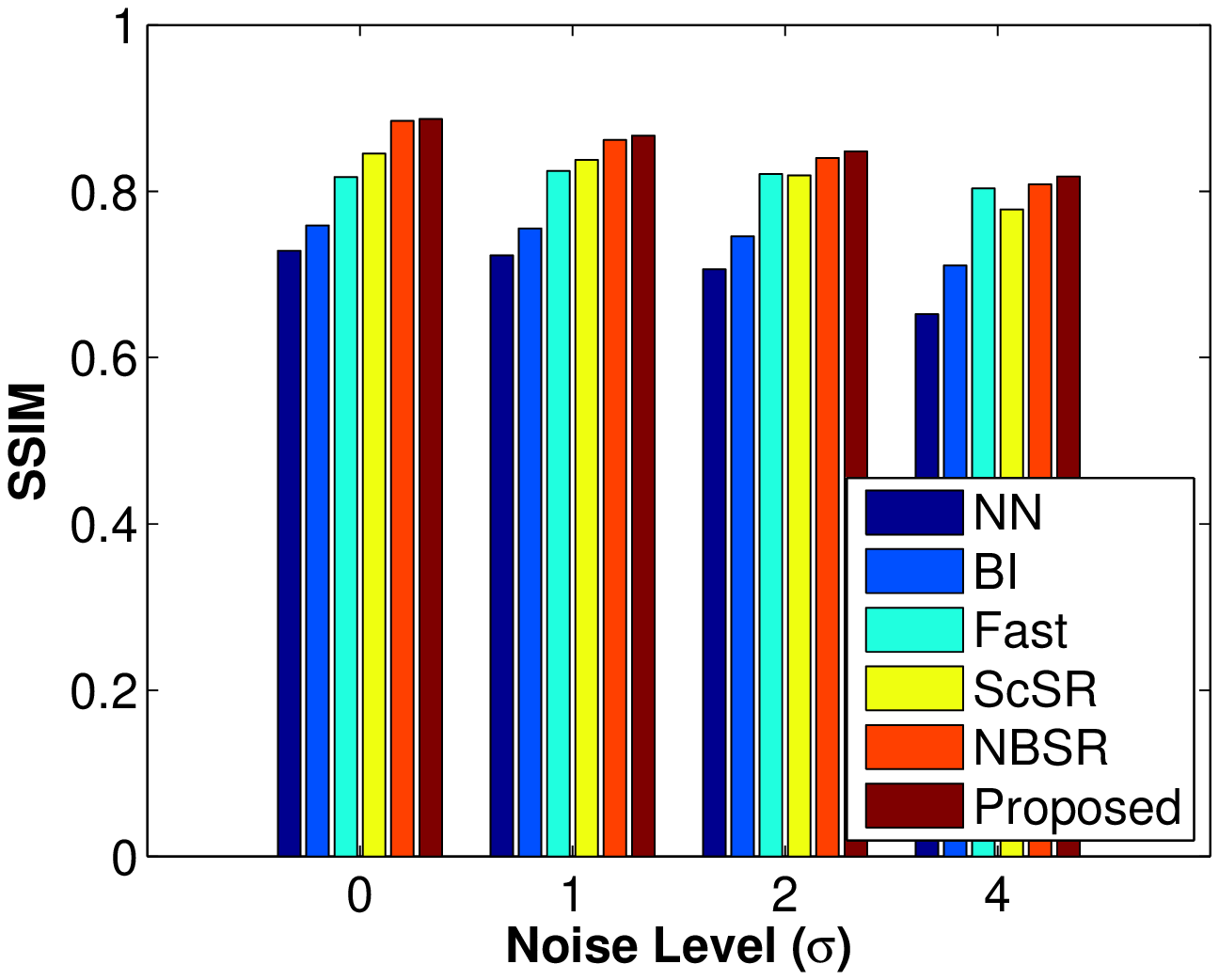}
\put(50,-6){\sffamily \footnotesize{\textcolor{black}{{(b)}}}}
\end{overpic}
\caption{Image SR results. (a) PSNR bar plot  and (b) SSIM bar plot with respect to increasing noise level ($\sigma \in \{0, 1, 2, 4\}$).}
\label{fig:NOisy_SR_Bar_Plot}
\end{figure}

In real-world SR tasks, the observed LR images are rarely noise-free, but are often contaminated by noise.
Therefore, it is important for the SR algorithm to be robust to the noise contained in the LR observations.
In this subsection, we evaluate the effectiveness of the proposed approach in the case of noisy SR situation, \emph{i.e.}, the given  input is both of low in resolution and is contaminated by noise.
The SR estimation results under the noise levels $\sigma\in\{0, 1, 2, 4\}$ are reported in Table~\ref{table:SR_Noisy}.
The  bar plots for the evaluated algorithms are shown in Figure~\ref{fig:NOisy_SR_Bar_Plot}.
The  PSNR and SSIM performance shown in Figure~\ref{fig:NOisy_SR_Bar_Plot} is the average performance over the test images.
As can been seen from Table~\ref{table:SR_Noisy} and Figure~\ref{fig:NOisy_SR_Bar_Plot},
the proposed method performs better than the other methods when the noise level is low;
as the noise level increases, the performances in terms of PSNR and SSIM measures of all the SR algorithms decreases.
In this case, the performance of the proposed method is still better than the other algorithms in general in terms of SSIM and is comparable to ScSR in terms of PSNR, indicating its robustness to the noise in the LR images and its applicability to general SR scenarios. It is worthwhile to point out that for the ScSR method, it is  given the ground truth noise level, and its regularization weight has to be adjusted according to the noise level.\footnote{In our experiments, we adjust the regularization weight the same way as suggested in \cite{Yang_SR_TIP}.} The proposed method, on the other hand,  does not require the noise level to be known, but can infer it from the noise LR image itself, while generating  results  comparable or better than those from ScSR, demonstrating its effectiveness for real-world SR tasks.

\section{Conclusion}
\label{sec:con}
An effective and efficient Bayeisna image SR algorithm with natural image prior is proposed in this paper.
The proposed method exploits the natural image statistics for image SR with a flexible high-order MRF model.
Specifically, an FoE model with products of GSM  potentials  is used for learning the prior model from natural images, which is further incorporated into a fully Bayesian framework for image SR.
Moreover, the GSM is adapted to the content of the HR image alongside with the estimation process of the HR image.
To estimate the latent HR image, we use the empirical Bayesian approach, which can generate high quality estimation while avoiding the time-consuming sampling.
Experimental results  under different settings compared with several \emph{state-of-the-art} SR algorithms in the literature verify the effectiveness of the proposed method.
Specifically, compared with our previous  sampling based SR approach~\cite{NBSR_Zhang}, the method proposed in this paper
can generate SR results with comparable or better quality than our previous  sampling based SR approach~\cite{NBSR_Zhang}, while with far less computational cost.
For future work, we would like to extend our current approach to other related inverse problems, such as multi-frame super resolution.
The extension of the current work to blind deblurring is our current ongoing work, which is encouraging based on our now obtained results.

\bibliographystyle{IEEEbib}
{
\bibliography{EBSR}
}

\end{document}